%% file: main.tex
\documentclass[11pt]{article}
\usepackage[left=1in, right=1in, top=1in]{geometry}
\usepackage{xargs}
\usepackage{natbib}
\usepackage{fancyhdr}
\usepackage{setspace}
\usepackage{lastpage}
\usepackage{upgreek}
\usepackage{amsmath,mathtools,amsthm}	
\usepackage{amssymb,dsfont,bbm}	
\usepackage{xcolor}  
\usepackage{bm}

\setcitestyle{authoryear}

\usepackage[colorlinks=true,breaklinks=true,bookmarks=true,urlcolor=blue,
     citecolor=blue,linkcolor=blue,bookmarksopen=false,draft=false]{hyperref}

\RequirePackage{graphicx}

\input{header.tex}
\input{def.tex}

\begin{document}

\title{Finite-time High-probability Bounds for Polyak-Ruppert Averaged Iterates of Linear Stochastic Approximation}

\author{A. Durmus~\footnote{Ecole Polytechnique, Paris, \texttt{alain.durmus@polytechnique.edu}. }, \, E. Moulines~\footnote{Ecole Polytechnique, Paris, \texttt{eric.moulines@polytechnique.edu}.}, \, A. Naumov~\footnote{HSE University, Russia,  \texttt{anaumov@hse.ru}.}, \, S. Samsonov~\footnote{HSE University, Russia,  \texttt{svsamsonov@hse.ru}.}}
\maketitle

\begin{abstract}
  This paper provides a finite-time analysis of linear stochastic approximation (LSA) algorithms with fixed step size, a core method in statistics and machine learning. LSA is used to compute approximate solutions of a $d$-dimensional linear system $\bar{\mathbf{A}} \theta = \bar{\mathbf{b}}$ for which $(\bar{\mathbf{A}}, \bar{\mathbf{b}})$ can only be estimated by (asymptotically) unbiased observations $\{(\mathbf{A}(Z_n),\mathbf{b}(Z_n))\}_{n \in \mathbb{N}}$.
 We consider here the case where $\{Z_n\}_{n \in \mathbb{N}}$ is an i.i.d. sequence or a uniformly geometrically ergodic Markov chain. We derive $p$-th moment and high-probability deviation bounds for the iterates defined by LSA and its Polyak-Ruppert-averaged version. Our finite-time instance-dependent bounds for the averaged LSA iterates are sharp in the sense that the leading term we obtain coincides with the local asymptotic minimax limit. Moreover, the remainder terms of our bounds admit a tight dependence on the mixing time $\taumix$ of the underlying chain and the norm of the noise variables. We emphasize that our result requires the SA step size to scale only with logarithm of the problem dimension $d$.
\end{abstract}

\section{Introduction}
\input{introduction.tex}

\section{Finite-time Moment and  High-probability Bounds in the Independent Noise Setting}
\label{sec:independent_case}
\input{independent_case}

\section{Finite-time Moment and  High-probability Bounds in the Markovian Noise Setting}
\label{sec:markov_case}
\input{markov_case}

\appendix

\section{Notations and Constants}
\label{appendix:constants}
\input{appendix_const}

\section{Independent case bounds}
\label{appendix:iid}
\input{appendix_iid}

\section{Independent case bounds under sub-Gaussian noise assumption}
\label{appendix:iid_subgaus}
\input{appendix_iid_subgaussian}

\section{Markov case bounds}
\label{appendix:Markov}
\input{appendix_markov}

\section{Technical bounds: Markov case}
\label{appendix:Markov_technical_lemmas}
\input{appendix_markov_technical}

\section{Technical lemmas}
\label{appendix:technical}
\input{appendix_technical}

\bibliography{references}

\end{document}

%% file: header.tex
\usepackage{xcolor}
\usepackage{srcltx}
\usepackage{etoolbox}
\usepackage{nameref}
\usepackage{comment}
\usepackage{mathrsfs}
\usepackage{enumerate}
\usepackage[shortlabels]{enumitem}
\usepackage{amsfonts} 
\usepackage{nicefrac} 
\usepackage{mathtools}
\usepackage{dsfont,mathrsfs}
\usepackage{cleveref}
\usepackage{xargs}
\usepackage{multirow}

\usepackage{autonum}
\usepackage{upgreek}

\newtheorem{assum}{A\hspace{-2pt}}

\newtheorem{assumID}{IND\hspace{-2pt}}

\newtheorem{theorem}{Theorem}
\crefname{theorem}{theorem}{Theorems}
\Crefname{theorem}{Theorem}{Theorems}

\newtheorem{lemma}{Lemma}
\crefname{lemma}{lemma}{lemmas}
\Crefname{lemma}{Lemma}{Lemmas}

\newtheorem{remark}{Remark}
\crefname{remark}{remark}{remarks}
\Crefname{remark}{Remark}{Remarks}

\newtheorem{corollary}{Corollary}
\crefname{corollary}{corollary}{corollaries}
\Crefname{corollary}{Corollary}{Corollaries}

\newtheorem{proposition}{Proposition}
\crefname{proposition}{proposition}{propositions}
\Crefname{proposition}{Proposition}{Propositions}

\crefname{definition}{definition}{definitions}
\Crefname{Definition}{Definition}{Definitions}

\crefname{example}{example}{examples}
\Crefname{Example}{Example}{Examples}

\crefname{figure}{figure}{figures}
\Crefname{Figure}{Figure}{Figures}

\crefname{table}{table}{tables}
\Crefname{Table}{Table}{Tables}

\crefname{assum}{A\hspace{-2pt}}{A\hspace{-2pt}}
\crefname{assumb}{B\hspace{-2pt}}{B\hspace{-2pt}}
\crefname{assumUGE}{UGE\hspace{-1pt}}{UGE\hspace{-1pt}}
\crefname{assumID}{IND\hspace{-1pt}}{IND\hspace{-1pt}}
\crefname{assumUE}{UE\hspace{-1pt}}{UE\hspace{-1pt}}
\crefname{assumSUP}{M\hspace{-1pt}}{M\hspace{-1pt}}

\newlist{renumerate}{enumerate}{3}
\setlist[renumerate]{wide, labelwidth=!, labelindent=0pt,label=(\roman*)}

\newlist{aenumerate}{enumerate}{3}
\setlist[aenumerate]{wide, labelwidth=!, labelindent=0pt,label=(\arabic*)}

\newlist{aaenumerate}{enumerate}{3}
\setlist[aaenumerate]{wide, labelwidth=!, labelindent=0pt,label=(\alph*)}

\newlist{aenumerateSpace}{enumerate}{3}
\setlist[aenumerateSpace]{wide, labelwidth=!,label=(\arabic*)}

\newlist{benumerate}{enumerate}{3}
\setlist[benumerate]{wide, labelwidth=!, labelindent=0pt,label=$\bullet$}

%% file: def.tex
\def\supconsteps{\supnorm{\funnoisew}}
\def\lesssimd{\lesssim_{d}}

\newcommand{\PE}{\mathbb{E}}
\newcommand{\PP}{\mathbb{P}}
\newcommandx{\genericb}[1][1=]{b_{#1}}
\newcommandx{\Constros}[1][1=]{\operatorname{C}_{\operatorname{Ros},#1}}
\newcommandx{\Constburk}[1][1=]{\operatorname{C}_{\operatorname{Burk}}}
\newcommandx{\driftW}[1][1=]{W_{#1}}

\def\Auxconst{\operatorname{c}}

\def\metricz{\mathsf{d}_{\Zset}}
\newcommandx{\metricd}[1][1=]{\mathsf{d}_{#1}}

\newcommandx\invmeasure[1][1=]{\Pi_{#1}}
\newcommandx{\PPjoint}[1][1=]{\PP^{\MKjoint[#1]}}
\newcommandx{\PEjoint}[1][1=]{\PE^{\MKjoint[#1]}}
\newcommandx{\PEMID}[1][1=\alpha]{\PE^{\MK[#1]}}
\newcommandx{\PPMID}[1][1=\alpha]{\PP^{\MK[#1]}}

\newcommand{\supnorm}[1]{\norm{ #1 }[\infty]}
\newcommandx{\MKjoint}[1][1=]{\bar{\operatorname{P}}_{#1}}
\newcommandx\costw[1][1=]{\mathsf{c}_{#1}}

\newcommandx\Intergrdist[1][1=]{\mathbb{M}_{1}(#1)}

\newcommandx{\mmarkov}[1][1=0]{m^{(\Markov)}_{#1}}

\def\AuxconstM{c^{(\operatorname{M})}}

\def\Deltatr{\Delta^{\operatorname{(tr)}}}
\def\Deltafl{\Delta^{\operatorname{(fl)}}}
\def\MarDeltatr{R^{\operatorname{(tr)}}}
\def\MarDeltafl{R^{\operatorname{(fl)}}}

\def\Zset{\mathsf{Z}}
\def\Zsigma{\mathcal{Z}}

\def\rset{\mathbb{R}}

\def\nset{\ensuremath{\mathbb{N}}}
\def\nsets{\ensuremath{\mathbb{N}^*}}
\def\nsetm{\ensuremath{\mathbb{N}_-}}
\def\zset{\ensuremath{\mathbb{Z}}}

\newcommand{\round}[1]{\ensuremath{\lfloor#1\rfloor}}

\newcommand{\msi}{\mathsf{I}}
\newcommand{\msj}{\mathsf{J}}

\newcommand{\SG}{\operatorname{SG}}
\newcommand{\Mat}[1]{{\bf{#1}}}
\def\MatB{B}
\def\MatP{P}

\newcommand{\Const}[1]{\operatorname{C}_{{#1}}}

\newcommand{\bConst}[1]{\operatorname{C}_{{\bf #1}}}

\newcommand{\smallConst}[1]{\operatorname{c}_{{#1}}}
\newcommand{\smallConstm}[1]{\operatorname{c}_{{#1}}^{(\Markov)} }

\def\thetainit{\theta_0}

\newcommandx\sequence[4][2=,3=,4=]
{\ifthenelse{\equal{#3}{}}{\ensuremath{\{ #1_{#2 #4}\}}}{\ensuremath{\{ #1_{#2 #4} \}_{#2 \in #3}}}}

\newcommandx\sequenceD[2][2=]
{\ifthenelse{\equal{#2}{}}{\ensuremath{\{ #1\}}}{\ensuremath{\{ #1\!~:\!~#2  \}}}}
\newcommandx\sequenceDouble[4][3=,4=]
{\ifthenelse{\equal{#3}{}}{\ensuremath{\{ (#1_{#3},#2_{#3})\}}}{\ensuremath{\{ (#1_{#3},#2_{#3})\}_{#3 \in #4}}}}

\newcommandx{\sequencen}[2][2=n\in\nset]{\ensuremath{\{ #1, \eqsp #2 \}}}
\newcommandx\sequencens[2][2=n]
{\ensuremath{\{ #1_{#2} \!~:\!~#2\in\nsets\}}}
\newcommandx\sequencet[4]
{\ensuremath{\{ #1{#2_{#3}} \, : \, \eqsp #3 \in #4 \}}}
\def\PE{\mathbb{E}}
\def\P{\mathbb{P}}
\def\ProdB{\Gamma}

\def\ProdBa{\ProdB^{(\alpha)}}

\newcommandx{\PVar}[1][1=]{\ensuremath{\operatorname{Var}_{#1}}}

\def\noisecov{\Sigma_\varepsilon}
\def\noisecovmarkov{\Sigma^{(\Markov)}_\varepsilon}

\newcommandx{\MK}[1][1=\alpha]{\mathrm{P}_{#1}}
\newcommandx\MKK[1][1=\alpha]{\mathrm{K}_{#1}}
\def\MKQ{\mathrm{Q}}

\newcommandx{\PEtilde}[1][1=]{\PE^{\mathrm{K}_{#1}}}
\newcommandx{\PPtilde}[1][1=]{\PP^{\mathrm{K}_{#1}}}
\newcommand{\PEext}{\tilde{\PE}}
\newcommand{\PPext}{\tilde{\PP}}

\newcommand{\abs}[1]{\vert #1\vert}

\newcommand{\absD}[1]{\left\vert #1\right\vert}
\newcommandx{\norm}[2][2=]{\Vert#1 \Vert_{{#2}}}
\newcommandx{\normLigne}[2][2=]{\Vert#1 \Vert_{{#2}}}
\newcommandx{\normLine}[2][2=]{\Vert#1 \Vert_{{#2}}}
\newcommandx{\normop}[2][2=]{\Vert{#1}\Vert_{{#2}}}
\newcommandx{\normopLigne}[2][2=]{\Vert{#1}\Vert_{{#2}}}
\newcommandx{\normopLine}[2][2=]{\Vert{#1}\Vert_{{#2}}}
\newcommandx{\osc}[2][1=]{\mathrm{osc}_{#1}(#2)}

\newcommandx{\normlip}[2][2=\operatorname{Lip}]{\Vert#1 \Vert_{{#2}}}
\newcommand{\lip}{\operatorname{L}}
\newcommandx{\lipspace}[1]{\lip_{#1}}

\newcommandx{\CPP}[3][1=]
{\ifthenelse{\equal{#1}{}}{{\mathbb P}\left(\left. #2 \, \right| #3 \right)}{{\mathbb P}_{#1}\left(\left. #2 \, \right | #3 \right)}}
\newcommandx{\CPPtilde}[3][1=]
{\ifthenelse{\equal{#1}{}}{{\tilde{\mathbb P}}\left(\left. #2 \, \right| #3 \right)}{{\tilde{\mathbb P}}_{#1}\left(\left. #2 \, \right | #3 \right)}}

\def\wrt{with respect to}

\def\rhs{right-hand side}

\def\iid{i.i.d.}

\newcommandx{\as}[1][1=\PP]{\ensuremath{#1\, -\mathrm{a.s.}}}

\newcommand{\ie}{i.e.}

\newcommand{\eqsp}{\;}

\newcommand{\Id}{\mathrm{I}}
\def\prtheta{\bar{\theta}}

\def\utheta{\tilde{\theta}^{\sf (tr)}}
\def\vtheta{\tilde{\theta}^{\sf (fl)}}

\newcommand{\ConstD}{\mathsf{D}}
\newcommand{\ConstDM}{\ConstD^{(\operatorname{M})}}

\newcommandx{\boundmetric}[1][1=]{\kappa_{\MKK[#1]}}

\newcommand{\Jnalpha}[2]{J_{#1}^{(#2)}}
\newcommand{\Hnalpha}[2]{H_{#1}^{(#2)}}

\newcommand{\coint}[1]{\left[#1\right)}

\newcommand{\ocint}[1]{\left(#1\right]}
\newcommand{\ocintLine}[1]{(#1]}
\newcommand{\ooint}[1]{\left(#1\right)}

\newcommand{\ccint}[1]{\left[#1\right]}

\newcommandx{\Nnorm}[2][1=V]{[ #2]_{#1}}
\newcommandx{\lipnorm}[2][1=g]{[ #1]_{#2}}

\newcommandx{\CPE}[3][1=]{{\mathbb E}^{#3}_{#1}\left[#2\right]}
\newcommandx{\CPEext}[3][1=]{\tilde{\mathbb E}^{#3}_{#1}\left[#2\right]}
\newcommandx{\CPEtilde}[3][1=]{{\tilde{\mathbb E}}^{#3}_{#1}\left[#2\right]}
\newcommandx{\CPEs}[3][1=]{{\mathbb E}^{#3}_{#1}[#2]}

\def\thetalim{\theta^\star}

\def\trace{\operatorname{Tr}}

\newcommand{\rme}{\mathrm{e}}
\newcommand{\rmd}{\mathrm{d}}
\def\funcAw{\mathbf{A}}
\def\funcBw{\mathbf{B}}
\newcommand{\funcA}[1]{\funcAw(#1)}

\def\funcbw{\mathbf{b}}
\newcommand{\funcb}[1]{\funcbw(#1)}
\newcommandx{\zmfuncA}[2][1=]{\tilde{\funcAw}^{#1}(#2)}
\newcommandx{\zmfuncAw}[1][1=]{\tilde{\funcAw}_{#1}}
\newcommandx{\zmfuncb}[2][1=]{\tilde{\funcbw}^{#1}(#2)}
\def\funnoisew{\varepsilon}
\newcommand{\funcnoise}[1]{\funnoisew(#1)}

\newcommandx{\funcct}[2][1=]{\funcctilde^{#1}(#2)}

\def\trace{\operatorname{Tr}}

\def\qcond{\kappa_{Q}}

\def\State{Z}
\def\state{z}

\def\taumix{t_{\operatorname{mix}}}

\newcommand{\indiacc}[1]{\boldsymbol{1}_{\{#1\}}}

\newcommand{\indinD}[1]{\boldsymbol{1}\{#1\}}

\newcommandx{\CovC}[1][1=u]{\operatorname{C}_{#1}}

\usepackage[textwidth=2.25cm,textsize=footnotesize]{todonotes}

\def\msa{\mathsf{A}}
\def\msb{\mathsf{B}}
\def\msz{\mathsf{Z}}
\def\mcz{\mathcal{Z}}

\def\plusinfty{+\infty}

\DeclareMathAlphabet{\mathpzc}{OT1}{pzc}{m}{it}

\def\lyapW{\mathpzc{W}}

\newcommand{\txts}{\textstyle}

\newcommandx{\bias}[1][1=\alpha]{\operatorname{B}_{#1}}

\newcommandx\probaMarkovTilde[2][2=]
{\ifthenelse{\equal{#2}{}}{{\widetilde{\mathbb{P}}_{#1}}}{\widetilde{\mathbb{P}}_{#1}\left[ #2\right]}}

\def\mcf{\mathcal{F}}

\newcommand{\parenthese}[1]{\left(#1 \right)}

\newcommand{\parentheseDeuxLigne}[1]{[ #1 ]}

\newcommand{\ceil}[1]{\left\lceil #1 \right\rceil}

\def\half{\nicefrac{1}{2}}

\def\bA{\bar{\mathbf{A}}}

\def\X{{\bf X}}
\def\Y{{\bf Y}}

\def\thetas{\thetalim}
\def\sphere{\mathbb{S}}
\def\realpart{\mathrm{Re}}
\def\transpose{\top}

\def\Am{{\bf A}}
\def\bm{{\bf b}}
\def\funcctilde{\tilde{c}_u}

\def\barb{\bar{\mathbf{b}}}
\newcommandx{\driftb}[1][1=p]{\bar{b}_{#1}}

\def\transpose{\top}

\def\Zbf{\mathbf{Z}}
\def\Abf{\mathbf{A}}
\def\Bbf{\mathbf{B}}

\def\bfA{\Abf}
\def\bfB{\Bbf}
\def\eps{\varepsilon}

\newcommandx{\boldb}[1][1={q}]{\mathsf{b}_{#1}}

\newcommandx{\ConstGW}[1][1={n,\lyapW}]{\operatorname{G}_{#1}}

\newcommandx{\ConstMW}[1][1={n,\lyapW}]{\operatorname{M}_{#1}}

\Crefname{assumptionC}{\textbf{C}\hspace{-1pt}}{\textbf{C}\hspace{-1pt}}
\crefname{assumptionC}{\textbf{C}}{\textbf{C}}

\newtheorem{assumptionM}{\textbf{UGE}\hspace{-1pt}}
\Crefname{assumptionM}{\textbf{UGE}\hspace{-1pt}}{\textbf{UGE}\hspace{-1pt}}
\crefname{assumptionM}{\textbf{UGE}}{\textbf{UGE}}

\def\distance{\mathsf{d}}

\newcommandx{\vartconstwas}[1][1=V]{c_{#1}}

\newcommandx{\deltawas}[1][1=*]{\delta_{#1}}

\newcommandx{\wasser}[4][1=\distance,4=]{\mathbf{W}_{#1}^{#4}\left(#2,#3\right)}
\newcommandx{\covcoeff}[2]{\rho_{#1}^{(#2)}}

\def\smallAconst{\smallConst{\funcAw}}
\def\smallAconstM{\smallConstm{\funcAw}}

\newcommand{\dobrush}{\mathsf{\Delta}}
\newcommandx{\dobru}[3][1=,3=]{\dobrush_{#1}^{#3}( #2)}  

\def\invariantQ{\pi}

\def\mtt{\mathtt{m}}

\def\HP{\mathrm{HP}}
\def\Etr{E^{{\sf tr}}}
\def\Efl{E^{{\sf fl}}}
\def\Gammabf{\boldsymbol{\Gamma}}
\def\checkA{\check{\mathbf{A}}}
\def\Ws{W^{\star}}

\def\tZs{\tilde{Z}^{\star}}
\def\tZ{\tilde{Z}}
\def\tmszn{\tilde{\msz}_{\nset}}
\def\tmczn{\tilde{\mcz}_{\nset}}
\def\qexponent{q}
\def\ppexponent{p}
\def\Markov{\mathrm{M}}
\def\bigO{\mathcal{O}}
\def\btheta{\bar{\theta}}

\newcommandx{\dlim}[1]{\ensuremath{\stackrel{#1}{\Longrightarrow}}}

%% file: introduction.tex
This paper is concerned with the linear stochastic approximation (LSA) algorithm for solving the linear system $\bA \theta = \barb$ with unique solution $\thetalim$, based on a sequence of observations $\{( \funcA{Z_n}, \funcb{Z_n})\}_{n \in \nset}$. Here $\Am: \msz \to \rset^{d \times d}$, $\bm: \msz \to \rset^d$ are measurable functions, and $(Z_k)_{k \in \nset}$ is
\begin{enumerate}[leftmargin=*, topsep=1mm]
\item  either an \iid\ sequence taking values in a state space $(\msz,\mcz)$ with distribution $\invariantQ$ satisfying $\PE [ \funcA{Z_1} ] = \bA$ and $\PE [ \funcb{Z_1} ] = \barb$;
\item or a $\msz$-valued ergodic Markov chain  with unique invariant distribution $\invariantQ$, such that \newline $\lim_{n \to \plusinfty} \PE [ \funcA{Z_n} ] = \bA$ and $\lim\nolimits_{n \to \plusinfty} \PE [ \funcb{Z_n} ] = \barb$.
\end{enumerate}
For a fixed step size $\alpha > 0$, burn-in period $n_0 \in \nset$, and  initialization $\theta_0$, we consider the sequences of estimates $\{\theta_n \}_{n \in \nset}, \{ \prtheta_{n} \}_{n \geq n_0+1}$ given by
\begin{equation}
\label{eq:lsa}
\begin{split}
\textstyle \theta_{k} &= \theta_{k-1} - \alpha \{ \funcA{Z_k} \theta_{k-1} - \funcb{Z_k} \} \eqsp,~~ k \geq 1, \\
\textstyle \prtheta_{n} &= (n-n_0)^{-1} \sum_{k=n_0}^{n-1} \theta_k \eqsp, ~~n \geq n_0+1 \eqsp.
\end{split}
\end{equation}
With a slight abuse of notation we drop the dependence upon $n_0$ in $\prtheta_{n}$. The sequence $\{ \theta_k \}_{k \in \nset}$ are the standard LSA iterates, while $\{ \prtheta_{n} \}_{n \geq n_0+1}$ corresponds to the Polyak-Ruppert (PR) averaged iterates; see \citet{ruppert1988efficient,polyak1992acceleration}.

The LSA algorithm is central in statistics, machine learning, and linear systems identification, see e.g. the works  \citet{eweda:macchi:1983,widrow1985adaptive,benveniste2012adaptive,kushner2003stochastic} and references therein. More recently, it has sparked a renewed interest in machine learning, especially for high-dimensional least squares and reinforcement learning (RL) problems; \citet{bertsekas2003parallel,bottou2018optimization,sutton:td:1988,bertsekas2019reinforcement,watkins1992q}. The LSA and LSA-PR recursions \eqref{eq:lsa} have been the subject of a wealth of work, and it is difficult to adequately acknowledge all contributions.
\citet{polyak1992acceleration,kushner2003stochastic,borkar:sa:2008,benveniste2012adaptive} provided asymptotic convergence guarantees (almost sure convergence, central limit theorem) under both \iid\ and Markovian noise settings. In particular, it has been established that LSA-PR can accelerate LSA and satisfies a central limit theorem with an asymptotically minimax-optimal covariance matrix.
\par
Although asymptotic convergence analysis is of theoretical interest, the current trend is to obtain nonasymptotic guarantees that take into account both the limited sample size and the dimension of the parameter space. For these reasons, non-asymptotic analysis of both \iid\ and Markovian SA procedures has recently attracted much attention.
\par
In the \iid\ setting,
\citet{rakhlin2012making,nemirovski2009robust,jain2018accelerating,pmlr-v99-jain19a}
investigated the finite-time mean squared error. Moreover, \citet{durmus2021stability} provided tight high-probability bounds for the LSA sequence $\{ \theta_n \}_{n \in \nset}$. For least squares regression problems with a symmetric matrix $\funcA{Z_n}$, \citet{bach:moulines:2013, jain2018parallelizing} showed that for a constant step size, the mean squared error (MSE) of
$\prtheta_{n} - \thetalim$ converges as ${\cal O}(1/n)$. For general
LSA, which includes instrumental variable methods for linear system identification and temporal differences in reinforcement learning
(TD), \citet{lakshminarayanan:2018a} showed a convergence rate of mean square error ${\cal O}(1/n)$. The LSA-PR procedure can also be viewed as a two-timescale SA algorithm, with Hoeffding-type non-asymptotic deviation bounds provided in \cite{dalal2020tale}. \citet{mou2020linear} provides a nonasymptotic high probability bound for LSA-PR with independent observations.
However, the proof of their main result \citet[Theorem 3]{mou2020linear} relies on tools from Markov chain theory that  assume strong conditions for
$\{(\funcAw(Z_n),\funcbw(Z_n)\}_{n \in\nset}$ and it is not clear how to adapt their method to the general case.
\par
For the Markovian setting, the literature is scarcer. Assuming an upper bound on the mixing time of the Markov chain, a projected variant of the LSA was analyzed by \citet{bhandari_finite_time}, yielding nonasymptotic rates of mean squared error (MSE) that are sharp in their dependence on sample size $n$ but not on dimension $d$. This result was later extended in \citet{srikant:1tsbounds:2019} with the analysis of LSA without the projection step. \citet{srikant:1tsbounds:2019} obtained the same convergence rate as \citet{bhandari_finite_time}. In \citet{chen2020explicit}, the authors obtained a sharp MSE bound for the last iteration of LSA assuming a $V$-uniformly ergodic Markov chain and decreasing  step sizes $\alpha_k = 1/k$. Recently, \citet{mou2021optimal} established
$p$-moment bounds for the last iterates of LSA and showed that the
mean-square error obtained with PR-LSA matches the local asymptotic minimax optimal limit.
\paragraph{Contributions and organization of the paper.}
Our main contribution is a unified framework for the finite-time analysis of LSA with \iid\ and Markov noise dynamics, based on the stochastic expansion for LSA \eqref{eq:lsa} introduced in \citet{aguech2000perturbation}. In this framework, we derive the finite time bounds for the $p$-th moment of
$\{\norm{\theta_n-\thetas}\}_{n\in\nset}$ and
$\{\norm{\btheta_{n}-\thetas}\}_{n\in\nset}$. The obtained bounds for PR-averaged LSA iterates are sharp in a sense that the leading term of these bounds coincides with the one of the central limit theorem. Moreover, as a corollary, for a fixed tolerance parameter $\delta \in \ooint{0,1}$ and the number of iterations $n$, we provide high-probability bounds on the error of LSA-PR iterates. In the \iid\ setup, our results extend and improve those obtained in \cite{mou2020linear} by providing a better dependence \wrt\ on the problem dimension $d$ and on the moment order $p$ for the remainder terms. The improvement w.r.t. dimension dependence comes through a logarithmic dependence of the maximal allowed step size upon the dimension $d$. In the case of Markovian noise, to the best of our knowledge, the results concerning the $p$-th moment of the LSA-PR error are novel. Moreover, in the Markovian setup the remainder terms of our bounds scale with the ratio $n/\taumix$, which can not be improved in general even for the case of MSE bounds (see e.g. \cite[Theorem~2]{mou2021optimal}).
\par
The paper is organized as follows. In \Cref{sec:stochastic-expansion} we introduce the decomposition of the error, which is key to our proof (see \citet{aguech2000perturbation}), and formulate our main assumptions. In \Cref{sec:independent_case} we present our results for the independent case. In \Cref{sec:markov} we extend our results when $\sequence{Z}[n][\nsets]$ is a uniformly geometrically ergodic Markov chain. The proofs are postponed to the appendix.
For reader's convenience the notations and key constants appearing in the text are summarized in \Cref{appendix:constants}.

\section{Stochastic expansions for LSA and LSA-PR}
\label{sec:stochastic-expansion}
As an introduction, we present tools and some preliminary results relevant to our analysis of LSA and
LSA-PR under both \iid\ and Markovian noise dynamics. Using the definition \eqref{eq:lsa} and some elementary algebra, we obtain
$$
\textstyle
\theta_{n} - \thetas = (\Id - \alpha \funcA{Z_n})(\theta_{n-1} - \thetas) - \alpha \funcnoise{\State_{n}}\eqsp,
$$
where we have set
\begin{equation}
\label{eq:def_center_version_and_noise}
\textstyle
\funcnoise{z} =  \zmfuncA{z} \thetas - \zmfuncb{z}\eqsp, \quad \zmfuncA{z}  = \funcA{z} - \bA \eqsp, \quad   \zmfuncb{z} = \funcb{z} - \barb \eqsp \eqsp.
\end{equation}
We expand the recurrence above using the notation $\ProdBa_{1:n}$ for the product of random matrices
\begin{equation} \label{eq:definition-Phi} \textstyle
\ProdBa_{m:n}  = \prod_{i=m}^n (\Id - \alpha \funcA{Z_i} ) \eqsp, \quad m,n \in\nsets, \quad m \leq n \eqsp,
\end{equation}
with the convention, $\ProdBa_{m:n}=\Id$ for $m > n$. We arrive at the decomposition of the LSA error:
$$
\theta_{n} - \thetas = \utheta_{n} + \vtheta_{n}\eqsp,
$$
where we have defined
\begin{equation}
\label{eq:LSA_recursion_expanded}
\textstyle
\utheta_{n} =  \ProdBa_{1:n} \{ \theta_0 - \thetas \} \eqsp, \quad \vtheta_{n} = - \alpha \sum_{j=1}^n \ProdBa_{j+1:n} \funcnoise{Z_j}\eqsp.
\end{equation}
Here $\utheta_{n}$ is a transient term (reflecting the forgetting of the initial condition) and $\vtheta_{n}$ is a fluctuation term (reflecting misadjustement noise). In both \iid\ and Markov noise dynamics, we proceed by treating the $\utheta_{n}$ and $\vtheta_{n}$ terms separately.

\paragraph{Bounding the transient term.} We first bound the $p$-th moments of $\{\norm{\utheta_{n}}\}_{n \in \nsets}$ by proving that the sequence of random matrices $\{ \funcA{Z_i} \}_{i \in \nsets}$ is \emph{exponentially stable} (see \citet{guo1995exponential,ljung2002recursive}). Formally, this means that for some $q \geq 1$, there exist constants $\mathsf{a}_q, \Const{q} > 0$ and $\alpha_{\infty,q} <\infty$, such that for any step size $\alpha \leq \alpha_{\infty,q}$, $m,n \in \nsets$, $m \leq n$,
\begin{equation}
\label{eq:L_V_q-exponential-stability} \txts
\PE[ \| \ProdBa_{m:n}\|^q ] \leq \Const{q} \exp\left( - \mathsf{a}_q \alpha (n - m)\right) \eqsp.
\end{equation}
Exponential stability is established in \Cref{fact:exponential-stability-product} for the \iid~setting and in \Cref{prop:products_of_matrices_UGE} for the Markovian setting. Intuitively, exponential stability means that the $q$-th moment of the product of random matrices $\ProdBa_{m:n}$ behaves similarly to the product of \emph{deterministic} matrices $(\Id - \alpha \bA)^{n-m}$ under the assumption that the matrix $-\bA$ is Hurwitz, i.e., for each eigenvalue $\lambda$ of $\bA$ we have $\realpart(\lambda) > 0$. To handle the product $(\Id - \alpha \bA)^{n-m}$, we use the Lyapunov contraction property: if the matrix $-\bA$ is Hurwitz, there exists  a symmetric positive definite matrix $Q$ such that $\Id - \alpha \bA$ is a strict contraction in $\normop{\cdot}[Q]$ (see \Cref{appendix:constants} for the relevant definitions). Precisely, the following result holds:
\begin{proposition}[\protect{\cite[Proposition 1]{durmus2021tight}}]
\label{fact:Hurwitzstability}
Assume that $-\bA$ is Hurwitz. Then there exists a unique symmetric positive definite matrix $Q$ satisfying the Lyapunov equation 
$\bA^\top Q + Q \bA =  \Id$. In addition, setting
\begin{equation}
\label{eq: kappa_def}
a = \normop{Q}^{-1}/2\eqsp, \quad
\text{and} \quad \alpha_\infty = (1/2) \normop{\bA}[Q]^{-2} \normop{Q}^{-1} \wedge \normop{Q} \eqsp,
\end{equation}
it holds for any $\alpha \in [0, \alpha_{\infty}]$ that $\normop{\Id - \alpha \bA}[Q]^2 \leq 1 - a \alpha$, and $\alpha a \leq 1/2$.
\end{proposition}
\paragraph{Bounding the fluctuation term.}
For the fluctuation term $\vtheta_{n}$ we use the perturbation expansion technique formalized in \cite{aguech2000perturbation}. We do not exploit the full power of this decomposition here and use only first- and second-order expansions. Higher order expansions are presented in \cite{aguech2000perturbation}. We emphasize that the considered error expansion remains the same in both the \iid\ and Markovian noise dynamics, so that $p$-th moment bounds on the error norm can be obtained in both cases. In the following, we briefly sketch how the decomposition of $\vtheta_{n}$ is constructed.
 By definition \eqref{eq:LSA_recursion_expanded} of $\vtheta_{n}$, it satisfies the recurrence
\begin{equation}
\label{eq:main_rec_fluct}
\textstyle
\vtheta_{n} = ( \Id - \alpha \funcA{\State_{n}} )\vtheta_{n-1} - \alpha \funcnoise{\State_{n}}\eqsp.
\end{equation}
Using the definition \eqref{eq:def_center_version_and_noise} of $\zmfuncAw(\cdot)$ and an induction argument, it is easy to verify that the following decomposition holds for any $n\in\nset$:
\begin{equation}
\label{eq:decomp_fluctuation}
\vtheta_{n} = \Jnalpha{n}{0}+ \Hnalpha{n}{0} \eqsp,
\end{equation}
where the latter terms are defined by the following pair of recursions
\begin{align}
\label{eq:jn0_main}
&\Jnalpha{n}{0} =\left(\Id - \alpha \bA\right) \Jnalpha{n-1}{0} - \alpha \funcnoise{\State_{n}}\eqsp, && \Jnalpha{0}{0}=0\eqsp, \\[.1cm]
\label{eq:hn0_main}
&\Hnalpha{n}{0} =\left( \Id - \alpha \funcA{\State_{n}} \right) \Hnalpha{n-1}{0} - \alpha \zmfuncA{\State_{n}} \Jnalpha{n-1}{0}\eqsp, && \Hnalpha{0}{0}=0\eqsp.
\end{align}
The recursion for $\Jnalpha{n}{0}$ is obtained by replacing $\funcA{\State_n}$ by its mean $\bA$ in \eqref{eq:main_rec_fluct} (cf. \eqref{eq:jn0_main} and \eqref{eq:main_rec_fluct}), and $\Hnalpha{n}{0}$ is remainder term. As a result $\Jnalpha{n}{0}$ becomes a linear statistic in $\{\funcnoise{\State_n}\}_{n \in \nsets}$ and is relatively easy to analyze with common concentration tools for \iid\ variables or Markov chains, see e.g. \cite[Section~2]{vershynin:2018} and \cite[Chapter~23]{douc:moulines:priouret:soulier:2018}. Interestingly, it can be shown that the term $\Jnalpha{n}{0}$ is the leading term with respect to $\alpha$ in the expansion \eqref{eq:decomp_fluctuation}. Moreover, the covariance matrix of $\Jnalpha{n}{0}$ is closely related to the asymptotic covariance matrix appearing in the central limit theorems for LSA with decreasing step size, see \cite{durmus2021tight} for further discussion on this topic.
We can further expand the decomposition \eqref{eq:decomp_fluctuation} since the recursion for the remainder term \eqref{eq:hn0_main} resembles the one in \eqref{eq:main_rec_fluct}. Hence, we can apply the same decomposition for $\Hnalpha{n}{0}$, and obtain
\begin{equation}
\label{eq:decomposition_H_n_0}
\textstyle \Hnalpha{n}{0} = \Jnalpha{n}{1} + \Hnalpha{n}{1} \eqsp,
\end{equation}
where we have set
\begin{equation}
\label{eq:jn_allexpansion_main}
\begin{aligned}
&\Jnalpha{n}{1} =\left(\Id - \alpha \bA\right) \Jnalpha{n-1}{1} -  \alpha \zmfuncA{\State_{n}} \Jnalpha{n-1}{0}\eqsp,
&& \Jnalpha{0}{1}=0  \eqsp, \\
& \Hnalpha{n}{1} =\left( \Id - \alpha \funcA{\State_{n}} \right) \Hnalpha{n-1}{1} - \alpha \zmfuncA{\State_{n}} \Jnalpha{n-1}{1} \eqsp, && \Hnalpha{0}{1}=0 \eqsp.
\end{aligned}
\end{equation}
Representation \eqref{eq:jn_allexpansion_main} can be elaborated further to decompose $\Hnalpha{n}{1}$, but this is not needed in this work. Combining \eqref{eq:decomp_fluctuation} and \eqref{eq:decomposition_H_n_0}, we obtain the decomposition which is the cornerstone of our analysis:
\begin{equation}
\label{eq:error_decomposition_LSA}
\textstyle \theta_{n} - \thetas =  \utheta_{n} + \Jnalpha{n}{0} + \Jnalpha{n}{1} + \Hnalpha{n}{1}\eqsp,
\end{equation}
where $\Jnalpha{n}{0}$, $\Jnalpha{n}{1}$, and $\Hnalpha{n}{1}$ are defined in \eqref{eq:jn0_main} and \eqref{eq:jn_allexpansion_main}, respectively.
Following the arguments in \citet{durmus2021tight}, this decomposition can be used to obtain sharp bounds on the $p$-th moment of the final LSA iterate  $\theta_n$.

\paragraph{Stochastic expansion for LSA-PR.} Our analysis of PR-LSA is based on  another useful error representation. Using \eqref{eq:lsa} and the definition of the noise term $\funcnoise{\cdot}$ in \eqref{eq:definition-Phi}, we get that
\begin{align}
\label{eq:pr_err_decompose}
\bA\left(\prtheta_{n} -\thetalim\right) & \textstyle =  \{\alpha (n - n_0) \}^{-1} ( \theta_{n_0}-\theta_{n} ) - (n-n_0)^{-1} \sum_{t=n_0}^{n-1} e\left(\theta_{t}, Z_{t+1} \right) \eqsp, \\
\label{eq:pr_e_definition}
e(\theta,z) & = \zmfuncA{z} \theta- \zmfuncb{z} = \funcnoise{z} + \zmfuncA{z} (\theta - \thetas) \eqsp.
\end{align}
We will establish in the sequel when bounding the error of the last LSA iterate, the term $\{\alpha n \}^{-1} (\theta_{n_0}-\theta_{n})$ is small compared to the second term in \eqref{eq:pr_err_decompose} with a suitably chosen $n_0$. Moreover, in the \iid~case it can be shown that $\{e\left(\theta_{t}, Z_{t+1} \right)\}_{t=0}^{n}$ are martingale increments, and the MSE bound of LSA-PR will follow directly from this observation (see \Cref{prop:mse_iid_with_burn_in}). This property of $\{e\left(\theta_{t}, Z_{t+1} \right)\}_{t=0}^{n}$ was also used in \cite{mou2020linear}.
\par 
To proceed with the $p$-th moment bounds of the LSA-PR method, we need to combine the extensions \eqref{eq:pr_err_decompose} and \eqref{eq:error_decomposition_LSA}. That is, we write
\begin{equation}
\label{eq:decompo_e_theta_z}
\textstyle \sum_{t=n_0}^{n-1} e\left(\theta_{t}, Z_{t+1} \right)= \Etr_{n} + \Efl_{n}\eqsp,
\end{equation}
where we have set
\begin{align}
\label{eq:def:Etr}
\Etr_{n}  &= \textstyle \sum_{t=n_0}^{n-1} \zmfuncA{\State_{t+1}} \ProdBa_{1:t} \{ \theta_0 - \thetas \} \eqsp,\\
\label{eq:def:Efl}
\Efl_{n} &= \textstyle \sum_{t=n_0}^{n-1} \funcnoise{\State_{t+1}} + \sum_{\ell=0}^{1}\sum_{t=n_0}^{n-1} \zmfuncA{\State_{t+1}} \Jnalpha{t}{\ell} + \sum_{t=n_0}^{n-1} \zmfuncA{\State_{t+1}} \Hnalpha{t}{1} \eqsp.
\end{align}
Based on the decompositions \eqref{eq:pr_err_decompose} and \eqref{eq:decompo_e_theta_z}, our analysis of the PR recursion consists in bounding the terms $\{\alpha (n - n_0) \}^{-1} ( \theta_{n_0}-\theta_{n} )$, $\Etr_{n}$ and $\Efl_{n}$ separately. For the first one, we use the bounds derived on the $p$-th moment of non-averaged LSA iterates $\theta_k - \thetas$. For the second one, we use the exponential stability for the products of random matrices $\{\ProdBa_{1:t}\}$ (see \eqref{eq:L_V_q-exponential-stability}). Finally, the fluctuation term $\Efl_{n}$  is dealt with the conditions we impose on the sequence $\sequence{Z}[n][\nsets]$.
\par 
We suppose from now on that the sample size $n$ is even, and fix the size of burn-in period $n_0 = n/2$. Thus we suppress the dependence upon $n_0$ in \eqref{eq:lsa} and use simplified notation 
\begin{equation}
\label{eq:bar_theta_n_fixed}
\prtheta_{n} = (2/n) \sum_{k=n/2}^{n-1} \theta_k\eqsp.
\end{equation}

\paragraph{Assumptions.} Throughout this paper (both in case of \iid\ and Markovian noise dynamics) we impose the following assumption regarding $z \mapsto \zmfuncAw(z)$ and $\bA$:
\begin{assum}
\label{assum:A-b}
$\bConst{A} = \sup_{z \in \msz} \normop{\funcA{z}} \vee \sup_{z \in \msz} \normop{\zmfuncA{z}} < \infty$ and  the matrix $-\bA$ is Hurwitz.
\end{assum}

In particular, the condition that $-\bA$ is Hurwitz implies that the linear system $\bA \theta = \barb$ has a unique solution $\thetalim$.

We further require the following
assumptions on the noise term $\funcnoise{z}$ and the stationary distribution $\invariantQ$ of the sequence $\sequence{Z}[n][\nsets]$:
\begin{assum}
\label{assum:noise-level}
$\int_{\Zset}\funcA{z}\rmd \pi(z) = \bA$ and $\int_{\Zset}\funcb{z}\rmd \pi(z) = \barb$. Moreover, $\supconsteps = \sup_{z \in \msz}\normop{\funcnoise{z}} < \plusinfty$.
\end{assum}
Our bounds in the \iid~noise case will also depend upon the covariance matrix of $\funcnoise{z}$, that is,
\begin{equation}
\label{eq:def_noise_cov}
\textstyle \noisecov = \int_{\Zset} \funcnoise{z}\funcnoise{z}^\top \rmd \pi(z) \eqsp.
\end{equation}
Assumption \Cref{assum:noise-level} can be generalized in certain directions. In particular, in \Cref{appendix:iid_subgaus} we provide the counterparts of the results of \Cref{sec:independent_case} under the assumption that the sequence $\{\funcnoise{\State_t}\}_{t \in \nsets}$ are \iid~sub-Gaussian random variables. The case of unbounded noise in the Markovian setting is much more technical and is left as a direction for future work.


%% file: independent_case.tex
In addition to \Cref{assum:A-b} and \Cref{assum:noise-level}, we consider in this section the following assumption:
\begin{assumID}
\label{assum:IID}
$\sequence{Z}[k][\nset]$ is a sequence of \iid\ random variables defined on a probability space $(\Omega,\mcf,\PP)$ with distribution $\pi$.
\end{assumID}
\par
The first result of this section provides $p$-th moment bounds for LSA-PR under \Cref{assum:IID}. Using the notations of \Cref{fact:Hurwitzstability} and assuming \Cref{assum:A-b}, we define for $q \geq 2$
\begin{align}
\label{eq:def_qcond_b_Q}
&  \qcond = \lambda_{\sf max}( Q )/\lambda_{\sf min}( Q )  \eqsp, \quad  b_{Q} = 2 \sqrt{\qcond} \bConst{A} \eqsp, \quad \\
  \label{eq:def_alpha_p_infty}
& \alpha_{q,\infty} = \alpha_{\infty} \wedge \smallAconst/q \eqsp, \quad \smallAconst = a/\{2b_Q^2\}\eqsp.
\end{align}
The quantity $\alpha_{q,\infty}$ defined above is a threshold on the step size that guarantees exponential stability for the $q$-th moment of the product of $\ProdBa_{1:n}$. Below we give our exponential stability result for the $p$-th moment of the product of $\ProdBa_{1:n}$.
\begin{proposition}[\protect{\citet[Corollary~1]{durmus2021tight}}]
\label{fact:exponential-stability-product}
Assume \Cref{assum:A-b} and \Cref{assum:IID}. Then, for any  $2 \leq p \leq q$, $\alpha \in \ocint{0, \alpha_{\infty}}$ and $n \in \nsets$, it holds
\begin{equation}
\label{eq:concentration iid}
 \PE^{1/p}\left[ \normop{\ProdBa_{1:n}}^{p} \right]
\leq \sqrt{\qcond} d^{1/q} (1 - a \alpha + (q-1) b_Q^2 \alpha^2)^{n/2}\eqsp.
\end{equation}
Moreover, for $\alpha \in \ocint{0, \alpha_{q, \infty}}$, it holds
\begin{equation}
\label{eq:concentration_iid_bounded_stepsize}
 \PE^{1/p}\left[ \normop{\ProdBa_{1:n}}^{p} \right]
\leq \sqrt{\qcond} d^{1/q} (1 - a \alpha / 2)^{n/2}\eqsp.
\end{equation}
\end{proposition}
\Cref{fact:exponential-stability-product} implies  that $\sup_{n\in\nsets} \PE[\normop{\ProdBa_{1:n}}^{p}]< \plusinfty$ for any $\alpha \in \ocint{0,\alpha_{q,\infty}}$ and $2 \leq p \leq q$. This condition connecting the choice of the step size $\alpha$ with $p$ and $q$ is unavoidable; see \citet[Example~1]{durmus2021tight}.
Roughly speaking, the condition on the step size $\alpha \in \ocint{0,\alpha_{q, \infty}}$ in \Cref{fact:exponential-stability-product} requires scaling the step size $\alpha$ with $1/q$. \cite[Theorem~9]{srikant:1tsbounds:2019} reports the same kind of dependence.
\begin{remark}
\label{rem:choice_stepsize_dimension_undep}
The flexibility achieved by using $q \geq p$ allows us to control the dependence on dimension $d$ independently of the choice of the $p$-th moment. In particular, we can choose $q =  c \log{d} $ with a suitable constant $c > 0$ and obtain $\PE^{1/p}[ \normop{\ProdBa_{1:n}}^{p}] \leq \sqrt{\qcond} \rme^{c}$ for $2 \leq p \leq c \log{d}$. This comes at the expense of taking a maximum step size $\alpha_{c\log(d),\infty}$ which scales with $1 / \log{d}$.
\end{remark}

We can now state the main result of this section, which is a $p$-th moment bound for $\PE ^{1/p}\left[\norm{\bA\left(\prtheta_{n}-\thetas\right)}^{p}\right]$. We write $\lesssimd$ for inequality up to a constant that depends on $\qcond$, $a$, $\bConst{A}$, and polylogarithmic factors in $d$.
\begin{theorem}
\label{th:theo_1_iid}
Assume \Cref{assum:A-b}, \Cref{assum:noise-level}, \Cref{assum:IID}. Then, for any $p \geq 2$, even $n \geq 2$, step size
\begin{equation}
\label{eq:step_size_pth_moment_refined}
\alpha(n,d,p) = \left( \alpha_{\infty} \wedge \smallAconst/\{1+\log{d}\} \right)(p n^{1/2})^{-1} \eqsp,
\end{equation}
and an initial parameter $\thetainit \in \rset^d$, it holds that
\begin{multline}
\label{eq:bound_p_th_moment_optimized_iid}
\PE^{1/p}\left[\norm{\bA\left(\prtheta_{n}-\thetas\right)}^{p}\right] \lesssimd \frac{\{\trace{\noisecov}\}^{1/2} p^{1/2}}{n^{1/2}} + \supconsteps \left(\frac{p}{n^{3/4}} + \frac{p^{2}}{n}\right) \\
 + p \norm{\thetainit - \thetas} \exp\left\{ - \frac{( \alpha_{\infty} \wedge \smallAconst) \sqrt{n}}{8 p (1+\log{d})} \right\}\eqsp.
\end{multline}
\end{theorem}
A generalization of \Cref{th:theo_1_iid} for an arbitrary step size $\alpha \in (0,\alpha_{q,\infty})$ is given below in \Cref{th:LSA_PR_error}, and its reformulation into high probability bounds is given in \Cref{cor:hp_bound_pr_iid}. The dependence of the step size $\alpha(n,d,p)$ on the sample size $n$ can be illustrated by optimizing the bounds provided by \Cref{th:LSA_PR_error} (see \eqref{eq:delta_transient_fluctuation_terms_simplified}). For completeness, we provide versions of \Cref{th:LSA_PR_error} and \Cref{cor:hp_bound_pr_iid} with exact constants in \Cref{sec:proof:LSA_PR_error_iid_consts}.
\par
Compared to \citet[Theorem~3]{mou2020linear}, \Cref{th:theo_1_iid} has a similar leading term, but with a different numerical factor. Instead of $\trace{\noisecov}$, the factor in \citet[Theorem~3]{mou2020linear} is the covariance matrix associated with the $\sequence{\theta}[k][\nset]$ when viewed as a Markov chain: $\noisecov^{(\alpha)} = n^{-1} \lim_{n \to \plusinfty} \PE [\sum_{i=1}^n (\theta_i - \thetas) (\theta_i-\thetas)^{\transpose}]$. \citet[Proposition 6]{durmus2021tight} shows that $\norm{\noisecov^{(\alpha)} - \noisecov} = \bigO(\alpha)$ with $\alpha \to 0$. It is worth noting that the conclusions of \citet[Theorem~3]{mou2020linear} regarding the choice of the optimal step size and the resulting high probability bounds differ slightly from ours since their optimization omits the dependence on the step size $\alpha$ in the term $\trace(\noisecov^{(\alpha)})$. \Cref{th:theo_1_iid} accounts for this additional factor in the optimization and leads to an optimal choice for $\alpha$ of order $n^{-1/2}$ and a residual term in $n^{-3/4}$, while the optimal choice of $\alpha$ from \citet[Theorem~3]{mou2020linear} has order $n^{-1/3}$ and leads to a residual term in $n^{-2/3}$. Moreover, \Cref{th:theo_1_iid} improves the scaling of the residual term \wrt\ $p$, and, unlike \citet[Theorem~3]{mou2020linear}, shows exponential forgetting of the initial condition.
Finally, an inspection of the proof of \citet[Theorem~3]{mou2020linear} shows that it relies heavily on results on additive functionals of Markov chains developed in \citet{joulin2010curvature}, while the main result in our derivation is a Rosenthal inequality for martingales. Applying the results from \citet{joulin2010curvature} requires log-Sobolev conditions for the noise distribution $(\funcnoise{\State_{n}})_{n \in \nset}$, which are very restrictive. \citet[Theorem~3]{mou2020linear} does not apply to the general framework considered here.
\par
\paragraph{Bounds for the non-averaged LSA iterates}
To bound $\PE^{1/p}\left[\norm{\bA\left(\prtheta_{n}-\thetas\right)}^{p}\right]$ in \Cref{th:theo_1_iid}, we first need to control the $p$-th moment of the last LSA iterate error $\sequenceD{\theta_n - \thetas}[n \in \nset]$. To this end, we use the decomposition \eqref{eq:decomp_fluctuation} and rely on the following $p$-th moment bounds for the sequence $\sequenceD{\Jnalpha{n}{0}}[n \in \nset]$.
\begin{proposition}
\label{prop:J_n_0_bound_iid}
Assume \Cref{assum:A-b}, \Cref{assum:noise-level}, and \Cref{assum:IID}. Then, for any $\alpha \in \ocint{0,\alpha_{\infty}}$, $p \geq 2$, and $n \in \nset$, it holds
\begin{equation}
\label{eq:J_n_0_bound_iid}
\PE^{1/p}\left[\norm{\Jnalpha{n}{0}}^{p}\right] \leq \ConstD_{1}\sqrt{\alpha a  p} \supconsteps \eqsp, \text{ where } \ConstD_1 =
\sqrt{2 \qcond} /a\eqsp.
\end{equation}
\end{proposition}
The proof is deferred to \Cref{sec:proof_j_n_0_iid}. The argument goes as follows.
Expanding the recurrence \eqref{eq:jn0_main}, we represent
\begin{equation}
\label{eq:J_n_0_repr_iid}
\textstyle
\Jnalpha{n}{0} = -\alpha\sum_{k=1}^{n}(\Id-\alpha \bA)^{n-k}\funnoisew(\State_k)\eqsp.
\end{equation}
Now the bound \eqref{eq:J_n_0_bound_iid} follows from a Hoeffding-type bound for sums of independent random vectors (see \cite[Theorem~3.1]{pinelis_1994}) in combination with \Cref{fact:Hurwitzstability}.

The constant $\ConstD_1$ is instance-dependent, since it depends on $a,\qcond$. Nevertheless, the product $\ConstD_1 \supconsteps$ is scale-invariant, that is, if we multiply $\bA$ by a positive constant $M$, both $a$ and $\supconsteps$ scales in the same way, leaving $\ConstD_1 \supconsteps$ unchanged. This property of scale invariance holds for all constants in the bounds that appear in the following statements. We emphasize that $\Jnalpha{n}{0}$ is the leading (with respect to the step size $\alpha$) term in the error decomposition \eqref{eq:error_decomposition_LSA}. Indeed, \eqref{eq:J_n_0_bound_iid} and the stability result (\Cref{fact:exponential-stability-product}) are sufficient to obtain a rough bound $\PE ^{1/p}[\norm{\Hnalpha{n}{0}}^{p}] \leq C\sqrt{\alpha}$ for a constant $C \geq 0$. Combining these results gives the following $p$-th moment bound for the LSA error $\norm{\theta_n - \thetas}$:
\begin{proposition}
\label{cor:LSA_err_bound_iid}
Assume \Cref{assum:A-b}, \Cref{assum:noise-level}, and \Cref{assum:IID}. Then, for any  $p,q \in\nset$, $2 \leq p \leq q$, $\alpha \in \ocint{0, \alpha_{q,\infty}}$, $n \in \nset$, and $\thetainit \in \rset^d$ it holds
\begin{align}
\label{eq:n_step_norm_propery}
\PE^{1/p}\left[\norm{\theta_n - \thetas}^{p}\right]
\leq d^{1/q} \qcond^{1/2} \left(1 - \alpha a/4\right)^{n} \norm{\thetainit - \thetas} + d^{1/q} \ConstD_{2} \sqrt{\alpha a  p} \supconsteps \eqsp,
\end{align}
where  $\ConstD_{2}$ is given in \eqref{eq:constants_D_definition}.
\end{proposition}
The proof is given in \Cref{sec:proof_h_n_0_iid}. It is based on the expansion \eqref{eq:decomp_fluctuation}, the stability result of \Cref{fact:exponential-stability-product}, and bounds on $\Jnalpha{n}{0}$ obtained in \Cref{prop:J_n_0_bound_iid}. We control the moments of $\Hnalpha{n}{0}$ with Holder's inequality and a bound for $\Jnalpha{n}{0}$. The bound thus obtained for $\Hnalpha{n}{0}$ is not optimal: the dependence of $\Hnalpha{n}{0}$ in $\alpha$ is improved below. However, this preliminary bound is sufficient to obtain the $p$-th moment bound for $\norm{\theta_{n} - \thetas}$, which is tight with respect to the dependence on the step size $\alpha$. Note that at the expense of the logarithmic dependence of step size $\alpha$ on the dimension, one can get rid of the dependence on dimension $d$ in \eqref{eq:n_step_norm_propery}; see \Cref{rem:choice_stepsize_dimension_undep}.
\par
\paragraph{MSE bound for LSA-PR}
We preface the proof of \Cref{th:theo_1_iid} by a separate bound on the mean square error of $n$-steps LSA-PR $\prtheta_{n}-\thetas$. While this result could be a consequence of \Cref{th:theo_1_iid}, we present here a separate and simpler derivation which leads to sharper bounds. Our strategy consists in using the decomposition \eqref{eq:pr_err_decompose} and the fact that $\{e(\theta_t,\State_{t+1})\}_{t=0}^{n-1}$ is a martingale increment sequence. Thus, we get from \eqref{eq:pr_err_decompose} that
\begin{equation}
\label{eq:theta_pr_mse_iid}
(n/2) \PE\left[\norm{\bA\left(\prtheta_{n}-\thetas\right)}^{2}\right] \leq \underbrace{ 4n^{-1}\sum\nolimits_{t=n/2}^{n-1}\PE\bigl[\norm{e(\theta_{t},\State_{t+1})}^{2}\bigr]}_{T_1} + \underbrace{4 (\alpha^2 n)^{-1}\PE[\norm{\theta_{n/2}-\theta_{n}}^2]}_{T_2}\eqsp.
\end{equation}
Since by definition $e(\theta_{t},\State_{t+1}) = \funcnoise{\State_{t+1}} + \zmfuncA{\State_{t+1}} (\theta_{t} - \thetas)$, the term $T_1$ contains the variance term $\PE [\norm{\funcnoise{\State_{t+1}}}^{2}] = \trace{\noisecov}$, which is the leading term when the step size $\alpha$ is small enough. This fact follows from \Cref{cor:LSA_err_bound_iid}, which implies that $\PE [\norm{\theta_{t}-\thetas}^2] \leq C\alpha$ up to the exponentially decreasing transient terms. Next result is obtained by deriving quantitative bounds for $T_1$ and $T_2$.

\begin{proposition}
\label{prop:mse_iid_with_burn_in}
Assume \Cref{assum:A-b}, \Cref{assum:noise-level}, and \Cref{assum:IID}. Then, for any even $n \geq 2$, $\alpha \in (0, \alpha_{\infty} \wedge \smallAconst/\{2 + 2\log{d}\})$,  $\thetainit \in \rset^d$,  it holds that
\[
(n/2) \PE\left[\norm{\bA\left(\prtheta_{n}-\thetas\right)}^{2}\right]
 \leq 4 \trace{\noisecov} + \Deltafl_{n,\alpha} \supconsteps + \rme^{-\alpha a n/4} \Deltatr_{n,\alpha} \norm{\thetainit - \thetas}^{2} \eqsp,
 \]
 where $\Deltafl_{n,\alpha}$ and $\Deltatr_{n,\alpha}$ are given in \eqref{eq:delta_transient_fluctuation_terms}.
\end{proposition}
The complete proof is postponed to \Cref{sec:proof:prop:mse_iid_with_burn_in}. In the previous statement, $\Deltatr_{n,\alpha} $ and $\Deltafl_{n,\alpha}$ correspond to the \emph{transient} and \emph{fluctuation} components of the LSA error. The initial condition's exponential forgetting is represented by the transient term, and the fluctuations of the non-averaged LSA iterates $\theta_n$ around $\thetas$ are captured by $\Deltafl_{n,\alpha}$.
It is worth noting that
\begin{equation}
\label{eq:delta_transient_fluctuation_terms_simplified}
\begin{aligned}
\Deltatr_{n,\alpha} \lesssimd (\alpha a n)^{-1} (1 + \alpha^{-1})
\eqsp, \qquad \Deltafl_{n,\alpha} \lesssimd (\alpha n)^{-1} + \alpha\eqsp.
\end{aligned}
\end{equation}
The above bounds can be simplified for a given choice of $\alpha$ as a function of sample size $n$. Optimizing the fluctuation error $\Deltafl$ in \eqref{eq:delta_transient_fluctuation_terms_simplified} for a fixed sample size $n$ suggests that $\alpha$ should scale with $n$ as $n^{-1/2}$. Then, choosing
\[
\alpha(n,d) = \left(\alpha_{\infty} \wedge \smallAconst/\{2 + 2\log{d}\} \right) n^{-1/2}\eqsp,
\]
we obtain from \Cref{prop:mse_iid_with_burn_in} the MSE bound
\begin{equation}
\label{eq:bound_prop_1_optimized_n_refined}
\PE\left[\norm{\bA\left(\prtheta_{n}-\thetas\right)}^{2}\right] \lesssimd \frac{\trace{\noisecov}}{n} + \frac{\supconsteps^{2}}{n^{3/2}} + \norm{\thetainit - \thetas}^2 \exp\left\{ - \frac{( \alpha_{\infty} \wedge \smallAconst) \sqrt{n}}{8(1+\log{d})} \right\}\eqsp.
\end{equation}

Note that the bound \eqref{eq:bound_prop_1_optimized_n_refined} has the same (optimal) leading term $n^{-1} \trace{\noisecov}$ as in \citet[Theorem~1]{mou2021optimal}, improving the dependence on sample size $n$ in the remainder term. To compare with \citet[Theorem~1]{mou2021optimal}, we assume that $\supconsteps \approx \sqrt{d}$. Then \eqref{eq:bound_prop_1_optimized_n_refined} yields a remainder term of order $d/n^{3/2}$, while the second-order term in \citet[Theorem~1]{mou2021optimal} scales as $(d/n)^{4/3}$.
\par
\paragraph{Outline of the proof of \Cref{th:theo_1_iid}}
To obtain \Cref{prop:mse_iid_with_burn_in}, we only used the expansion \eqref{eq:decomp_fluctuation}. But this decomposition is not sufficient to show \emph{scale separation} with respect to the step size $\alpha$ between $\sequenceD{\Jnalpha{n}{0}}[n\in\nset]$ and $\sequenceD{\Hnalpha{n}{0}}[n\in\nset]$. More precisely, in the proof of \Cref{prop:mse_iid_with_burn_in}, we only show that $ \sup_{n\in\nset} \PE ^{1/p}[\normLine{H_n^{(0)}}^{p}] \leq C \alpha^{1/2}$ for $\alpha$ small enough and a constant $C\geq 0$. To refine this bound, we use the expansion \eqref{eq:decomposition_H_n_0} to obtain that $ \sup_{n\in\nset} \PE^{1/p}[\normLine{H_n^{(0)}}^{p}] \leq C \alpha$, if $\alpha$ is small enough, for a constant $C\geq 0$. We formalize this result in the following proposition.
\begin{proposition}
\label{prop:J_n_1_2_bound_iid} Assume \Cref{assum:A-b}, \Cref{assum:noise-level}, and \Cref{assum:IID}. Then, for any $\alpha \in \ocint{0,\alpha_{\infty}}$, $p \geq 2$, and $n \in \nset$, it holds
\begin{equation}
\label{eq:J_n_1_bound_iid_bounded}
\PE^{1/p}\left[\norm{\Jnalpha{n}{1}}^{p}\right] \leq \ConstD_3 \alpha a  p^{3/2} \supconsteps \eqsp,
\end{equation}
where $\ConstD_3$ is defined in \eqref{eq:def:constD_3_4}.
Moreover, for any $2 \leq p \leq q$ and $\alpha \in \ocint{0, \alpha_{q,\infty}}$, $n \in \nset$,
\begin{equation}
\label{eq:H_n_1_bound_iid_bounded}
\PE^{1/p}\left[\norm{\Hnalpha{n}{1}}^{p}\right] \leq \ConstD_{4} \alpha a p^{3/2} d^{1/q} \supconsteps \eqsp,
\end{equation}
where $\ConstD_4$ is defined in \eqref{eq:def:constD_3_4}.
\end{proposition}
The proof is provided in \Cref{sec:proof_j_n_1_iid}. Using \Cref{prop:J_n_1_2_bound_iid}, we obtain $p$-th moment error bounds for LSA-PR.

\begin{theorem}
\label{th:LSA_PR_error}
Assume \Cref{assum:A-b}, \Cref{assum:noise-level}, \Cref{assum:IID}. Then, for any $p \geq 2$, even $n \geq 2$, $\alpha \in (0, \alpha_{p(1+\log{d}), \infty})$, $\thetainit \in \rset^d$, it holds that
\begin{equation}
\label{eq:p_th_moment_LSA_bound}
(n/2)^{1/2}\PE^{1/p}\left[\norm{\bA\left(\prtheta_{n}-\thetas\right)}^{p}\right]
\leq  \bConst{\sf{Rm}, 1} \{\trace{\noisecov}\}^{1/2} p^{1/2} + \Deltafl_{n,p,\alpha} \supconsteps + \rme^{- \alpha a n/8}
 \Deltatr_{n,p,\alpha}\norm{\thetainit - \thetas}\eqsp,
\end{equation}
where $\bConst{\sf{Rm}, i}$, $i=1,2$,  are defined in \Cref{appendix:constants} and $\Deltatr_{n,p,\alpha}$, $\Deltafl_{n,p,\alpha}$ are given in \eqref{eq:delta_fl_p_th_moment}.
\end{theorem}
The proof is postponed to \Cref{sec:proof:th:LSA_PR_error}. Similarly to the proof \Cref{th:LSA_PR_error}, we rely on the decomposition \eqref{eq:decompo_e_theta_z} but use \Cref{prop:J_n_1_2_bound_iid} to bound $p$-th moments of the fluctuation term \eqref{eq:def:Efl}.
Here again the leading term is $\sum_{t=n/2}^n \funcnoise{\State_{t+1}}$. We bound the $p$-th moment of this sum with Rosenthal's inequality for martingales from \citet[Theorem~4.1]{pinelis_1994} and using that $\PE [\norm{\funnoisew(\State)}^{2}] = \trace{\noisecov}$. The dependence in $p^{1/2}$ comes from the leading Gaussian term in this inequality. The other terms come from controlling the $p$-th moments  in Rosenthal's inequality and from majorizing the remainder terms. Simplified expressions for $\Deltatr_{n,p,\alpha}$ and $\Deltafl_{n,p,\alpha}$ are given by
\begin{equation}
\label{eq:delta_fl_p_th_moment_simplified}
\Deltafl_{n,p,\alpha} \lesssimd p^{1/2}(\alpha n)^{-1/2} +  \alpha p^{5/2} +  p n^{-1/2}  +  \alpha^{1/2} p^{3/2} \eqsp,\quad
\Deltatr_{n,p,\alpha} \lesssimd  \alpha^{-1} n^{-1/2} +  n^{1/2}\eqsp.
\end{equation}
We again can simplify the bounds of \Cref{th:LSA_PR_error} with a special choice of the step size $\alpha$, proceeding as in \eqref{eq:bound_prop_1_optimized_n_refined}. The fluctuation error term \eqref{eq:delta_fl_p_th_moment_simplified} and the stability result \Cref{fact:exponential-stability-product} suggests that $\alpha$ should scale with $n$ and $p$  as $(pn^{1/2})^{-1}$, therefore justifying the choice of $\alpha(n,d,p)$ given in \eqref{eq:step_size_pth_moment_refined}. Then the bound of \Cref{th:LSA_PR_error} can be re-stated as a high probability bound using the Markov inequality with $p = \log(3\rme/\delta)$. Namely, the following inequality holds:
\begin{corollary}
\label{cor:hp_bound_pr_iid}
Assume \Cref{assum:A-b}, \Cref{assum:noise-level}, \Cref{assum:IID} and set $\delta \in \ooint{0,1}$. Then, for any $\thetainit \in \rset^{d}$, $n \in \nset$, with $\alpha = \alpha(n,d,\log(3\rme/\delta))$ defined in \eqref{eq:step_size_pth_moment_refined}, it holds with probability at least $1-\delta$, that
\begin{equation}
n^{1/2} \norm{\bA\left(\prtheta_{n}-\thetas\right)} \lesssimd  \sqrt{\{\trace{\noisecov}\} \log (3\rme/\delta)}
+ \Delta^{(\HP)}(n,\thetainit, \delta)\eqsp,
\end{equation}
where
\begin{multline}
\Delta^{(\HP)}(n,\thetainit, \delta) = n^{-1/4}\supconsteps \log^{3/2}(3\rme/\delta) \\ + (\log(3\rme/\delta) + \sqrt{n}) \norm{\thetainit - \thetas}
\exp\left\{ - \frac{( \alpha_{\infty} \wedge \smallAconst) \sqrt{n}}{8(1+\log{d})\log(3\rme/\delta)} \right\}  \eqsp.
\end{multline}
\end{corollary}
For completeness, we give the statement of \Cref{cor:hp_bound_pr_iid} with exact constants in \Cref{sec:proof:LSA_PR_error_iid_consts}.

%% file: markov_case.tex
\label{sec:markov}
We now consider the Markov case. Let $(\msz,\metricz)$ be a Polish space endowed with its Borel $\sigma$-field denoted by $\mcz$ and let $(\msz^{\nset}, \mcz^{\otimes \nset})$ be the corresponding canonical space. Consider a Markov kernel $\MKQ$ on $\msz\times \mcz$ and denote by  $\PP_{\xi}$ and $\PE_{\xi}$ the corresponding probability distribution and expectation with initial distribution $\xi$. Without loss of generality, assume that $(Z_k)_{k \in \nset}$  is the associated canonical process. By construction, for any $\msa \in \mcz$, $\CPP[\xi]{Z_k \in \msa}{Z_{k-1}}= \MKQ(Z_{k-1},\msa)$, $\PP_\xi$-a.s. In the case $\xi = \updelta_z$, $z \in \msz$, $\PP_{\xi}$ and $\PE_{\xi}$ are  denoted by $\PP_{z}$ and $\PE_{z}$.
\par
In this section we impose the following assumption on the mixing properties of $\MKQ$:
\begin{assumptionM}
\label{assum:drift}
The Markov kernel $\MKQ$ admits $\invariantQ$ as an invariant distribution and is uniformly geometrically ergodic, that is, there exists $\taumix \in \nsets$ such that for all $k \in \nsets$,
\begin{equation}
\label{eq:drift-condition}
\dobru{\MKQ^k} = \sup_{z,z' \in \Zset} (1/2) \norm{\MKQ^k(z, \cdot) - \MKQ^k(z',\cdot)}[\sf{TV}] \leq (1/4)^{\lfloor k / \taumix \rfloor} \eqsp.
\end{equation}
\end{assumptionM}
Here, $\taumix$ is the mixing time of $\MKQ$. With \eqref{eq:drift-condition} it is easy to see that
\begin{equation}
\label{eq:crr_koef_sum_tau_mix}
\textstyle \sum_{k=0}^{\infty}\dobru{\MKQ^k} = \sum_{\ell =0}^{\taumix-1}\sum_{r=0}^{\infty}\dobru{\MKQ^{\ell + r\taumix}} \leq (4/3)\taumix\eqsp.
\end{equation}
\Cref{assum:drift} implies that $\pi$ is the unique invariant distribution of $\MKQ$. \Cref{assum:drift} is equivalent to the condition that $\MKQ$ satisfies a uniform minorization condition (see \citet[Theorem 18.2.5]{douc:moulines:priouret:soulier:2018}), \ie, there exists a probability measure $\nu$ such that for all $z \in \Zset$, $\msa \in \mcz$,
$\MKQ^{\taumix}(z,\msa) \geq (3/4) \nu(\msa)$. Under \Cref{assum:A-b}, we define the quantity
\begin{align}
\label{eq:alpha_infty_makov}
\alpha_{\infty}^{(\Markov)} &= \left[\alpha_{\infty} \wedge \qcond^{-1/2} \bConst{A}^{-1}\, \wedge\, a/(6\rme \qcond \bConst{A})\right] \times \lceil{8 \qcond^{1/2}\bConst{A} / a\rceil}^{-1}\eqsp, \\
  \label{eq:def_C_Gamma}
\bConst{\Gamma} &= 4(\qcond^{1/2}\bConst{A} + a/6)^{2} \times \lceil 8  \qcond^{1/2}\bConst{A} /a \rceil \eqsp,
\end{align}
where $ \alpha_{\infty}$, $a,\qcond$ are defined in \eqref{eq: kappa_def} and \eqref{eq:def_qcond_b_Q}, respectively. Now we use $\alpha_{\infty}^{(\Markov)}$ and $\bConst{\Gamma}$ to define, for $q \geq 2$,
\begin{equation}
\label{eq:def_alpha_p_infty_Markov}
\alpha^{(\Markov)}_{q,\infty} = \alpha_{\infty}^{(\Markov)} \wedge \smallAconstM/q \eqsp, \quad \smallAconstM = a/\{12 \bConst{\Gamma}\}\eqsp.
\end{equation}
We will see that $\alpha^{(\Markov)}_{q,\infty} \taumix^{-1}$ is a natural counterpart of the stability threshold $\alpha_{q,\infty}$ from \eqref{eq:def_alpha_p_infty}. Our goal now is to prove the counterpart of the stability result for the product of random matrices (cf.~\Cref{fact:exponential-stability-product}) under Markov conditions \Cref{assum:drift}. The main difference with the \iid\ scenario is that the maximum step size which allows for matrix product stability scales with $\taumix^{-1}$. Similar scaling was reported in \cite{srikant:1tsbounds:2019} and \cite{mou2021optimal}.
\begin{proposition}
\label{prop:products_of_matrices_UGE}
Assume \Cref{assum:A-b} and \Cref{assum:drift}. Then, for any $2 \leq p \leq q$, $\alpha \in \ocintLine{0, \alpha_{\infty}^{(\Markov)}\taumix^{-1}}$, $n \in \nset$, and probability distribution $\xi$ on $(\Zset,\Zsigma)$, it holds
\begin{equation}
\label{eq:concentration UGE}
\PE_{\xi}^{1/p}\left[ \normop{\ProdBa_{1:n}}^{p} \right]
\leq  \sqrt{\qcond} \rme^2 d^{1/q} \exp\{- n \alpha a/6  + n (q-1)\alpha^2 \bConst{\Gamma}\} \eqsp,
\end{equation}
where $\alpha_{\infty}^{(\Markov)}$ is defined in \eqref{eq:alpha_infty_makov}. Moreover, for $\alpha \in \ocintLine{0,\alpha^{(\Markov)}_{q,\infty}\taumix^{-1}}$, it holds
\begin{equation}
\label{eq:concentration_UGE_simple}
\PE_{\xi}^{1/p}\left[ \normop{\ProdBa_{1:n}}^{p} \right]
\leq \sqrt{\qcond} \rme^2 d^{1/q} \rme^{- a\alpha n/12} \eqsp.
\end{equation}
\end{proposition}
The proof is given in \Cref{sec:matrix_product_matrix}.
It is based on a simplification of the arguments in \citet{durmus2021stability} together with a new result about the matrix concentration for the product of random matrices, using a proof method introduced in \citet{huang2020matrix}.

Similar to the \iid~case in \Cref{fact:exponential-stability-product}, there is an unavoidable interaction between the choice of step size $\alpha$ and the maximum controlled moment $q$. Moreover, \Cref{rem:choice_stepsize_dimension_undep} can be applied to obtain dimension-independent bounds for $\PE _{\xi}^{1/p}[ \normop{\ProdBa_{1:n}}^{p}]$.
\par
With the above notations, we are ready to state and prove the Markov counterpart of \Cref{th:LSA_PR_error}. Under \Cref{assum:noise-level} and \Cref{assum:drift}, we define the matrix $\noisecovmarkov$ as
\begin{equation}
\label{eq:cov_matrix_epsilon_markov}
\txts \noisecovmarkov = \PE_{\invariantQ}[\funcnoise{\State_0}\funcnoise{\State_0}^\top] + 2\sum_{\ell=0}^{\infty}\PE_{\invariantQ}[\funcnoise{\State_0}\funcnoise{\State_\ell}^\top]\eqsp.
\end{equation}
For any initial probability measure $\xi$ on $(\Zset,\Zsigma)$,  \cite[Theorem 21.2.10]{douc:moulines:priouret:soulier:2018} implies that
$n^{-1/2} \sum_{t=0}^{n-1} \funcnoise{\State_t}$
converges in distribution to the zero-mean Gaussian distribution with covariance matrix
$\noisecovmarkov$.
Hence, $\noisecovmarkov$ is a  counterpart of the covariance matrix $\noisecov$, and we expect it to be the leading term in the bound for $\PE_{\xi}^{1/p}\left[\norm{\bA\left(\prtheta_{n}-\thetas\right)}^{p}\right]$.
\begin{theorem}
\label{th:theo_2_markov}
Assume \Cref{assum:A-b}, \Cref{assum:noise-level}, and \Cref{assum:drift}. Then, for any $p \geq 2$, even $n \geq 4 \vee \taumix$, step size
\begin{equation}
\label{eq:step_size_pth_moment_refined_markov}
\alpha^{(\Markov)}(n,d,p,\taumix) = \left( \alpha_{\infty}^{(\Markov)} \wedge \smallAconstM/\{1+\log{d}\} \right)(p n^{2/3} \taumix^{1/3})^{-1} \eqsp,
\end{equation}
initial parameter $\thetainit \in \rset^d$, and initial probability measure $\xi$ on $(\Zset,\Zsigma)$, it holds that
\begin{multline}
\label{eq:bound_p_th_moment_optimized_markov}
\PE^{1/p}_{\xi}\left[\norm{\bA\left(\prtheta_{n}-\thetas\right)}^{p}\right] \lesssimd \frac{\{\trace{\noisecovmarkov}\}^{1/2} p^{1/2}}{n^{1/2}} + \supconsteps \left(\frac{\taumix^{2/3} p \log{n}}{n^{2/3}} + \frac{\taumix p^{2}}{n}\right) \\
 + p n^{1/2}\norm{\thetainit - \thetas} \exp\left\{ - \frac{(\alpha_{\infty}^{(\Markov)} \wedge \smallAconstM) n^{1/3}}{24 p\taumix^{1/3} (1+\log{d})} \right\}\eqsp.
\end{multline}
\end{theorem}
As in \iid\ case, we provide the generalization of \Cref{th:theo_2_markov} for the case of an arbitrary step size $\alpha \in (0,\alpha^{(\Markov)}_{q,\infty}\taumix^{-1})$ in \Cref{th:LSA_PR_error_markov}, together with the corresponding high-probability bounds (see \Cref{cor:hp_bound_pr_markov}). The expression for the step size \eqref{eq:step_size_pth_moment_refined_markov} now differs from the \iid\ case \eqref{eq:step_size_pth_moment_refined}. Unsurprisingly, the stability result \Cref{prop:products_of_matrices_UGE} requires to scale the step size with $\taumix^{-1}$. At the same time, the optimization upon the sample size $n$ in \Cref{th:LSA_PR_error_markov} suggests now that the optimal step size should scale as $n^{-2/3}$. This is due to the fact that, unlike with the \iid\ noise, the Polyak-Ruppert estimate is no longer an unbiased estimate of $\thetas$ (see \Cref{prop:bias_estimate_PR_markov} and the corresponding discussion).

\par
\Cref{th:theo_2_markov} generalizes and improves the results of \citet[Theorem~1]{mou2021optimal}. First, \citet[Theorem~1]{mou2021optimal} considers only the mean squared error, while in \Cref{th:theo_2_markov} we derive bounds for arbitrary $p$-th moments of the LSA-PR error. These bounds are further used to derive high probability bounds in \Cref{cor:hp_bound_pr_markov} below. Second, the refined bound \eqref{eq:bound_p_th_moment_optimized_markov} for $p=2$ yields the same leading term of order $\{\trace{\noisecovmarkov}\}^{1/2} n^{-1/2}$ and improves the dependence of the residual term on dimension. Indeed, for comparison with \citet[Theorem~1]{mou2021optimal}, we assume that $\supconsteps \approx \sqrt{d}$. This leads to a residual term with a dependence of order $d^{1/2}$ in \Cref{th:theo_2_markov} instead of $d^{2/3}$ in \citet[Theorem~1]{mou2021optimal}. Moreover, the optimal step size $\alpha$ in \eqref{eq:step_size_pth_moment_refined_markov} scales with $d$ as $(1+\log{d})^{-1}$, unlike $d^{-1/3}$ in \citet[Theorem~1]{mou2021optimal}.

\paragraph{Bounds on the non-averaged LSA iterates} Similar to the \iid\ setting, we first obtain a preliminary bound on  the $p$-th moment of the LSA error $\norm{\theta_n - \thetas}$. In this preliminary result, we are especially  interested in obtaining a sharp bound with respect to the step size $\alpha$, for which we only rely on the first decomposition \eqref{eq:decomp_fluctuation}. We now give the bounds
 for $\PE ^{1/p}_{\xi}\bigl[\norm{\Jnalpha{n}{0}}^{p}\bigr]$ and $\PE_\xi^{1/p} \left[\norm{\theta_n - \thetas}^{p}\right]$, which match the corresponding bounds of \Cref{prop:J_n_0_bound_iid} and \Cref{cor:LSA_err_bound_iid} up to absolute constants and the factor $\sqrt{\taumix}$.
\begin{proposition}
\label{prop:J_n_0_bound_Markov}
Assume \Cref{assum:A-b}, \Cref{assum:noise-level}, and \Cref{assum:drift}. Then, for any $\alpha \in \ocintLine{0,\alpha_{\infty}}$, $p \geq 2$, initial probability measure $\xi$ on $(\Zset,\Zsigma)$, and $n \in \nset$, it holds
\begin{equation}
\label{eq:J_n_0_bound_Markov}
\PE^{1/p}_{\xi}\bigl[\norm{\Jnalpha{n}{0}}^{p}\bigr]  \leq \ConstDM_{1} \sqrt{\alpha a p \taumix} \supconsteps \eqsp,
\end{equation}
where $\ConstDM_{1}$ is defined in \eqref{eq:const_D_1_Markov}.
\end{proposition}
The proof is provided in \Cref{sec:proof_j_n_0_bound_Markov}.
By definition \eqref{eq:J_n_0_repr_iid}, $\Jnalpha{n}{0}$ is a linear statistics of the Markov chain $(\State_k)_{k \in \nset}$. Thus the desired result follows from a Mac-Diarmid type inequality under \Cref{assum:drift} (see \citet[Corollary~2.10]{paulin_concentration_spectral}). Note that the bound \eqref{eq:J_n_0_bound_Markov} is similar to the one established in \Cref{prop:J_n_0_bound_iid} up to an additional $\sqrt{\taumix}$ factor.
\begin{proposition}
\label{prop:LSA_error_Markov}
Assume \Cref{assum:A-b}, \Cref{assum:noise-level}, and \Cref{assum:drift}. Let $2 \leq p \leq q/2$ and $\alpha^{(\Markov)}_{q,\infty}$ be defined in \eqref{eq:def_alpha_p_infty_Markov}. Then, for any $\alpha \in \ocintLine{0, \alpha_{q,\infty}^{(\Markov)} \taumix^{-1}}$, $\thetainit \in \rset^d$, initial probability measure $\xi$ on $(\Zset,\Zsigma)$, and $n \in \nset$, it holds
\begin{equation}
\label{eq:n_step_drift_main}
\PE_\xi^{1/p} \left[\norm{\theta_n - \thetas}^{p}\right] \leq \sqrt{\qcond} \rme^2 d^{1/q} \rme^{-\alpha a n/12} \norm{\thetainit- \thetas} + \ConstDM_{2} d^{1/q} \sqrt{\alpha a p \taumix} \supconsteps\,,
\end{equation}
where  $\ConstDM_2$ is defined in \eqref{eq:definition:ConstDM_2}.
\end{proposition}
The proof is postponed to \Cref{proof:prop:LSA_error_Markov}.
\Cref{prop:LSA_error_Markov} improves the results obtained in \citet[Proposition~1]{mou2021optimal}. First, we derive a better scaling with respect to $p$ for the fluctuation term. Indeed, \citet[Proposition 1]{mou2021optimal} implies that this term scales with $p^{3/2}$, while we obtain $p^{1/2}$. Moreover, the constraints on the step size $\alpha$ are relaxed. \Cref{prop:LSA_error_Markov} holds for the step size $\alpha \lesssim 1/[p(1+\log{d})\taumix]$, while \citet[Proposition 1]{mou2021optimal} requires that $\alpha \lesssim 1/[p^3d \taumix]$.
\paragraph{Outline of the proof of \Cref{th:theo_2_markov}}
We state below a counterpart of \Cref{prop:J_n_1_2_bound_iid}. The objective, as in the i.i.d. case, is to obtain a better control of $\PE^{1/p}_\xi\bigl[\norm{\Hnalpha{n}{0}}^{p}\bigr]$, which we do here by using the decomposition \eqref{eq:decomposition_H_n_0}.
\begin{proposition}
\label{prop:J_n_1_2_bound_markov}
Assume \Cref{assum:A-b}, \Cref{assum:noise-level}, and \Cref{assum:drift}. Then, for any $p \geq 2$, $\alpha \in \ocint{0,\alpha_{\infty}}$, and initial probability measure $\xi$ on $(\Zset,\Zsigma)$, it holds that
\begin{equation}
\label{eq:j_n_1_bound_markov}
\PE^{1/p}_{\xi}[\normop{\Jnalpha{n}{1}}^{p}] \leq \supconsteps (\alpha a \taumix) \{\ConstDM_{J,1} \sqrt{\log{(1/\alpha a)}} p^{2} + \ConstDM_{J,2}  (\alpha a \taumix)^{1/2} p^{1/2}\}\eqsp,
\end{equation}
where $\ConstDM_{J,1}$ and $\ConstDM_{J,2}$ are defined in \eqref{eq:def:ConstDM_J,1} and \eqref{eq:def:ConstDM_J,2} respectively. In addition, for any $p,q \geq 2$, satisyfing $2 \leq p \leq q/2$, $\alpha \in \ocintLine{0,\alpha^{(\Markov)}_{q,\infty}\taumix^{-1}}$, and initial probability measure $\xi$ on $(\Zset,\Zsigma)$, it holds that
\begin{equation}
\label{eq:H_n_1_bound_markov}
 \PE^{1/p}_{\xi}[\normop{\Hnalpha{n}{1}}^{p}] \leq d^{1/q} \supconsteps (\alpha a \taumix)\bigl[\ConstDM_{H,1}   \sqrt{\log{(1/\alpha a)}} p^{2} + \ConstDM_{H,2} (\alpha a \taumix)^{1/2} p^{1/2}\bigr]\eqsp,
\end{equation}
where  $\ConstDM_{H,1}$ and $\ConstDM_{H,2}$ are defined in \eqref{eq:def:ConstDM_H}.
\end{proposition}
The proof is postponed to \Cref{sec:proof:prop:J_n_1_2_bound_markov}.
Unlike the case of \iid-noise, $\Jnalpha{n}{1}$ is no longer a martingale, so we cannot directly apply Rosenthal-type inequalities to upper bound $\PE ^{1/p}_{\xi}[\normop{\Jnalpha{n}{1}}^{p}]$. Instead, we rely on Berbee's lemma (\citet[Lemma~5.1]{riobook}). The leading term in the bound of $\PE ^{1/p}_{\xi}[\normop{\Jnalpha{n}{1}}^{p}] $ is $(\alpha a \taumix) \sqrt{\log{(1/\alpha a)}}$ in the Markov case instead of $\alpha a$ in the \iid\ case. The factor $\sqrt{\log{(1/\alpha a)}}$ is a byproduct of  Berbee's inequality and is most likely an artifact of the proof.

With the above estimates, we are ready to state and prove the counterpart of \Cref{th:LSA_PR_error} under \Cref{assum:drift}.
Unlike the \iid\ case, the Polyak-Ruppert estimate \eqref{eq:lsa} is not an unbiased estimate for $\thetas$. We quantify this resulting bias in our next result.
\begin{proposition}
\label{prop:bias_estimate_PR_markov}
Assume \Cref{assum:A-b}, \Cref{assum:noise-level}, and \Cref{assum:drift}. Then, for any $\alpha \in \ocintLine{0, \alpha_{2(1+\log{d}),\infty}^{(\Markov)} \taumix^{-1}}$, $\thetainit \in \rset^d$, initial probability measure $\xi$ on $(\Zset,\Zsigma)$, and even $n \geq 2$, it holds that
\begin{equation}
\label{eq:bias_estimate_PR_markov}
\norm{\PE_\xi[\prtheta_{n}] - \thetas} \leq \frac{\ConstDM_{4} \rme^{-\alpha a n/24} \norm{\thetainit- \thetas}}{\alpha a n} + \ConstDM_{5} \supconsteps (\alpha a \taumix) \sqrt{\log{(1/\alpha a)}} + \ConstDM_{6} \supconsteps (\alpha a \taumix)^{3/2} \eqsp,
\end{equation}
where  $\ConstDM_{4}, \ConstDM_{5}, \ConstDM_{6}$ are defined in \eqref{eq:def:const_D_4_M}.
\end{proposition}
The proof is given in \Cref{prop:bias_estimate_PR_markov_proof}. Note that our bound on the bias scales as $\mathcal{O}(\alpha \taumix)$ up to the factor $\sqrt{\log{(1/\alpha a)}}$. Similar bounds on the bias of LSA-PR procedure are provided in \cite[Section~2]{meyn_bias_sa}. Moreover, one could conclude that $\prtheta_{n}$ is not an unbiased estimate of $\thetas$ using the decomposition \eqref{eq:pr_err_decompose}. Indeed, we notice that in the fluctuation term $\Efl_{n}$, defined in \eqref{eq:def:Efl}, the term $\sum_{t=n/2}^{n-1}  \zmfuncA{\State_{t+1}} \Jnalpha{t}{0}$ has not mean zero. This comes in contrast to the \iid~case where  $\zmfuncA{\State_{t+1}}$ and $\Jnalpha{t}{0}$ are independent and have zero mean. In fact, $\norm{\PE_\xi[\zmfuncA{\State_{t+1}} \Jnalpha{t}{0}]}$ scales as $\mathcal{O}(\alpha \taumix)$. The precise statement is given in \Cref{prop:bias_Markov_PR} in appendix.

Equipped with the above bounds, we can prove the $p$-th moment bound of the LSA-PR error under the Markovian noise dynamics.
\begin{theorem}
\label{th:LSA_PR_error_markov}
Assume \Cref{assum:A-b}, \Cref{assum:noise-level}, and \Cref{assum:drift}. Then, for any $p \geq 2$, $\alpha \in \ocintLine{0, \alpha_{p(1+\log{d}),\infty}^{(\Markov)} \taumix^{-1}}$, even $n \geq 4$, $\thetainit \in \rset^d$, and initial probability measure $\xi$ on $(\Zset,\Zsigma)$, it holds
\begin{equation}
\label{eq:p_th_moment_LSA_bound_Markov}
\begin{split}
(n/2)^{1/2}\PE_{\xi}^{1/p}\left[\norm{\bA\left(\prtheta_{n}-\thetas\right)}^{p}\right]
 &\leq  \bConst{\sf{Rm}, 1} \{\trace{\noisecovmarkov}\}^{1/2} p^{1/2} + \supconsteps \MarDeltafl_{n,p,\alpha,\taumix} \\
 &\qquad + \MarDeltatr_{n,p,\alpha,\taumix} \norm{\thetainit- \thetas} \exp\left\{-\alpha a n/24\right\}\eqsp,
\end{split}
\end{equation}
where
$\MarDeltafl_{n,p,\alpha,\taumix}$ and  $\MarDeltatr_{n,p,\alpha,\taumix}$ are defined in \eqref{eq:def:const:LSA_PR_markov}.
\end{theorem}
The proof is postponed to \Cref{proof:LSA_PR_error_markov}. The terms $\MarDeltafl_{n,p,\alpha,\taumix}$ and $\MarDeltatr_{n,p,\alpha,\taumix}$ correspond to the fluctuation and transient terms of the error decomposition, respectively.  Simplified expressions of $\MarDeltafl_{n,p,\alpha,\taumix}$ and  $\MarDeltatr_{n,p,\alpha,\taumix}$  are given by
\begin{align}
\nonumber
\textstyle &\MarDeltafl_{n,p,\alpha,\taumix} \lesssimd  (p \taumix)^{1/2} (\alpha n)^{-1/2} + p \taumix^{3/4} n^{-1/4} \\
&\qquad \qquad \qquad \qquad\qquad + \taumix [ (\log(1/\alpha))^{1/2} p^2 + (\alpha \taumix)^{1/2} p^{1/2} ] (n^{-1/2} + \alpha n^{1/2})
\eqsp, \\
\label{eq:def:const:LSA_PR_markov_simplified}
\textstyle & \MarDeltatr_{n,p,\alpha,\taumix} \lesssimd  \alpha^{-1} n^{-1/2} +  n^{1/2} \eqsp.
\end{align}
The bound for the transient term is similar to the i.i.d. case, only the numerical constants are affected. The expression for the fluctuation term is  more complicated.
Unlike the \iid\ case, $\{e(\theta_t,\State_{t+1})\}_{t \in \nset}$ is no longer a martingale increment sequence (see the decomposition \eqref{eq:pr_err_decompose} and \eqref{eq:pr_e_definition}). By \eqref{eq:def:Efl} we have the decomposition
\begin{multline}
\bA\left(\prtheta_{n}-\thetas\right) = 2(\alpha n)^{-1}(\theta_{n/2}-\theta_{n}) - 2n^{-1}\Etr_{n/2,n} + \\
\underbrace{2 n^{-1} \sum_{t=n/2}^{n-1} \funcnoise{\State_{t+1}} + 2 n^{-1} \sum_{t=n/2}^{n-1}\zmfuncA{\State_{t+1}} \Jnalpha{t}{0} + 2 n^{-1} \sum_{t=n/2}^{n-1} \zmfuncA{\State_{t+1}} \Hnalpha{t}{0}}_{\text{fluctuation term}}\eqsp.
\end{multline}
The fluctuation term above consists of $3$ summands. The first one is $(n/2)^{-1} \sum_{t=n/2}^{n-1} \funcnoise{\State_{t+1}}$, which is an additive functional of the uniformly geometrically ergodic Markov chain $\{\State_t\}_{t\in\nsets}$ under \Cref{assum:drift}. Using a novel Rosenthal inequality from \cite[Theorem~1]{moulines23_rosenthal}, we get that the leading term of the $p$-th moment of this quantity scales as $\{\trace{\noisecovmarkov}\}^{1/2} p^{1/2}$. This is also the leading term in the bound \eqref{eq:p_th_moment_LSA_bound_Markov}. The remainder terms in this inequality are more involved than in the \iid\ case, which explains the occurrence of the term of order $p \taumix^{3/4} n^{-1/4}$. Regarding the second term in the fluctuation component, we already mentioned that  in general $\PE_{\xi}[\zmfuncAw(\State_{t+1}) \Jnalpha{t}{0}] \neq 0$, in contrast to the \iid~case. Moreover, we provide the bound on the quantity $\PE_{\xi}[\zmfuncAw(\State_{t+1}) \Jnalpha{t}{0}]$ in \Cref{lem:bias_estimate_J_n_0_Markov}. This bound is of the same order (up to the factor $\sqrt{\log{1/\alpha a}}\,$) with respect to the step size $\alpha$ as the bound for $\Hnalpha{t}{0}$, that can be obtained through \Cref{prop:J_n_1_2_bound_markov} (recall that $\Hnalpha{n}{0} = \Jnalpha{n}{1} + \Hnalpha{n}{1}$). This fact can also be applied to control the last summand in the decomposition and explains why we did not consider the higher-order expansions for $H_{n,1}$.
\par
The bound of \Cref{th:LSA_PR_error_markov} can be refined under the special choice of the step size $\alpha$. The fluctuation error term \eqref{eq:def:const:LSA_PR_markov_simplified} suggests that $\alpha$ should scale with $n$ as $n^{-2/3}$, which explains the dependence of $\alpha^{(\Markov)}(n,d,p,\taumix)$ in \eqref{eq:step_size_pth_moment_refined_markov} upon the sample size $n$. Similar to \Cref{cor:hp_bound_pr_iid}, the bound of \Cref{th:LSA_PR_error_markov} can be reformulated as a high probability bound using the Markov inequality.
\begin{corollary}
\label{cor:hp_bound_pr_markov}
Assume \Cref{assum:A-b}, \Cref{assum:noise-level}, \Cref{assum:drift}, and set $\delta \in \ooint{0,1}$. Then, for any $\thetainit \in \rset^{d}$, sample size $n \in \nsets$, $n \geq 4 \vee \taumix$, and initial probability measure $\xi$ on $(\Zset,\Zsigma)$, choosing the step size $\alpha = \alpha^{(\Markov)}(n,d,\log(3\rme/\delta),\taumix)$ defined in \eqref{eq:step_size_pth_moment_refined_markov}, it holds with probability at least $1-\delta$, that
\begin{equation}
n^{1/2} \norm{\bA\left(\prtheta_{n}-\thetas\right)} \lesssim_{d} \sqrt{\{\trace{\noisecovmarkov}\} \log (3\rme/\delta)} + c_1^{(\Markov)} R^{(\HP)}(n,\thetainit, \delta,\taumix)\eqsp,
\end{equation}
where
\begin{align}
&R^{(\HP)}(n,\thetainit, \delta,\taumix) = \supconsteps \log(3\rme/\delta) \left(n^{-1/6}  \log{(n)} \taumix^{2/3} + n^{-1/2} \taumix \log(3\rme/\delta)\right) \\
&\qquad + (n^{1/6}\taumix^{1/3}\log(3\rme/\delta) + n^{1/2}) \norm{\thetainit - \thetas}
\exp\left\{ - \frac{(\alpha_{\infty}^{(\Markov)} \wedge \smallAconstM)n^{1/3}}{24\taumix^{1/3}(1+\log{d})\log(3\rme/\delta)} \right\}\eqsp.
\end{align}
\end{corollary}


%% file: appendix_const.tex
Denote $\nsets = \nset \setminus \{0\}$ and $\nsetm = \zset \setminus \nsets$.
Let $d \in \nsets$ and $Q$ be a symmetric positive definite $d \times d$ matrix. For $x \in \rset^d$, we denote $\norm{x}[Q]= \{x^\top Q x\}^{\half}$. For brevity, we set $\norm{x}= \norm{x}[\Id_d]$. We denote $\normop{A}[Q]= \max_{\norm{x}[Q]=1} \norm{Ax}[Q]$, and the subscriptless norm $\normop{A} = \normop{A}[\Id]$ is the standard spectral norm.
For a function $g: \Zset \to \rset^d$, we denote $\supnorm{g} = \sup_{z \in \Zset}\norm{g(z)}$.

We denote $\sphere^{d-1} = \{x \in \rset^{d} \, : \, \norm{x} = 1\}$. Let $A_{1},\ldots,A_N$ be $d$-dimensional matrices. We denote $\prod_{\ell=i}^j A_\ell = A_j \ldots A_i$ if $i\leq j$ and by convention $\prod_{\ell=i}^j A_\ell = \Id$ if $i > j$. We say that a centered random variable (r.v.) $X$ is subgaussian with variance proxy factor $\sigma^2$ and denote $X \in \SG (\sigma^2)$ if for all $\lambda \in \rset$, $\log \PE [\rme^{\lambda X}] \leq \lambda^2\sigma^2/2$. 

The readers can refer to the following table on the variables, constants and notations that are used across the paper for references.

\begin{center}
\begin{tabular}{l l l}
\hline
\bfseries Variable & \bfseries Description & \bfseries Reference \\
\hline
$Q$ & Solution of Lyapunov equation for $\bA$ & \Cref{fact:Hurwitzstability} \\
$\qcond$ & $\lambda_{\sf min}^{-1}(Q) \lambda_{\sf max}(Q)$ & \Cref{fact:Hurwitzstability} \\
$a$ & Real part of minimum eigenvalue of $\bA$ & \Cref{fact:Hurwitzstability} \\
$\ProdBa_{m:n}$ & Product of random matrices with step size $\alpha$ & \eqref{eq:definition-Phi} \\
$\funnoisew(\State_n)$ & Noise in LSA procedure & \eqref{eq:def_center_version_and_noise} \\
$\utheta_{n}, \vtheta_{n}$ & Transient and fluctuation terms of LSA error & \eqref{eq:LSA_recursion_expanded} \\

$\alpha_{p,\infty}$ (resp. $\alpha^{(M)}_{p,\infty}$) & \multirow{2}{*}{\shortstack[l]{Stability threshold for $\ProdBa_{m:n}$ to have bounded \\ $p$-th moment under \Cref{assum:IID} (resp. \Cref{assum:drift})}} & \eqref{eq:def_alpha_p_infty} \\
& & \\
$\Jnalpha{n}{0}$ & Dominant term in $\vtheta_n$ & \eqref{eq:jn0_main} \\
$\Hnalpha{n}{0}$ & Residual term $\vtheta_n - \Jnalpha{n}{0}$ & \eqref{eq:jn0_main} \\
$\Jnalpha{n}{1},\Hnalpha{n}{1}$,  & Expansion of $\Hnalpha{n}{0}$ & \eqref{eq:decomposition_H_n_0}-\eqref{eq:jn_allexpansion_main} \\
$\noisecov$ & Noise covariance $\PE[\eps_1 \eps_1^{\transpose}]$ & \Cref{assum:noise-level} \\
$\noisecovmarkov$ & \multirow{2}{*}{\shortstack[l]{Asymptotic covariance matrix \\ under Markovian noise}} & \eqref{eq:cov_matrix_epsilon_markov} \\
& & \\
$\bConst{\sf{Rm}, 1} = 60 \rme$ & Constant in martingale Rosenthal's inequality & \cite[Theorem~4.1]{pinelis_1994}  \\
$\bConst{\sf{Rm}, 2} = 60$ & Constant in martingale Rosenthal's inequality & \cite[Theorem~4.1]{pinelis_1994} \\
$\bConst{\sf{Ros}, 1}^{(\Markov)}, \bConst{\sf{Ros}, 2}^{(\Markov)}$ & Constants in Rosenthal's inequallity under \Cref{assum:drift} & \Cref{theo:rosenthal_uge_arbitrary_init} \\
$\{\mcf_t\}_{t \in\nset}$ & filtration $\mcf_t= \sigma(\State_s \, : \,  1 \leq s \leq t)$ with $\mcf_0 = \{\emptyset, \msz\}$ &
\\
 $\PE^{\mcf_t}$ &  the conditional expectation with respect to $\mcf_t$ & \\
\hline
\end{tabular}
\end{center} 

%% file: appendix_iid.tex
In the lemmas below we use the shorthand notations $\zmfuncAw[n], \Am_{n} ,\funnoisew_{n}$ for $\zmfuncA{\State_{n}}, \Am(\State_n)$, and $\funcnoise{\State_{n}}$, respectively, where $\funnoisew(z): \zset \to \rset^{d}$ is defined in \eqref{eq:def_center_version_and_noise}.
For $t \in \nset$, we define the filtration $\mcf_t= \sigma(\State_s \, : \,  1 \leq s \leq t)$, $\mcf_0 = \{\emptyset, \msz\}$, and denote by $\PE^{\mcf_t}$ the conditional expectation with respect to $\mcf_t$.

\subsection{Proof of \Cref{prop:J_n_0_bound_iid}}
\label{sec:proof_j_n_0_iid}
Recall that the constant $\ConstD_{1}$ is defined as $\ConstD_1 = \sqrt{2 \qcond}/a$.
With the decomposition \eqref{eq:jn0_main}, we expand $\Jnalpha{n}{0}$ as
\begin{equation}
\textstyle \Jnalpha{n}{0} = \alpha \sum_{j=1}^{n}\left(\Id - \alpha \bA \right)^{n-j}\funnoisew_{j} =: \alpha \sum_{j=1}^{n}\eta_{n,j}\eqsp, \text{ where } \eta_{n,j} = \left(\Id - \alpha \bA \right)^{n-j}\funnoisew_{j}\eqsp.
\end{equation}
\Cref{fact:Hurwitzstability} implies that $\normop{\left(\Id - \alpha \bA \right)^{n-j}} \leq \qcond^{1/2} (1-\alpha a)^{(n-j)/2}$. Hence, using  \Cref{lem:bounded_differences_norms} and \Cref{assum:noise-level}, we get for any $t \geq 0$ that
\begin{align}
\PP(\normLine{\Jnalpha{n}{0}} \geq t) \leq 2\exp\{-t^2/(2\sigma^{2}_{\alpha,n})\}\eqsp,
\end{align}
where
\begin{equation}
\textstyle \sigma^{2}_{\alpha,n} = \alpha^{2}  \qcond \supconsteps^{2} \sum_{j=1}^{n} (1-\alpha a)^{n-j} \leq \alpha  \qcond \supconsteps^{2} /a\eqsp.
\end{equation}
Combining this result with \Cref{lem:bound_subgaussian} yields \eqref{eq:J_n_0_bound_iid}.

\subsection{Proof of \Cref{cor:LSA_err_bound_iid}}
\label{sec:proof_h_n_0_iid}
Define the constant $\ConstD_{2}$ as 
\begin{equation}
\label{eq:constants_D_definition}
\textstyle \ConstD_{2} =(2 \qcond)^{1/2} a^{-1}(1 + 4 \qcond^{1/2} \bConst{A}  a^{-1})\eqsp.
\end{equation}
Using the main expansion \eqref{eq:decomp_fluctuation} and Minkowski's inequality,
\begin{equation}
\label{eq:p_norm_lsa_err_appendix}
\PE^{1/p}\left[\norm{\theta_{n} - \thetas}^{p}\right] \leq
\PE^{1/p}\left[\norm{\ProdBa_{1:n} (\thetainit - \thetas)}^{p}\right] +
\PE^{1/p}\left[\norm{\Jnalpha{n}{0}}^{p}\right] + \PE^{1/p}\left[\norm{\Hnalpha{n}{0}}^{p}\right]\eqsp.
\end{equation}
Applying \Cref{fact:exponential-stability-product}, using that $\alpha a \leq 1/2$ by \Cref{fact:Hurwitzstability}, and $(1-t)^{1/2} \leq 1-t/2$ for $t \in\ccint{0,1}$,
\begin{equation}
\label{eq:transient_term_bound}
\PE^{1/p}\left[\norm{\ProdBa_{1:n}(\thetainit - \thetas)}^{p}\right] \leq \qcond^{1/2} d^{1/q} (1 - \alpha a/4)^{n}\norm{\thetainit-\thetas}\eqsp.
\end{equation}
With \Cref{prop:J_n_0_bound_iid}, we get $\PE^{1/p}[\norm{\Jnalpha{n}{0}}^{p}] \leq \ConstD_1 \sqrt{\alpha a  p}\supconsteps$. It remains to bound $\PE^{1/p}[\norm{\Hnalpha{n}{0}}^{p}]$. Expanding the recurrence \eqref{eq:jn0_main}, we represent
\begin{equation}
\textstyle \Hnalpha{n}{0} = -\alpha \sum_{j=1}^{n}\ProdBa_{j+1:n}\zmfuncA{Z_j} \Jnalpha{j-1}{0}\eqsp.
\end{equation}
Using Minkowski's inequality and since $\zmfuncA{Z_j}$ and $\Jnalpha{j-1}{0}$ are independent under \Cref{assum:IID}, we obtain with \Cref{assum:A-b}, that
\begin{equation}
\label{eq:H_n_alpha_expansion}
\textstyle \PE^{1/p}\left[\norm{\Hnalpha{n}{0}}^{p}\right] \leq \alpha \bConst{A} \sum_{j=1}^{n} \PE^{1/p}\left[\normop{\ProdBa_{j+1:n}}^{p}\right] \PE^{1/p}\left[\norm{\Jnalpha{j-1}{0}}^{p}\right]\eqsp.
\end{equation}
Now \eqref{eq:concentration_iid_bounded_stepsize} and \Cref{prop:J_n_0_bound_iid} yield
\begin{equation}
\label{eq:H_n_alpha_bound}
\textstyle \PE^{1/p}\left[\norm{\Hnalpha{n}{0}}^{p}\right] \leq \Auxconst_{1} d^{1/q} \sqrt{\alpha  a  p} \supconsteps \eqsp, \text{ where }  \Auxconst_{1} = 4 \ConstD_1 \bConst{A} \qcond^{1/2}/ a \eqsp.
\end{equation}Combining the bounds above in \eqref{eq:p_norm_lsa_err_appendix} completes the proof.

\subsection{Proof of \Cref{prop:mse_iid_with_burn_in}}
\label{sec:proof:prop:mse_iid_with_burn_in}
Define
\begin{equation}
\label{eq:delta_transient_fluctuation_terms}
\begin{aligned}
\Deltatr_{n,\alpha} = 32\rme \qcond/(\alpha^{2} n) + 128 \rme \qcond \bConst{A}^{2}/(7 \alpha a n) 
\eqsp, \qquad \Deltafl_{n,\alpha} = 64\rme  \ConstD_{2}^{2}/(\alpha n) + 16 \rme \alpha \bConst{A}^2  \ConstD_{2}^{2}\eqsp.
\end{aligned}
\end{equation}
Let $q \geq 2$ be a number to be fixed later, and assume that $\alpha \in (0, \alpha_{q,\infty})$. We need this additional degree of freedom to ensure that our bounds are dimension-free. Recall for $k \in \nset$, $\mcf_k= \sigma(\State_s \, : \,  1 \leq s \leq k)$, $\mcf_0 = \{\emptyset, \msz\}$. 
Our proof is based on the decomposition \eqref{eq:pr_err_decompose}. Under \Cref{assum:IID}, $\CPE{e(\theta_t,Z_{t+1})}{\mcf_t}= 0$ $\PP$-a.s., showing that $e(\theta_t,Z_{t+1})$ is an $\mcf_t$-martingale increment. Thus, exploiting \eqref{eq:pr_err_decompose}, we proceed with decomposition \eqref{eq:theta_pr_mse_iid} and estimate the terms $T_1$ and $T_2$ separately. To control the remainder term $T_2$, we apply \Cref{cor:LSA_err_bound_iid} with $p=2$, and obtain
\begin{align}
\label{eq:bound_t2_prop_mse_iid}
T_2 \leq \frac{32 d^{2/q} \qcond \norm{\thetainit - \thetas}^{2}(1- \alpha a/4)^{n}}{\alpha^2 n} + \frac{64  d^{2/q}\ConstD_{2}^{2} a \supconsteps^{2}}{\alpha n} \eqsp.
\end{align}
Now we bound $T_1$. Recall that for $\theta \in \rset^{d}$, $z \in \Zset$, $e(\theta,z) = \funcnoise{z} + \zmfuncA{z} (\theta - \thetas)$. Hence,
\begin{align}
\PE\bigl[\norm{\rme(\theta_{t},\State_{t+1})}^{2}\bigr]
\leq 2 \trace{\noisecov} + 2 \PE\bigl[\norm{\zmfuncA{\State_{t+1}} \{\theta_{t} - \thetas\}}^{2}\bigr]\eqsp,
\end{align}
where we used that $\PE[\norm{\funnoisew_{t}}^{2}] = \trace{\noisecov}$. \Cref{cor:LSA_err_bound_iid} together with $\alpha \leq \alpha_{\infty}$ and $\alpha a \leq 1/2$ yields
\begin{multline}
\textstyle \sum_{t=n/2}^{n-1} \PE\bigl[\norm{\rme(\theta_{t},\State_{t+1})}^{2}\bigr] \leq n \trace{\noisecov} + 4 \alpha a n d^{2/q} \ConstD_2^{2}\bConst{A}^{2} \supconsteps^{2} \\ \nonumber +
\frac{32 d^{2/q} \qcond \bConst{A}^{2} \norm{\thetainit - \thetas}^{2}}{7\alpha a}(1-\alpha a/4)^{n}\eqsp.
\end{multline}
It remains to combine the above bounds and to choose $q = 2(1+\log{d})$, and the elementary inequality $d^{2/(2+2\log{d})} \leq \rme$.

\subsection{Proof of \Cref{prop:J_n_1_2_bound_iid}}\label{sec:proof_j_n_1_iid}
Define
\begin{equation}
\label{eq:def:constD_3_4}
\ConstD_3 =  2 \qcond  \bConst{A}/a^{2} \eqsp, \quad \text{and}\quad \ConstD_{4} = 4 \qcond^{1/2} \bConst{A} \ConstD_{3}/a \eqsp.
\end{equation}
\subsubsection{Moment bounds for $\Jnalpha{n}{1}$} We begin with the proof of \eqref{eq:J_n_1_bound_iid_bounded}.
Expanding the recurrence \eqref{eq:jn_allexpansion_main} with $\ell=1$ and using that $\Jnalpha{k-1}{0} = -\alpha \sum_{i=1}^{k-1}  (\Id - \alpha \bA)^{k-i-1} \varepsilon_{i}$ yields
\begin{equation}
\label{eq:S_ell_def}
\textstyle 
\Jnalpha{n}{1} = \alpha^2 \sum_{i=1}^{n-1} S^{(1)}_{i+1:n} \funnoisew_{i}, ~~\text{where} ~~ S^{(1)}_{i+1:n} = \sum_{k=i+1}^{n} (\Id - \alpha \bA)^{n-k} \zmfuncAw[k] (\Id - \alpha \bA)^{k-1 - i} \eqsp.
\end{equation}
Recall for $k \in \nset$, $\mcf_k= \sigma(\State_s \, : \,  1 \leq s \leq k)$, $\mcf_0 = \{\emptyset, \msz\}$.
It is easy to check that the sequence $\{ S^{(1)}_{i+1:n} \funnoisew_{i}\}_{i=1}^{n-1}$ is a martingale-difference with respect to the filtration $(\mcf_k)_{k \in\nset}$: $\PE[S^{(1)}_{i+1:n} \funnoisew_{i} | \mcf_{i-1}] = 0$.  Applying the Burkholder inequality \citet[Theorem 8.6]{osekowski:2012} and the Minkowski inequality, we get
\begin{align}
\label{eq:pinelis_J_n_1_bounded}
\textstyle \PE[\norm{\Jnalpha{n}{1}}^p ]
  &\leq \textstyle  p^p \alpha^{2p} \PE [(\sum_{i=1}^{n-1}\norm{S^{(1)}_{i+1:n} \funnoisew_{i} }^2 )^{p/2}] \\
 &\leq  \textstyle p^{p} \alpha^{2p} ( \sum_{i=1}^{n-1}  \PE^{2/p}[\norm{S^{(1)}_{i+1:n} \funnoisew_{i}}^p])^{p/2} \eqsp.
\end{align}
 Let us denote $v_{i} = \funnoisew_{i} / \norm{\funnoisew_{i}}$. Then, using \Cref{assum:IID}, we get
\begin{align}
\label{eq:cond_expectation_inserted}
\PE[\norm{S^{(1)}_{i+1:n} \funnoisew_{i}}^p] = \PE\Bigl[\norm{\funnoisew_{i}}^{p} \CPE{\norm{S^{(1)}_{i+1:n} v_{i}}^p}{\mcf_i} \Bigr] \leq \PE[\norm{\funnoisew_{i}}^{p}] \sup_{u \in \sphere^{d-1}}\PE[\norm{S^{(1)}_{i+1:n} u}^p]\eqsp.
\end{align}
\Cref{assum:A-b} and \Cref{fact:Hurwitzstability} imply that $\norm{(\Id - \alpha \bA)^{n-k} \zmfuncAw[k] (\Id - \alpha \bA)^{k-1 - i}} \leq \qcond \bConst{A}(1-\alpha a)^{(n-i-1)/2}$. Hence, applying \Cref{lem:bounded_differences_norms}, we get for any $t \geq 0$ and $u \in \sphere^{d-1}$ that
\begin{align}
\PP\bigl(\norm{S^{(1)}_{i+1:n} u} \geq t\bigr) \leq 2\exp\left\{-\frac{t^2}{2\qcond^2\bConst{A}^2(n-i)(1-\alpha a)^{n-i-1}}\right\}\eqsp.
\end{align}
Applying \Cref{lem:bound_subgaussian}, we get for any  $u \in \sphere^{d-1}$
\begin{equation}
\label{eq:l_p_bound_V_ell_bounded}
\PE^{2/p}[\norm{S^{(1)}_{i+1:n} u}^p] \leq 2 p \bConst{A}^2\qcond^2(n-i)(1-\alpha a)^{n-i-1} \eqsp.
\end{equation}
Combining \eqref{eq:pinelis_J_n_1_bounded}, \eqref{eq:l_p_bound_V_ell_bounded}, and \Cref{assum:noise-level}, we get
\begin{equation}
\label{eq:J_n_1_fin}
\begin{split}
\PE^{1/p}[\norm{\Jnalpha{n}{1}}^p]
& \textstyle \leq 2 \supconsteps p^{3/2} \alpha^{2}  \bConst{A}\qcond ( \sum_{i=1}^{n-1} (n-i) (1-\alpha a)^{n-i-1} )^{1/2} \\
&\leq \ConstD_3 \alpha a  p^{3/2} \supconsteps \eqsp,
\end{split}
\end{equation}
where $\ConstD_3$ is defined in \eqref{eq:J_n_1_bound_iid_bounded}. In the above we have used that $\sum_{k=1}^{\infty} k \rho^{k-1} = (1-\rho)^{-2}$ for $\rho \in\coint{0,1}$ together with $\alpha a \leq 1/2$.
\subsubsection{Moment bounds for $\Hnalpha{n}{1}$}
The decomposition \eqref{eq:jn_allexpansion_main} implies that
\begin{equation}
\textstyle \Hnalpha{n}{1} = -\alpha \sum_{\ell=1}^{n}\ProdBa_{\ell+1:n} \zmfuncAw[\ell] \Jnalpha{\ell-1}{1}\eqsp.
\end{equation}
Hence, using Minkowski's inequality together with \Cref{assum:IID},
\begin{equation}
\textstyle \PE^{1/p}[\normop{\Hnalpha{n}{1}}^{p}] \leq \alpha \sum_{\ell=1}^{n}  \PE^{1/p}[\normop{\ProdBa_{\ell+1:n} \zmfuncAw[\ell]}^{p}] \PE^{1/p}[\norm{\Jnalpha{\ell-1}{1}}^{p}]\eqsp.
\end{equation}
Applying \Cref{fact:exponential-stability-product} and \eqref{eq:J_n_1_fin}, we get using the definition \eqref{eq:H_n_1_bound_iid_bounded} of $\ConstD_{4}$
\begin{align}
\PE^{1/p}[\normop{\Hnalpha{n}{1}}^{p}] &\leq \qcond^{1/2} \bConst{A} \ConstD_{3} \alpha^{2} a  d^{1/q} p^{3/2} \supconsteps \sum_{\ell=1}^{n}(1 - \alpha a/4)^{n} \\
& \leq \ConstD_{4} d^{1/q} \alpha a  p^{3/2} \supconsteps\eqsp.
\end{align}

\subsection{Proof of \Cref{th:theo_1_iid} and \Cref{th:LSA_PR_error}}
\label{sec:proof:th:LSA_PR_error}
Define
\begin{equation}
\label{eq:delta_fl_p_th_moment}
\begin{split}
\Deltafl_{n,p,\alpha} &= \frac{4\rme^{1/p} \ConstD_{2} (a p)^{1/2}}{(\alpha n)^{1/2}} + \rme^{1/p} \bConst{A} (\ConstD_3 + \ConstD_4)  \alpha a p^{5/2} + \frac{2 \bConst{\sf{Rm}, 2} p}{n^{1/2}}
  \\
  & \qquad \qquad + \bConst{A} \ConstD_{1} (\alpha a)^{1/2} p^{3/2} \eqsp,\\
\Deltatr_{n,p,\alpha} &= \rme^{1/p} \qcond^{1/2} (4/(\alpha n^{1/2}) + 2^{-1/2} n^{1/2} \bConst{A})\eqsp.
\end{split}
\end{equation}

We begin with the proof of \Cref{th:LSA_PR_error}. The result of \Cref{th:theo_1_iid} will directly follow from it using the step size $\alpha$ fixed in \eqref{eq:step_size_pth_moment_refined}.

\proof{Proof of \Cref{th:LSA_PR_error}.} 
Let $q \geq 2$ be a number to be fixed later, and assume that $\alpha \in (0, \alpha_{q,\infty})$. The proof is based on exploiting the decomposition \eqref{eq:pr_err_decompose}. Below we use shorthand notations $\zmfuncAw[t], \Am_{t} ,\funnoisew_{t}$ for $\zmfuncA{\State_{t}}, \Am(\State_t)$, and $\funcnoise{\State_{t}}$, respectively. Applying \eqref{eq:pr_err_decompose} and Minkowski's inequality, we get
\begin{align}
\label{eq:bound_p_moment_iid_T_1_T_2_decomp}
  &(n/2)\PE^{1/p}\left[\norm{\bA\left(\prtheta_{n}-\thetas\right)}^{p}\right] \leq T_1 + T_2 \eqsp, \\
&        \textstyle T_1 = \PE^{1/p}\bigl[\norm{\sum\nolimits_{t=n/2}^{n-1}\rme(\theta_{t},\State_{t+1})}^{p}\bigr] \eqsp, \quad T_2 = \alpha^{-1}\PE^{1/p}[\norm{\theta_{n/2}-\theta_{n}}^p]\eqsp.
\end{align}
The term $T_2$ is a remainder one, which is controlled with \Cref{cor:LSA_err_bound_iid} and Minkowski's inequality:
\begin{align}
\label{eq:T_2_bound_iid}
T_2 \leq 2 \alpha^{-1} d^{1/q} \qcond^{1/2}\left(1 - \alpha a/4\right)^{n/2} \norm{\thetainit - \thetas} + 2 \alpha^{-1/2} d^{1/q} \ConstD_{2} (a p)^{1/2} \supconsteps \eqsp.
\end{align}
Now we proceed with the leading term $T_1$. Using Minkowski's inequality, \eqref{eq:pr_e_definition} and \eqref{eq:error_decomposition_LSA},
\begin{equation}
\label{eq:T_1_bound_iid}
\begin{aligned}
\textstyle  T_1
&\leq \textstyle \PE^{1/p}\bigl[\norm{\sum_{t=n/2}^{n-1}\funnoisew_{t+1}}^{p}\bigr] + \PE^{1/p}\parentheseDeuxLigne{\norm{\sum_{t=n/2}^{n-1} \zmfuncAw[t+1] \ProdBa_{1:t} \{ \theta_0 - \thetas \}}^p} \\
& \textstyle \,  + \PE^{1/p}\parentheseDeuxLigne{ \norm{\sum_{t=n/2}^{n-1} \zmfuncAw[t+1] \Jnalpha{t}{0}}^p} + \PE^{1/p}\parentheseDeuxLigne{\norm{\sum_{t=n/2}^{n-1} \zmfuncAw[t+1] \Jnalpha{t}{1}}^p} + \PE^{1/p}\parentheseDeuxLigne{ \norm{\sum_{t=n/2}^{n-1} \zmfuncAw[t+1] \Hnalpha{t}{1}}^p}\eqsp.
\end{aligned}
\end{equation}
We first estimate the leading term $\PE ^{1/p}\bigl[\norm{\sum_{t=n/2}^{n-1}\funnoisew_{t+1}}^{p}\bigr]$. Applying Rosenthal's inequality for martingales from \citet[Theorem~4.1]{pinelis_1994} and using that $\PE [\norm{\funnoisew(\State)}^{2}] = \trace{\noisecov}$, we get
\begin{equation}
 \PE^{1/p}\bigl[\norm{\sum_{t=n/2}^{n-1}\funnoisew_{t+1}}^{p}\bigr]
\leq \bConst{\sf{Rm}, 1} p^{1/2}(n/2)^{1/2}\{\trace{\noisecov}\}^{1/2}  +
 \bConst{\sf{Rm}, 2} p \,\PE^{1/p}[\max_{t}\norm{\funnoisew_{t+1}}^{p}]\eqsp.
\end{equation}
With the assumption \Cref{assum:noise-level}, we get from the previous bound
\begin{equation}
  \label{eq:bound_p_moment_epsi_iid}
\textstyle \PE^{1/p}\bigl[\norm{\sum_{t=n/2}^{n-1}\funnoisew_{t+1}}^{p}\bigr] \leq \bConst{\sf{Rm}, 1} p^{1/2}(n/2)^{1/2}\{\trace{\noisecov}\}^{1/2} + \bConst{\sf{Rm}, 2} \supconsteps p\eqsp.
\end{equation}
We now proceed with the other terms. The term $\sum_{t=n/2}^{n-1}\zmfuncAw[t+1] \utheta_{t}$ is upper-bounded using Minkowski's inequality and the \Cref{fact:exponential-stability-product}:
\begin{align}
     \label{eq:bound_p_moment_iid_utheta}
\textstyle \PE^{1/p}\bigl[\norm{\sum_{t=n/2}^{n-1}\zmfuncAw[t+1] \ProdBa_{1:t} \{ \theta_0 - \thetas \}}^{p}\bigr] \leq \bConst{A} (n-n_0) \qcond^{1/2} d^{1/q} (1- \alpha a/4)^{n/2} \norm{\thetainit - \thetas}\eqsp.
\end{align}
Note that the sequences $\{ \zmfuncAw[t+1] \Jnalpha{t}{0}\}_{t=n/2}^{n-1}$, $\{\zmfuncAw[t+1] \Jnalpha{t}{1}\}_{t=n/2}^{n-1}$, and $\{\zmfuncAw[t+1] \Hnalpha{t}{1}\}_{t=n/2}^{n-1}$ are $(\mcf_t)_{t\in \nset}$-martingale increments. Hence, applying the Burkholder inequality \citet[Theorem 8.6]{osekowski:2012}  and the Minkowski  inequality,
\begin{align}
\textstyle \PE^{1/p}\bigl[\norm{\sum_{t=n/2}^{n-1}\zmfuncAw[t+1]\Hnalpha{t}{1}}^{p}\bigr]
  & \textstyle \leq  p \Bigl(\sum_{t=n/2}^{n-1}\PE^{2/p}[\norm{\zmfuncAw[t+1] \Hnalpha{t}{1}}^p]\Bigr)^{1/2} \\
     \label{eq:bound_p_moment_iid_H_1}
&\textstyle \leq \bConst{A} \ConstD_{4} (n/2)^{1/2} p^{5/2} \alpha a d^{1/q} \supconsteps\eqsp,
\end{align}
where the last inequality follows from \Cref{prop:J_n_1_2_bound_iid}. Similarly, using \Cref{prop:J_n_0_bound_iid} and \Cref{prop:J_n_1_2_bound_iid}, we get
\begin{align}
     \label{eq:bound_p_moment_iid_J_0}
\textstyle\PE^{1/p}[\norm{\sum_{t=n/2}^{n-1} \zmfuncAw[t+1] \Jnalpha{t}{0}}^p ]
&\textstyle \leq  p \Bigl(\sum_{t=n/2}^{n-1}\PE^{2/p}[\norm{\zmfuncAw[t+1] \Jnalpha{t}{0}}^p]\Bigr)^{1/2} \\
&\textstyle \leq   \bConst{A} \ConstD_{1} (n/2)^{1/2} p^{3/2} (\alpha a)^{1/2} \supconsteps\eqsp.
\end{align}
 By the same reasoning, with \Cref{prop:J_n_1_2_bound_iid}, we get
 \begin{equation}
   \label{eq:bound_p_moment_iid_J_1}
\textstyle \PE^{1/p}[\norm{\sum_{t=n/2}^{n-1}\zmfuncAw[t+1] \Jnalpha{t}{1}}^p ] \leq  \bConst{A} \ConstD_{3} (n/2)^{1/2} p^{5/2} \alpha  a \supconsteps\eqsp.
\end{equation}
It remains to choose $q = p(1+\log{d})$ and combine the bounds above in \eqref{eq:bound_p_moment_iid_T_1_T_2_decomp}.

\subsection{Version of \Cref{th:theo_1_iid} and \Cref{cor:hp_bound_pr_iid} with exact constants}
\label{sec:proof:LSA_PR_error_iid_consts}
\begin{corollary}
\label{prop:hp_iid_exact}
Assume \Cref{assum:A-b}, \Cref{assum:noise-level}, \Cref{assum:IID} and let $n \geq 2$, $p \geq 2$ and consider the step size $\alpha = \alpha(n,d,p)$ specified in \eqref{eq:step_size_pth_moment_refined}. Then it holds that
\begin{align}
\label{eq:bound_p_th_moment_optimized_iid_with_const}
&(n/2)^{1/2}\PE^{1/p}\left[\norm{\bA\left(\prtheta_{n}-\thetas\right)}^{p}\right]
\leq \bConst{\sf{Rm}, 1} \{\trace{\noisecov}\}^{1/2} p^{1/2} \\ &\qquad \qquad \qquad + \rme^{1/p} \supconsteps \left(\frac{c_3 (1+\log{d})^{1/2} p}{n^{1/4}} + \frac{c_4 p}{n^{1/2}}\right) \\
&\qquad \qquad \qquad+ \rme^{1/p} c_{5} (1+\log{d})(p+n^{1/2})\norm{\thetainit - \thetas} \exp\left\{ - \frac{( \alpha_{\infty} \wedge \smallAconst) \sqrt{n}}{8 p (1+\log{d})} \right\}\eqsp,
\end{align}
where $c_3$, $c_4$ and $c_5$ are given by
\begin{align}
c_{3} &= \frac{4 a^{1/2}\ConstD_2}{(\alpha_{\infty} \wedge
\smallAconst)^{1/2}} + 2 \bConst{\sf{Rm}, 2} + (\alpha_{\infty} \wedge
\smallAconst)^{1/2} a^{1/2}\bConst{A} \ConstD_{1}\eqsp, \, c_{5} = \qcond^{1/2} \left(\frac{4}{\alpha_{\infty} \wedge \smallAconst} + \bConst{A}\right)\eqsp, \\
c_4 &= \bConst{A} (\ConstD_3 + \ConstD_4) a ( \alpha_{\infty} \wedge \smallAconst)\eqsp.
\end{align}
Moreover, let us fix $\delta \in \ooint{0,1}$. Then for any $\thetainit \in \rset^{d}$, $n \in \nset$, with $\alpha = \alpha(n,d,\log(3\rme/\delta))$ defined in \eqref{eq:step_size_pth_moment_refined}, it holds with probability at least $1-\delta$, that
\begin{equation}
n^{1/2} \norm{\bA\left(\prtheta_{n}-\thetas\right)} \leq  3\rme\sqrt{2} \sqrt{\{\trace{\noisecov}\} \log (3\rme/\delta)}
+ c_{2} \Delta^{(\HP)}(n,\thetainit, \delta)\eqsp,
\end{equation}
where
\begin{equation}
\label{eq:aux_const_c_2_iid}
c_{2} = 3\rme \sqrt{2} \left((c_3 + c_4) (1+\log{d})^{1/2} \vee c_5 (1+\log{d})\right)\eqsp.
\end{equation}
\end{corollary}

\proof{Proof.}
The inequality \eqref{eq:bound_p_th_moment_optimized_iid_with_const} follows from \Cref{th:LSA_PR_error} after substituting the step size $\alpha(n,d,p)$ given in \eqref{eq:step_size_pth_moment_refined}. Now \Cref{cor:hp_bound_pr_iid} with $c_2$ defined in \eqref{eq:aux_const_c_2_iid} follows from the Markov inequality applied with $p = \log(3\rme/\delta) > 2$.
\endproof


%% file: appendix_iid_subgaussian.tex
Assumption \Cref{assum:noise-level} can be relaxed to a sub-Gaussian-type conditions on the noise variable $\funnoisew(\State)$. Consider the following assumption:
\begin{assum}
\label{assum:effective-noise-level-subgaus}
For any $u \in \sphere^{d-1}$, and $\lambda \in \rset$, $\log\{\PE[ \exp(\lambda u^{\top} \funcnoise{\State})]\} \leq \lambda^2 \sigma_\funnoisew^{2} / 2$, where $\State$ is a random variable with distribution $\invariantQ$.
\end{assum}
Note that \Cref{assum:noise-level} implies \Cref{assum:effective-noise-level-subgaus}, and \Cref{assum:effective-noise-level-subgaus} can be written more concisely as  $u^{\top} \funcnoise{\State} \in \SG (\sigma_\funnoisew^2)$ for any $u \in \sphere^{d-1}$. For instance, this condition holds when $\funcnoise{Z_{t+1}}$ is an outer product of sub-Gaussian random variables in the canonical coordinates; see \citet[Assumption~2]{mou2021optimal}\footnote{The condition can be further relaxed to cover heavier-tail setting in which $\funcnoise{Z_{t+1}}$ has only a finite number of moments or is sub-exponential (instead of sub-gaussian).}. Note that, for any $u\in \sphere^{d-1}$, and $t \geq 0$,
\begin{equation}
  \label{eq:tail_sub_gaussian}
  \PP(\abs{u^\top \funcnoise{\State}} \geq t) \leq 2 \exp(-t^2/(2\sigma_{\funnoisew}^2))\eqsp.
\end{equation}

Below we state the counterpart of \Cref{prop:J_n_0_bound_iid} and \Cref{cor:LSA_err_bound_iid}.
\begin{proposition}
\label{prop:J_n_0_bound_iid_subgaus}
Assume \Cref{assum:A-b}, \Cref{assum:IID} and \Cref{assum:effective-noise-level-subgaus}. Then, for any $\alpha \in \ocint{0,\alpha_{\infty}}$, $p \geq 2$, $u\in\sphere^d$ and $n \in \nset$,
\begin{equation}
\label{eq:j_n_0_bound_subgauss}
\PE^{1/p}\bigl[\bigl\vert u^{\top} \Jnalpha{n}{0}\bigr\vert^{p}\bigr] \leq \ConstD_{1} \sqrt{\alpha a p \sigma_{\funnoisew}^{2}}\eqsp,
\end{equation}
where $\ConstD_{1}$ is given in \eqref{eq:J_n_0_bound_iid}. Moreover, for any $p,q \in \nset$, $2 \leq p \leq q$, $\alpha \in \ocint{0, \alpha_{q,\infty}}$, $n \in \nset$,  $u\in\sphere^d$ and $\thetainit \in \rset^d$,
\begin{align}
\label{eq:n_step_norm_propery_subgauss}
\PE^{1/p}\left[\abs{u^{\top}(\theta_n - \thetas)}^{p}\right]
\leq d^{1/q} \qcond^{1/2} \left(1 - \alpha a/4\right)^{n} \norm{\thetainit - \thetas} + \ConstD d^{1/q} \sqrt{\alpha a p \sigma^2_{\funnoisew}} \eqsp,
\end{align}
where the constant $\ConstD$ is given by
\begin{equation}
\label{eq:definition-D-subgaussian}
\ConstD =(2 \qcond)^{1/2} a^{-1}(1 + 4 \qcond^{1/2} \bConst{A}  a^{-1})\eqsp
\end{equation}
\end{proposition}
\proof{Proof.}
We first show the bound \eqref{eq:j_n_0_bound_subgauss}. Expanding \eqref{eq:jn0_main}, we get for any $u \in \sphere^{d-1}$, that
\begin{equation}
\label{eq:j_n_alpha_expr}
\textstyle
u^{\top} \Jnalpha{n}{0} = \alpha \sum_{j=1}^{n}\eta_{n,j} \eqsp, \text{ where }
\eta_{n,j} = \, u^{\top}\left(\Id - \alpha \bA \right)^{n-j}\funnoisew_{j}\eqsp.
\end{equation}
Note that $\{ \eta_{n,j} \}_{j=1}^n$ are sub-Gaussian random variables. With \Cref{fact:Hurwitzstability}, for any $\lambda \in \rset$,
\begin{equation}
\textstyle \log \PE [\exp\{\lambda \eta_{n,j}\}] \leq (1/2)\lambda^2 \norm{u^{\top}\left(\Id - \alpha \bA \right)^{n-j}}^{2} \sigma_{\funnoisew}^{2} \leq (1/2) \lambda^2 \qcond (1-\alpha a)^{n-j} \sigma_{\funnoisew}^{2}\eqsp.
\end{equation}
Hence, $\eta_{n,j} \in \SG(\sigma_{n,j}^2)$, where
$\sigma_{n,j}^2 = \qcond(1-\alpha a)^{n-j}\sigma_{\funnoisew}^2$. \Cref{assum:IID} and \eqref{eq:j_n_alpha_expr} imply that $u^\top \Jnalpha{n}{0}$ is also sub-Gaussian random variable, that is,
\begin{equation}
\textstyle
u^\top \Jnalpha{n}{0} \in \SG(\sigma_{\alpha,n}^2)\eqsp, \quad \sigma_{\alpha,n}^2 = \alpha^2 \sum_{j=1}^{n}\sigma_{n,j}^2 \leq a^{-1}\qcond \sigma_{\funnoisew}^2 \alpha\eqsp.
\end{equation}
Using \eqref{eq:tail_sub_gaussian} and applying \Cref{lem:bound_subgaussian}, we obtain for $p \geq 2$ that
\begin{equation}
\label{eq:j_n_alpha_bound}
\PE^{1/p}\bigl[\bigl\vert u^{\top} \Jnalpha{n}{0}\bigr\vert^{p}\bigr] \leq \ConstD_{1} \sigma_{\funnoisew} \sqrt{\alpha a p}\eqsp, \text{ where } \ConstD_1 =
2 \qcond^{1/2}a^{-1}\eqsp.
\end{equation}
Now the proof of the bound \eqref{eq:n_step_norm_propery_subgauss}  follows the same line as the proof of \Cref{cor:LSA_err_bound_iid} and is omitted.
\endproof

\begin{proposition}
\label{prop:mse_iid_with_burn_in_subgaussian}
Assume \Cref{assum:A-b}, \Cref{assum:IID} and \Cref{assum:effective-noise-level-subgaus}. Then, for any even $n \geq 2$, $\alpha \in (0, \alpha_{\infty} \wedge[ \smallAconst/\{2 + 2\log{d}\}])$,  $\thetainit \in \rset^d$, $u \in\sphere^{d-1}$, it holds
\[
(n/2) \PE\left[\abs{u^{\top}\bA\left(\prtheta_{n}-\thetas\right)}^{2}\right]
 \leq 4 u^{\top} \noisecov u + \Deltafl_{n,\alpha} \sigma_{\funnoisew} + \rme^{-\alpha a n/4} \Deltatr_{n,\alpha} \norm{\thetainit - \thetas}^{2}\eqsp,
\]
where $\Deltafl_{n,\alpha},\Deltatr _{n,\alpha}$ are given in \eqref{eq:delta_transient_fluctuation_terms}.
\end{proposition}
\proof{Proof.}
The proof follows the same line as \Cref{prop:mse_iid_with_burn_in}.
\endproof
\begin{proposition}
\label{cor:J_n_1_2_bound_subgaussian_noise}
Assume \Cref{assum:A-b}, \Cref{assum:effective-noise-level-subgaus}, and \Cref{assum:IID}. Then, for any $\alpha \in \ocint{0,\alpha_{\infty}}$, $p \geq 2$, $ u \in \sphere^{d-1}$ and $n \in \nset$, it holds
\begin{equation}
\label{eq:J_n_1_bound_subgaussian_noise}
\PE^{1/p}\bigl[|u^\top\Jnalpha{n}{1}|^{p}\bigr] \leq \ConstD_3 \alpha a p^{2} \sigma_{\funnoisew}\eqsp, \text{ where }  \ConstD_3 =  4 \qcond  \bConst{A}  /a^{2} \eqsp.
\end{equation}
 Moreover, for any $2 \leq p \leq q$ and $\alpha \in \ocint{0, \alpha_{q,\infty}}$, $n \in \nset$,
\begin{equation}
\label{eq:H_n_1_bound_subgaussian_noise}
\PE^{1/p}\bigl[|u^\top\Hnalpha{n}{1}|^{p}\bigr] \leq \ConstD_{4} \alpha a p^{2} d^{1/q} \sigma_{\funnoisew}\eqsp, \text{ where } \ConstD_{4} = 4 \qcond^{1/2} \bConst{A} \ConstD_{3}/a^{2} \eqsp.
\end{equation}
\end{proposition}
\proof{Proof.}
We begin with the bound \eqref{eq:J_n_1_bound_subgaussian_noise}.
We use \eqref{eq:S_ell_def}. The sequence $\{ S^{(1)}_{i+1:n} \funnoisew_{i}\}_{i=1}^{n-1}$ is a martingale-difference with respect to the filtration $(\mcf_k)_{k \in\nset}$:   Applying Burkholder's inequality \citet[Theorem 8.6]{osekowski:2012} and Minkowski's inequality, we get
\begin{align}
\label{eq:pinelis_J_n_1}
\textstyle \PE[|u^\top \Jnalpha{n}{1}|^p ]
  &\leq \textstyle  p^p \alpha^{2p} \PE [(\sum_{i=1}^{n-1} (u^\top S^{(1)}_{i+1:n} \funnoisew_{i} )^2 )^{p/2}] \\
\textstyle  & \leq  \textstyle p^{p} \alpha^{2p} ( \sum_{i=1}^{n-1}  \PE^{2/p}[|u^\top S^{(1)}_{i+1:n} \funnoisew_{i}|^p])^{p/2} \eqsp.
\end{align}
Set $v_{i+1:n}= [S_{i+1:n}^{(1)}]^T u$.  Then, using \Cref{assum:IID}, we get
\begin{align}
\label{eq:cond_expectation_inserted}
\PE[|v_{i+1:n}^\top \funnoisew_{i}|^p] \leq \PE\bigl[ \| v_{i+1:n} \|^p \bigr] \sup_{v \in \sphere^{d-1}} \PE[| v^\top \funnoisew_i|^p ]\eqsp.
\end{align}
Using the same arguments as in \Cref{prop:J_n_1_2_bound_iid}, we get
\begin{align}
\PP\bigl(\norm{v_{i+1:n}} \geq t\bigr) \leq 2\exp\left\{-\frac{t^2}{2\qcond^2\bConst{A}^2(n-i)(1-\alpha a)^{n-i-1}}\right\}\eqsp.
\end{align}
Hence, applying \Cref{lem:bound_subgaussian}, we get for any  $u \in \sphere^{d-1}$
\begin{equation}
\label{eq:l_p_bound_V_ell}
\PE^{2/p}[\norm{v_{i+1:n}}^p] \leq 4 p \bConst{A}^2\qcond^2(n-i)(1-\alpha a)^{n-i-1} \eqsp.
\end{equation}
Combining \eqref{eq:pinelis_J_n_1}, \eqref{eq:l_p_bound_V_ell}, and \Cref{assum:noise-level}, we get
\begin{equation}
\label{eq:J_n_1_fin_subgaus}
\begin{split}
\PE^{1/p}[|u^\top \Jnalpha{n}{1}|^p]
& \textstyle \leq 4  \sigma_{\funnoisew} p^{2} \alpha^{2}  \bConst{A} \qcond ( \sum_{i=1}^{n-1} (n-i) (1-\alpha a)^{n-i-1} )^{1/2} \\
&\leq \ConstD_3 \alpha a p^{2} \sigma_\funnoisew\eqsp.
\end{split}
\end{equation}
We now consider \eqref{eq:H_n_1_bound_subgaussian_noise}.  Recall that $\Hnalpha{n}{1} = -\alpha \sum_{\ell=1}^{n}\ProdBa_{\ell+1:n} \zmfuncAw[\ell] \Jnalpha{\ell-1}{1}$.
Hence, using Minkowski's inequality together with \Cref{assum:IID},
\begin{align}
\textstyle
\PE^{1/p}[\normop{\Hnalpha{n}{1}}^{p}] \leq \alpha \sum_{\ell=1}^{n}  \PE^{1/p}[\normop{\ProdBa_{\ell+1:n} \zmfuncAw[\ell]}^{p}] \sup_{u \in \sphere^{d-1}} \PE^{1/p}[| u^\top \Jnalpha{\ell-1}{1}|^{p}]\eqsp.
\end{align}
Applying \Cref{fact:exponential-stability-product} and \eqref{eq:J_n_1_fin_subgaus}, we get using the definition \eqref{eq:H_n_1_bound_iid_bounded} of $\ConstD_{4}$
\begin{align}
\textstyle
\PE^{1/p}[\normop{\Hnalpha{n}{1}}^{p}] \leq \qcond^{1/2}  d^{1/q} \bConst{A} \ConstD_{3} \alpha^2   \sigma_{\funnoisew} p^{2} \sum_{\ell=1}^{n}(1 - \alpha a/4)^{n} \leq \ConstD_{4} d^{1/q} \sigma_{\funnoisew} \alpha a  p^{2}\eqsp.
\end{align}
\endproof
Using the bounds of \Cref{prop:mse_iid_with_burn_in_subgaussian}, we obtain the $p$-th moment error bound for LSA-PR procedure similarly to \Cref{th:LSA_PR_error}. Proceeding as in \eqref{eq:delta_fl_p_th_moment}, we introduce the fluctuation and transient components of the LSA-PR error
\begin{equation}
\label{eq:delta_fl_p_th_moment_appendix}
\begin{split}
\Deltafl_{n,p,\alpha} &= \frac{4\rme^{1/p} \ConstD_{2} p^{1/2}}{(\alpha n)^{1/2}} + \rme^{1/p} \bConst{A} (\ConstD_3 + \ConstD_4)  \alpha a p^{3} + \frac{3\sqrt{2} \bConst{\sf{Rm}, 2} \sqrt{\log\{\rme n\}} p^{3/2}}{n^{1/2}}
  \\
  & \qquad \qquad + \bConst{A} \ConstD_{1} \alpha^{1/2} p^{3/2} \eqsp,\\
\Deltatr_{n,p,\alpha} &= \rme^{1/p} \qcond^{1/2} (2\sqrt{2}/(\alpha n^{1/2}) + 2^{-1/2} n^{1/2} \bConst{A})\eqsp.
\end{split}
\end{equation}

\begin{theorem}
\label{th:LSA_PR_error_subgaus}
Assume Assume \Cref{assum:A-b}, \Cref{assum:IID}, and \Cref{assum:effective-noise-level-subgaus}. Then, for any even $n \geq 2$, $p \geq 2$, $\alpha \in (0, \alpha_{\infty} \wedge \smallAconst/\{p(1+\log{d})\})$, $\thetainit \in \rset^d$, $u \in \sphere_{d-1}$, it holds
\begin{multline}
\label{eq:p_th_moment_LSA_bound_subgaus}
(n/2)^{1/2}\PE^{1/p}\left[\abs{u^{\top}\bA\left(\prtheta_{n}-\thetas\right)}^{p}\right]
\leq  \bConst{\sf{Rm}, 1} \{u^{\top} \noisecov u\}^{1/2} p^{1/2} + \sigma_{\funnoisew} \Deltafl_{n,p,\alpha} \\
+ \Deltatr_{n,p,\alpha} \left(1 - \alpha a/4\right)^{n/2} \norm{\thetainit - \thetas}\eqsp,
\end{multline}
where $\bConst{\sf{Rm}, i}$, $i=1,2$ are defined in \Cref{appendix:constants}.
\end{theorem}
\proof{Proof.}
The proof follows the lines of \Cref{th:LSA_PR_error} and is omitted. The only difference with the mentioned proof is related with the term $\PE^{1/p}\bigl[\abs{\sum_{t=n/2}^{n-1}u^{\top} \funnoisew_{t+1}}^{p}\bigr]$. Application of Rosenthal's inequality yields
\begin{multline}
\textstyle \PE^{1/p}\bigl[\abs{\sum_{t=n/2}^{n-1}u^{\top} \funnoisew_{t+1}}^{p}\bigr] \leq \bConst{\sf{Rm}, 1} p^{1/2}(n-n_0)^{1/2}\{u^{\top} \noisecov u\}^{1/2} \\ +
\textstyle  \bConst{\sf{Rm}, 2} p \,\PE^{1/p}[\max_{t \in \{n/2,\ldots,n-1\}}\abs{u^{\top} \funnoisew_{t+1}}^{p}]\eqsp.
\end{multline}
Since $u^{\top} \funnoisew_{t+1} \in \SG(\sigma^{2}_{\funnoisew})$ for any $t \in \nsets$, we obtain using \citet[Lemma~4]{durmus2021tight}, that
\begin{equation}
\textstyle \PE^{1/p}[\max_{t \in \{n/2,\ldots,n-1\}}\abs{u^{\top} \funnoisew_{t+1}}^{p}] \leq 3 \sigma_{\funnoisew} p^{1/2} \sqrt{1 + \log(n/2)}\eqsp.
\end{equation}
This modification affects the fluctuation term $\Deltafl_{n,p,\alpha}$ in \eqref{eq:delta_fl_p_th_moment_appendix}.
\endproof

%% file: appendix_markov.tex
\subsection{Proof of \Cref{prop:products_of_matrices_UGE}}
\label{sec:matrix_product_matrix}
We first provide a  result on the product of  dependent  random matrices. The proof is based on  \citet{huang2020matrix}. Let $(\Omega, \mathfrak F, \sequence{\mathfrak{F}}[\ell][\nset], \P)$ be a filtered probability space. For the matrix $\MatB \in \rset^{d \times d}$ we denote by $( \sigma_\ell(\MatB) )_{ \ell=1 }^d$ its singular values. For $\qexponent \geq 1$, the Shatten $\qexponent$-norm is denoted by $\norm{\MatB}[\qexponent] = \{\sum_{\ell=1}^d \sigma_\ell^\qexponent (\MatB)\}^{1/\qexponent}$. For $\qexponent, \ppexponent \geq 1$ and a random matrix $\X$ we write $\norm{\X}[\qexponent,\ppexponent] = \{ \PE[\norm{\X}[\qexponent]^\ppexponent] \}^{1/\ppexponent}$.
\par 

\begin{proposition}
\label{th:general_expectation UGE}
Let $\sequence{\Y}[\ell][\nset]$ be a sequence of random matrices adapted to the filtration \(\sequence{\mathfrak{F}}[\ell][\nset]\)   and $\MatP$ be a positive definite matrix. Assume that for each $\ell \in \nsets$ there exist $\mtt_{\ell} \in \ocintLine{0,1}$  and $\sigma_{\ell} > 0$ such that
\begin{equation}
  \label{eq:form_bound_ass_prod_mat}
  \norm{\PE^{\mathfrak{F}_{\ell-1}}[\Y_\ell]}[\MatP]^2  \leq 1 - \mtt_{\ell} \text{  and } \norm{\Y_\ell - \PE^{\mathfrak{F}_{\ell-1}}[\Y_\ell]}[\MatP]  \leq \sigma_\ell \quad \text{ $\PP$-a.s.}  \eqsp.
\end{equation}
Define $\Zbf_n = \prod_{\ell = 0}^n \Y_\ell= \Y_n \Zbf_{n-1}$, for $n \geq 1$. Then, for any $2 \le \ppexponent \le \qexponent$ and $n \geq 1$,
\begin{equation}
\label{eq:gen_expectation}
\textstyle \norm{\Zbf_n}[\qexponent,\ppexponent]^2 \leq \kappa_P \prod_{\ell=1}^n (1- \mtt_{\ell} + (\qexponent-1)\sigma_{\ell}^2) \norm{\MatP^{1/2}\Zbf_0 \MatP^{-1/2}}[\qexponent, \ppexponent]^2 \eqsp,
\end{equation}
where $\kappa_P = \lambda_{\sf max}( \MatP )/\lambda_{\sf min}( \MatP )$ and $\lambda_{\sf max}( \MatP ),\lambda_{\sf min}( \MatP )$ correspond to the largest and smallest eigenvalues of $\MatP$.
\end{proposition}
\proof{Proof.}
Let $n \in\nsets$ and $2 \leq \ppexponent \leq \qexponent$. We begin with the decomposition
\begin{equation}
\Zbf_n = \Y_n \Zbf_{n-1} = (\Y_n - \CPE{\Y_n}{\mathfrak{F}_{n-1}}) \Zbf_{n-1}+ \CPE{\Y_n}{\mathfrak{F}_{n-1}} \Zbf_{n-1}\eqsp.
\end{equation}
Let us define $f_P : \rset^{d \times d } \to \rset^{d \times d}$ as $f_P(B) = \MatP^{1/2} B \MatP^{-1/2}$. Therefore, for any $n \in\nset$, it holds $f_P(\Zbf_n) = \bfA_n + \bfB_n$, where
\begin{equation}
\bfA_n = f_P((\Y_n - \CPE{\Y_n}{\mathfrak{F}_{n-1}}) \Zbf_{n-1}) \eqsp, \quad \Bbf_n = f_P(\CPE{\Y_n}{\mathfrak{F}_{n-1}}) f_P( \Zbf_{n-1} )\eqsp.
\end{equation}
Since \(\CPE{\bfA_n}{\bfB_n} = \CPE{\CPE{\bfA_n}{\mathfrak{F}_{n-1}}}{\bfB_n} = 0\),  \citet[Proposition 4.3]{huang2020matrix} implies that
\begin{equation}
\label{eq:bound_decomp_proof_mat_prod}
\norm{f_P(\Zbf_n)}[\qexponent,\ppexponent]^2 \leq \norm{\bfB_n}[\qexponent,\ppexponent]^2 + (\qexponent-1) \norm{\bfA_n}[\qexponent,\ppexponent]^2 \eqsp.
\end{equation}
It remains to bound the two terms on the right-hand side. To this end, we use \citet[Theorem~6.20]{hiai:petz:2014} which implies that for any $B_1,B_2 \in \rset^{d \times d}$,
\begin{equation}
\label{eq:submultip}
\norm{B_1 B_2}[\qexponent,\ppexponent]\leq \normop{B_1} \norm{B_2}[\qexponent,\ppexponent]  \eqsp.
\end{equation}
Combining \eqref{eq:submultip} with $\norm{B}[\MatP] = \normop{f_P(B)}$, and $\norm{\Y_n - \CPE{\Y_n}{\mathfrak{F}_{n-1}}}[\MatP]  \leq \sigma_{n}$, we get
\begin{align}
\nonumber
\norm{\bfA_n}[\qexponent,\ppexponent] &= \left(\PE\left[ \norm{ f_P(\Y_n - \CPE{\Y_n}{\mathfrak{F}_{n-1}})  f_P(\Zbf_{n-1})}[\qexponent]^\ppexponent \right]\right)^{1/\ppexponent} \\
\label{eq:bound_MP_1_proof}
&\leq  \left(\PE\left[ \normop{\Y_n - \CPE{\Y_n}{\mathfrak{F}_{n-1}}}[\MatP]^\ppexponent  \norm{f_P(\Zbf_{n-1})}[\qexponent]^\ppexponent  \right]\right)^{1/\ppexponent} \leq\sigma_{n} \norm{ f_P(\Zbf_{n-1})}[\qexponent,\ppexponent] \eqsp.
\end{align}
Similarly, applying $\norm{\CPE{\Y_n}{\mathfrak{F}_{n-1}}}[\MatP]^2  \leq 1 - \mtt_n$
\begin{align}
\nonumber
\norm{\bfB_n}[\qexponent,\ppexponent]^2 &= \left(\PE\left[ \norm{ f_P(\CPE{\Y_n}{\mathfrak{F}_{n-1}})  f_P(\Zbf_{n-1})}[\qexponent]^\ppexponent \right]\right)^{2/\ppexponent} \\
\label{eq:bound_MP_2_proof}
&\leq \left(\PE\left[ \norm{\CPE{\Y_n}{\mathfrak{F}_{n-1}}}[\MatP]^\ppexponent \norm{f_P(\Zbf_{n-1})}[\qexponent]^\ppexponent \right]\right)^{2/\ppexponent} \leq (1 - \mtt_n) \norm{f_P(\Zbf_{n-1})}[\qexponent,\ppexponent]^2 \eqsp.
\end{align}
Combining \eqref{eq:bound_MP_1_proof} and \eqref{eq:bound_MP_2_proof} in \eqref{eq:bound_decomp_proof_mat_prod} yields
\begin{equation}
  \label{eq:2}
\textstyle \norm{f_P(\Zbf_n)}[\qexponent,\ppexponent]^2 \leq (1 - \mtt_n + (\qexponent-1)\sigma_{n}^2)  \norm{f_P(\Zbf_{n-1})}[\qexponent,\ppexponent]^2 \leq \prod_{i=1}^n (1 - \mtt_i + (\qexponent-1)\sigma_{i}^2)  \norm{f_P(\Zbf_{0})}[\qexponent,\ppexponent]^2 \eqsp.
\end{equation}
The proof is completed using \eqref{eq:submultip} which implies that
\[
\norm{\Zbf_n}[\qexponent,\ppexponent]= \norm{\MatP^{-1/2}f_P(\Zbf_n) \MatP^{1/2}}[\qexponent,\ppexponent] \leq \sqrt{\kappa_P} \norm{f_P(\Zbf_n)}[\qexponent,\ppexponent] \eqsp.
\]
\endproof
In the lemmas below we aim to prove the bound \eqref{eq:form_bound_ass_prod_mat_v2_Z}. Recall that $\Y_1 = \prod_{i=1}^{h} (\Id - \alpha
\funcA{\State_{i}})$.

\begin{lemma}
\label{lem:proof_uge_1}
Assume \Cref{assum:A-b} and \Cref{assum:drift}. Then for any $\alpha \in \ocintLine{0, \alpha_\infty^{(\Markov)}\taumix^{-1}}$ with $\alpha_\infty^{(\Markov)}$ defined in \eqref{eq:alpha_infty_makov}, and any probability $\xi$ on $(\Zset,\Zsigma)$,
\begin{equation}
\label{eq:h_markov_def}
\norm{\PE_{\xi}[\Y_1]}[Q]^2 \leq 1 -  a \alpha h/6\eqsp, \quad \text{where } h = 1 \vee \lceil 8 \qcond^{1/2} \bConst{A} \taumix / a \rceil \eqsp.
\end{equation}
\end{lemma}
\proof{Proof.}
We decompose the matrix product $\Y_1$ as follows:
\begin{equation} \label{eq:split_main}
\Y_1  = \Id - \alpha h \bA   - \Mat{S}_1  + \Mat{R}_1 \eqsp,
\end{equation}
where $\Mat{S}_1 = \alpha \sum_{k = 1}^{h}\bigl\{\funcA{\State_{k}} - \bA\bigr\}$ is  linear statistics in $\{\funcA{\State_{k}}\}_{k=1}^h$, and the remainder $\Mat{R}_1$ collects the higher-order terms in the products

\begin{equation} \label{eq:RlRlbar_def}
    \Mat{R}_1 = \sum_{r=2}^{h}(-1)^{r} \alpha^{r} \sum_{(i_1,\dots,i_r)\in\msi_r^\ell}\prod_{u=1}^{r}\funcA{\State_{i_u}}\eqsp.
\end{equation}
with $\msi_r^{\ell} = \{(i_1,\ldots,i_r) \in \{1,\ldots,h\}^r\, : \, i_1 < \cdots < i_r \}$. Using $\norm{M}[Q] = \norm{Q^{1/2} M Q^{-1/2}}$, it is straightforward to check that $\PP$-a.s. it holds
\begin{equation}
\label{eq:10}
\norm{\Mat{R}_{1}}[Q] \leq \sum_{r=2}^{h} (\alpha \qcond^{1/2} \bConst{A} )^r \binom{h}{r}  \leq (\qcond^{1/2} \bConst{A} \alpha h)^2  (1 + \qcond^{1/2} \bConst{A} \alpha)^{h} = T_2 \eqsp.
\end{equation}
On the other hand, using \Cref{assum:drift}, we have for any $k \in \nsets$, that
\begin{align}
\normop{\PE_{\xi}[\funcA{\State_{k}} - \bA]} = \sup_{u,v \in \sphere^{d-1}}[\PE_{\xi}[u^{\top} \funcA{\State_{k}} v] - u^{\top} \bA v] \leq \bConst{A} \dobru{\MKQ^k}\eqsp.
\end{align}
Hence, with the triangle inequality and \eqref{eq:crr_koef_sum_tau_mix},
\begin{align}
\norm{\PE_{\xi}[\Mat{S}_1]}[Q]
&\leq \textstyle \alpha \qcond^{1/2} \sum_{k = 1}^{h} \normop{\PE_{\xi}[\funcA{\State_{k}} - \bA]} \leq \alpha \qcond^{1/2} \bConst{A} \sum_{k = 1}^{h} \dobru{\MKQ^k} \\
&\leq \textstyle (4/3) \alpha  \taumix \qcond^{1/2} \bConst{A} = T_1\eqsp.
\end{align}
This result combined with \eqref{eq:10} in \eqref{eq:split_main} implies that
\begin{equation}
\label{eq: T_2 def markov product}
\norm{\PE_{\xi}[ \Y_{1}]}[Q] \leq \norm{\Id -  \alpha h \bA}[Q] + T_1 + T_2 \eqsp.
\end{equation}
First, by definition  \eqref{eq:h_markov_def} of $h$ , we have
\begin{equation}
\label{eq:bound_T_1_product_uge}
  T_1 \leq \alpha a h / 6 \eqsp.
\end{equation}
With the definition of $\alpha_\infty^{(\Markov)}$ in \eqref{eq:alpha_infty_makov}, $\alpha \leq \alpha_\infty^{(\Markov)} \leq  (\qcond^{1/2} \bConst{A} h)^{-1} \wedge [ a/(6\rme \qcond \bConst{A}^2 h)]$, and
\begin{equation}
  \label{eq:bound_T_2_product_uge}
T_2 \le (\qcond^{1/2} \bConst{A}  \alpha h)^2 \rme \leq \alpha ah/6 \eqsp.
\end{equation}
Finally, \Cref{fact:Hurwitzstability} implies that, for $\alpha h \leq \alpha_{\infty}$,
\begin{equation}
\label{eq:bound_main_uge}
\textstyle \norm{\Id -  \alpha h \bA}[Q] \leq 1 - \alpha a h/2\eqsp.
\end{equation}
Combining \eqref{eq:bound_T_1_product_uge}, \eqref{eq:bound_T_2_product_uge}, and \eqref{eq:bound_main_uge} yield $\norm{\PE_{\xi}[ \Y_{1}]}[Q] \leq 1 - \alpha a h/6$, and the statement follows.
\endproof

\begin{lemma}
\label{lem:proof_uge_2}
Assume \Cref{assum:A-b} and \Cref{assum:drift}, and let $\alpha \in \ocintLine{0, \alpha_\infty^{(\Markov)}\taumix^{-1}}$. Then, for any probability $\xi$ on $(\Zset,\Zsigma)$, we have
\begin{equation}
\label{eq:norm_bound_h_lemma}
\norm{\Y_{1} - \PE_{\xi}[ \Y_{1}]}[Q] \leq \bConst{\sigma} \alpha h \eqsp, \text{ where } \bConst{\sigma} = 2(\qcond^{1/2}\bConst{A} + a/6)\eqsp,
\end{equation}
and $h$ is given in \eqref{eq:h_markov_def}.
\end{lemma}
\proof{Proof.}
Using \eqref{eq:split_main}, we obtain
\begin{equation}
\textstyle
\norm{\Y_{1} - \PE_{\xi}[ \Y_{1}]}[Q] \leq \alpha \sum_{k = 1}^{h} \norm{\funcA{\State_k} - \PE_{\xi}[\funcA{\State_k}]}[Q] + \norm{\Mat{R}_{1} - \PE_{\xi}[\Mat{R}_{1}]}[Q]\eqsp.
\end{equation}
Applying the definition of $\Mat{R}_{1}$ in \eqref{eq:10}, the definition of $h$,$\alpha_\infty^{(\Markov)}$, and $T_2$ in \eqref{eq:bound_T_2_product_uge}, we get from the above inequalities
\begin{equation}
\textstyle \norm{\Y_{1} - \PE_{\xi}[ \Y_{1}]}[Q] \leq 2\alpha \qcond^{1/2}\bConst{A}h + \alpha a h / 3\eqsp,
\end{equation}
and the statement follows.
\endproof
\par
We have now all we need to show 
\Cref{prop:products_of_matrices_UGE}.

\proof{Proof of \Cref{prop:products_of_matrices_UGE}}
Denote by $h \in \nset$ a block length, the value of which is determined later. Define the sequence $j_0 = 0, \, j_{\ell+1} = \min(j_\ell + h, n)$. By construction $j_{\ell+1} - j_{\ell} \leq h$. Let $N = \ceil{n/h}$. Now we introduce the decomposition
\begin{equation}
\label{eq:decomp_Gamma_proof_main}
\ProdBa_{1:n} = \prod_{\ell=1}^N \Y_\ell\eqsp, \quad \text{where} \quad \Y_\ell = \prod_{i=j_{\ell-1}}^{j_\ell} (\Id - \alpha
\funcA{\State_{i}}) \eqsp, ~~\ell \in \{1,\ldots,N\} \eqsp.
\end{equation}
Using a crude bound $\norm{\Y_{N}} \leq (1 + \alpha \bConst{A})^h$, we get
\begin{equation}
\textstyle \PE^{1/p}_{\xi}[\normop{\ProdBa_{1:n}}^{p}] \leq (1 + \alpha \bConst{A})^h \PE_{\xi}^{1/p}[ \normopLigne{\prod_{\ell=1}^{N-1} \Y_{\ell} }^{p}]\eqsp.
\end{equation}
Now we aim to bound $\PE_{\xi}^{1/p}[\normopLigne{\prod_{\ell=1}^{N-1} \Y_{\ell} }^{p}]$ with the technique introduced in \Cref{th:general_expectation UGE}. To do so, we define, for $\ell \in \{1,\ldots,N-1\}$, the filtration $\mathcal{H}_{\ell} = \sigma(Z_k \,: \, k \leq j_{\ell})$ and establish  almost sure bounds on $\norm{\CPE[\xi]{\Y_\ell}{\mathcal{H}_{\ell-1}}}[Q]$ and $\norm{\Y_{\ell}-\CPE[\xi]{\Y_\ell}{\mathcal{H}_{\ell-1}}}[Q]$ for $\ell \in \{1,\ldots,N-1\}$. More precisely, by the Markov property, it is sufficient to show that there exist $\mtt \in \ocintLine{0,1}$ and $\sigma > 0$ such that for any probabilities $\xi, \xi'$ on $(\Zset,\Zsigma)$,
\begin{equation}
\label{eq:form_bound_ass_prod_mat_v2_Z}
\norm{\PE_{\xi'}[\Y_1]}[Q]^2  \leq 1 - \mtt \text{  and } \norm{\Y_1 - \PE_{\xi'}[\Y_1]}[Q] \leq \sigma \eqsp, \quad \text{ $\PP_{\xi}$-a.s.}\eqsp.
\end{equation}
Such bounds require the blocking procedure, since \eqref{eq:form_bound_ass_prod_mat_v2_Z} not necessarily holds with $h = 1$. Set
\begin{equation}
\label{eq:block_size_h}
h = \lceil 8 \qcond^{1/2} \bConst{A} /a \rceil \taumix \eqsp.
\end{equation}
Applying \Cref{lem:proof_uge_1} and \Cref{lem:proof_uge_2}, we show that \eqref{eq:form_bound_ass_prod_mat_v2_Z} hold with $\mtt = a \alpha h/6$ and $\sigma = \bConst{\sigma} \alpha h$, with $\bConst{\sigma} = 2(\qcond^{1/2}\bConst{A} + a/6)$. Then, applying \Cref{th:general_expectation UGE},
\begin{align}
\textstyle \PE_{\xi}^{1/p}\left[ \normop{\ProdBa_{1:n}}^{p} \right]
&\leq  \textstyle \PE_{\xi}^{1/q}\left[ \normop{\ProdBa_{1:n}}^{q}\right] \leq \sqrt{\qcond} d^{1/q} \rme^{\alpha \bConst{A} h} (1 - a \alpha h /6+ (q-1) \bConst{\sigma}^2 \alpha^2 h^2 )^{N-1}  \nonumber \\
& \leq \sqrt{\qcond} d^{1/q} \rme^{\alpha \bConst{A} h}  \rme^{- a\alpha h (N-1)/6 + (q-1) \alpha^2 \bConst{\sigma}^2 h^2 (N-1)} \nonumber \\
& \leq \sqrt{\qcond} d^{1/q} \rme^{\alpha h (\bConst{A} + a/6)}  \rme^{- a\alpha n/6  + (q-1) \alpha^2 n \bConst{\sigma}^2 h} \nonumber \\
\label{def: constGamma}
& \leq \sqrt{\qcond} \rme^2 d^{1/q} \rme^{- a\alpha n/6  + (q-1) \bConst{\Gamma} \alpha^2 n}\eqsp.
\end{align}
Here we used that by definition of $h$ and since  $\alpha \in \ocintLine{0,\alpha_\infty^{(\Markov)}\taumix^{-1}}$, $\alpha h \bConst{A} \leq 1$, and $\alpha h a /6\leq 1$ by \eqref{eq:h_markov_def}.
\endproof

\subsection{Proof of \Cref{prop:J_n_0_bound_Markov}}
\label{sec:proof_j_n_0_bound_Markov}
Define
\begin{equation}
\label{eq:const_D_1_Markov}
\ConstDM_{1} = 2^{7/2}\qcond^{1/2} a^{-1} \{\rme^{-1/4} + \sqrt{2\pi \rme} \bConst{A} a^{-1}\}\eqsp.
\end{equation}
We first apply the Abel transform  to $\Jnalpha{n}{0}$. Using the representation \eqref{eq:jn0_main}, we obtain that
\begin{align}
\Jnalpha{n}{0}
&= \textstyle \alpha \sum_{j=1}^{n} \left(\Id - \alpha \bA \right)^{n-j}\funnoisew(\State_j)  \label{eq:J_n0_repr_Abel} \\
&= \textstyle \alpha  \left(\Id - \alpha \bA \right)^{n-1} \sum_{k=1}^{n}\funnoisew(\State_k) - \alpha^2 \sum_{j=1}^{n-1}\left(\Id - \alpha \bA \right)^{n-j-1} \bA \sum_{k=j+1}^{n}\funnoisew(\State_k)\eqsp.  \nonumber
\end{align}
Note that $\invariantQ(\funnoisew) = 0$, and for any $z \in \Zset$, \Cref{assum:noise-level} implies $\norm{\funnoisew(z)} \leq \supconsteps$. Hence, applying \Cref{lem:bounded_differences_norms_markovian}, we get for any $j \in \nset$ and $t > 0$, that
\begin{equation}
\label{eq:hp_bound_stationary}
\PP_{\xi}\bigl(\normop{\sum\nolimits_{k=j+1}^{n}\funnoisew(\State_k)} \geq t \bigr)
\leq 2\exp\bigl\{-t^2/(2\beta_{n-j}^{2}) \bigr\}\eqsp,
\end{equation}
where for $\ell \in \nset^*$,
\begin{equation}
\label{eq:beta_ell_def_markov}
\beta_{\ell} = 8 \sqrt{\ell \taumix} \supconsteps \eqsp.
\end{equation}
\Cref{lem:bound_subgaussian} and \eqref{eq:hp_bound_stationary} imply that, for any $p \geq 2$,
\begin{align}
\textstyle \PE_{\xi}^{1/p}\bigl[\normop{\sum_{k=j+1}^{n}\funcnoise{\State_k}}^{p}\bigr] \leq 2^{7/2} \sqrt{(n-j) p \taumix} \supconsteps\eqsp.
\end{align}
Then, applying Minkowski's inequality to \eqref{eq:J_n0_repr_Abel}, we get
\begin{multline}
\textstyle \PE_{\xi}^{1/p}\left[\norm{\Jnalpha{n}{0}}^{p}\right]
\leq 2^{7/2} \alpha \normop{\left(\Id - \alpha \bA \right)^{n-1}}\sqrt{n p \taumix} \supconsteps \eqsp \\ +
\textstyle 2^{7/2} \alpha^2 \sum_{j=1}^{n-1}\norm{\left(\Id - \alpha \bA \right)^{n-j-1}\bA}\sqrt{(n-j) p \taumix} \supconsteps\eqsp.
\end{multline}
Using \Cref{assum:A-b} and \Cref{fact:Hurwitzstability}, for $j \in \{1,\dots,n\}$, $\normop{\left(\Id - \alpha \bA \right)^{n-j}} \leq \sqrt{\qcond} (1-\alpha a)^{(n-j)/2}$.
Note also that, since $a \alpha \leq 1/2$,
\begin{multline}
\label{eq:sum_gamma_bound_markov}
\sum_{j=1}^{n-1}(1-\alpha a)^{(n-j-1)/2} \sqrt{n-j} \leq \rme^{\alpha a} \sum_{k=1}^{n-1}\exp\{-\alpha a (k+1)/2\}\sqrt{k} \\
\leq \frac{2^{3/2}\rme^{\alpha a}}{(\alpha a)^{3/2}}\int_{0}^{+\infty}\exp\{-y\}\sqrt{y}\,\rmd y \leq \frac{2^{1/2}\uppi^{1/2} \rme^{1/2}}{(\alpha a)^{3/2}}\eqsp.
\end{multline}
It remains to combine the previous bounds with an elementary inequality, using $\alpha a \leq 1/2$, for any $x > 0$,
\[
(1-\alpha a)^{(x-1)/2}\sqrt{x} \leq \rme^{\alpha a /2} \exp\{-\alpha a x/2\}\sqrt{x}\leq \frac{\rme^{1/4}}{(\alpha a)^{1/2}} \sup_{u\geq 0} \{u\rme^{-u}\}^{1/2} \leq \frac{1}{(\alpha a)^{1/2} \rme^{1/4}}\eqsp.
\]
Combining the bounds above yield \eqref{eq:J_n_0_bound_Markov} with the constant $\ConstDM_{1}$ defined in
\eqref{eq:const_D_1_Markov}.

\subsection{Proof of \Cref{prop:LSA_error_Markov}}
\label{proof:prop:LSA_error_Markov}
Define
\begin{equation}
\label{eq:definition:ConstDM_2}
\ConstDM_2 = \ConstDM_{1} (1 + 24\sqrt{2}\rme^2 \sqrt{\qcond} \bConst{A} a^{-1}) \,,
\end{equation}
where $\ConstDM_1$ is given in \eqref{eq:const_D_1_Markov}. Proceeding as in \eqref{eq:p_norm_lsa_err_appendix}, we get
\begin{equation}
\label{eq:theta_n_bound_Markov_general}
\PE_\xi^{1/p}\left[\norm{\theta_{n} - \thetas}^{p}\right]
\leq \PE^{1/p}_\xi\left[\norm{\ProdBa_{1:n} (\thetainit - \thetas)}^{p}\right] +
\PE_\xi^{1/p}\left[\norm{\Jnalpha{n}{0}}^{p}\right]
+\PE_\xi^{1/p} \left[\norm{\Hnalpha{n}{0}}^{p}\right]\eqsp.
\end{equation}
The first two terms are bounded using   \eqref{eq:concentration_UGE_simple} and \Cref{prop:J_n_0_bound_Markov}, respectively. Regarding the last one, the recurrence \eqref{eq:jn0_main}, $\Hnalpha{n}{0} = -\alpha \sum_{j=1}^{n}\ProdBa_{j+1:n}\zmfuncA{Z_j} \Jnalpha{j-1}{0}$, and Minkowski's inequality yields
\begin{equation}
\txts \PE^{1/p}_\xi\left[\norm{\Hnalpha{n}{0}}^{p}\right]
\leq \alpha \sum_{j=1}^{n} \bigl\{\PE_\xi\bigl[\norm{\ProdBa_{j+1:n}}^{2p}\bigr]\bigr\}^{1/2p}  \bigl\{\PE_\xi\bigl[\norm{ \zmfuncA{Z_j} \Jnalpha{k-1}{0}}^{2p}\bigr]\bigr\}^{1/2p} \eqsp.
\end{equation}
Using \Cref{prop:J_n_0_bound_Markov} and $\rme^{-x} \leq 1 - x/2$, valid for $x \in [0,1]$,  we get
\begin{align}
\textstyle \PE^{1/p}_\xi\bigl[\norm{\Hnalpha{n}{0}}^{p}\bigr]  &\leq \textstyle  \alpha d^{1/q} \supconsteps \rme^2\sqrt{\qcond}  \bConst{A} \ConstDM_{1}\sqrt{2 \alpha a p \taumix} \sum_{j=1}^n (1-a\alpha/24)^n  \eqsp.
\end{align}
This completes the proof.

\subsection{Proof of \Cref{prop:J_n_1_2_bound_markov}}
\label{sec:proof:prop:J_n_1_2_bound_markov}
Define 
\begin{align}
\label{eq:def:ConstDM_J,1}
\ConstDM_{J,1} &= 64\qcond \bConst{A} a^{-2} \left((\sqrt{2} + \qcond^{1/2})/\sqrt{2\log{2}}  + 2 \uppi^{1/2} \qcond^{1/2} + \qcond^{1/2}/\sqrt{\log{2}}\right) \\
\label{eq:def:ConstDM_J,2}
\ConstDM_{J,2} &= (128/3) \qcond^{3/2} \bConst{A} a^{-2} \\
\label{eq:def:ConstDM_H}
\ConstDM_{H,1} &= 96 a^{-1} \bConst{A} \rme^{2} \qcond^{1/2} \ConstDM_{J,1} \quad \text{and} \quad \ConstDM_{H,2} = 48 a^{-1} \bConst{A} \rme^{2} \qcond^{1/2} \ConstDM_{J,2}.
\end{align}
We preface the proof of this proposition by giving a statement of the Berbee lemma, which plays an essential role. Consider the extended measurable space $\tmszn = \msz^{\nset} \times [0,1]$, equipped with the $\sigma$-field $\tmczn = \mcz^{\otimes \nset} \otimes \mathcal{B}([0,1])$. For each probability measure $\xi$ on $(\Zset,\Zsigma)$, we consider the probability measure $\PPext_{\xi} = \PP_{\xi} \otimes \mathbf{Unif}([0,1])$ and denote by $\PEext_{\xi}$ the corresponding expectated value. Finally, we denote by $(\tZ_k)_{k \in\nset}$ the canonical process $\tZ_k : ((z_i)_{i\in\nset},u) \in \tmszn \mapsto z_k$ and $U :((z_i)_{i\in\nset},u) \in \tmszn \mapsto u$. Under $\PPext_{\xi}$, $\sequence{\tZ}[k][\nset]$ is by construction a Markov chain with initial distribution $\xi$ and Markov kernel $\MKQ$ independent of $U$. The distribution of $U$ under $\PPext_{\xi}$ is uniform over $\ccint{0,1}$.
\begin{lemma}
  \label{lem:construction_berbee}
  Assume  \Cref{assum:drift}, let $m \in \nsets$ and $\xi$ be a probability measure on $(\msz,\mcz)$.   Then,  there exists a random process $(\tZs_{k})_{k\in\nset}$ defined on $(\tmszn, \tmczn, \PPext_{\xi})$ such that for any $k \in \nset$,
  \begin{enumerate}[wide,label=(\alph*)]
  \item \label{lem:construction_a} $\tZs_{k}$ is independent of $\tilde{\mcf}_{k+m} = \sigma\{\tZ_{\ell} \, : \, \ell \geq k+m\}$;
  \item \label{lem:construction_b} $\PPext_{\xi}(\tZs_k \neq \tZ_k) \leq  \dobru{\MKQ^m}$;
   \item \label{lem:construction_c} the random variables $\tZs_k$ and $\tZ_k$ have the same distribution under $\PPext_\xi$.
  \end{enumerate}
\end{lemma}
\proof{Proof.}  Berbee's lemma \citet[Lemma~5.1]{riobook} ensures that for any $k$,
  there exists $\tZs_k$ satisfying \ref{lem:construction_a},
  \ref{lem:construction_c} and
  $\PPext_{\xi}(\tZs_k \neq \tZ_k) = \beta_{\xi}(\sigma(\tZ_{k}),
  \tilde{\mcf}_{k+m})$. Here for two sub $\sigma$-fields $\mathfrak{F}$,
  $\mathfrak{G}$ of $\tmczn$,
  \begin{equation}
    \label{eq:1}
    \beta_{\xi}(\mathfrak{F},\mathfrak{G}) = \frac{1}{2}\sup \sum_{i \in \msi} \sum_{j \in \msj} | \PPext_{\xi}( \msa_i \cap \msb_j)- \PPext_{\xi}(\msa_i)\PPext_{\xi}(\msb_j)|\eqsp,
  \end{equation}
  and the supremum is taken over all pairs of partitions $\{\msa_i\}_{i\in\msi} \in \mathfrak{F}^\msi$ and $\{\msb_j\}_{j\in\msj}\in \mathfrak{G}^\msj$ of $\tmszn$ with $\msi$ and $\msj$ finite. Applying \citet[Theorem 3.3]{douc:moulines:priouret:soulier:2018} with \Cref{assum:drift} completes the proof.
\endproof

\proof{Proof of \Cref{prop:J_n_1_2_bound_markov}.}
Recall that $\Jnalpha{n}{1} = \alpha^2 \sum_{\ell=1}^{n-1} S_{\ell+1:n} \funnoisew(\State_\ell)$, where 
\begin{equation}
\label{eq:S_l_n_def}
\textstyle S_{\ell+1:n} = \sum_{k=\ell+1}^{n} (\Id - \alpha \bA)^{n-k} \zmfuncA{\State_{k}} (\Id - \alpha \bA)^{k-1 - \ell} \eqsp.
\end{equation}
We first set a constant block size $m \in \nset^{\star}, m \geq \taumix$ (to be determined later).
In order to proceed with $S_{\ell+1:n}\funnoisew(\State_{\ell})$, we split $S_{\ell+1:n}$ into a part measurable \wrt\, $\mcf_{\ell+m}^{n}= \sigma(Z_{k} : k \geq m+\ell)$ and a remainder term. Indeed, using its definition \eqref{eq:S_l_n_def},
\begin{align}
S_{\ell+1:n} &=(\Id - \alpha \bA)^{n-m-\ell} S_{\ell+1:\ell+m} + S_{\ell+m+1:n} (\Id - \alpha \bA)^{m}\eqsp.
\end{align}
Let $N = \round{(n-1)/m}$. With these notations, we can decompose $\Jnalpha{n}{1}$ as a sum of three terms:
$\Jnalpha{n}{1} = T_1 + T_2 +T_3$, with
\begin{align}
T_1 &= \alpha^2 \sum_{\ell=1}^{m(N-1)} (\Id - \alpha \bA)^{n-m-\ell} S_{\ell+1:\ell+m} \funnoisew(\State_\ell)\\
T_2 & = \alpha^2 \sum_{\ell=1}^{m(N-1)} S_{\ell+m+1:n} (\Id - \alpha \bA)^{m} \funnoisew(\State_\ell) \eqsp, \quad T_3  = \alpha^2\sum_{\ell=m(N-1) + 1}^{n-1} S_{\ell+1:n}  \funnoisew(\State_\ell) \eqsp.
\end{align}
We bound the terms $T_1,T_2$ and $T_3$ separately. Using Minkowski's inequality together with \Cref{fact:Hurwitzstability}, \Cref{lem:S_l_bound_lin}, and the definition \eqref{eq:S_l_n_def}, we get
\begin{align}
\PE^{1/p}_{\xi}\bigl[\norm{T_{1}}^{p}\bigr]
&\leq \alpha^2 \sum_{\ell=1}^{m(N-1)} {\qcond}^{1/2} (1-\alpha a)^{(n-m-\ell)/2} \PE^{1/p}_{\xi}\bigl[\norm{S_{\ell+1:\ell+m}\funnoisew(\State_{\ell})}^{p}\bigr] \\
&\leq   16 \alpha^{2} \kappa_Q^{3/2}  \bConst{A}
\ConstDM_{S} \supconsteps \sqrt{m \taumix \{\trace{\noisecov}\} p}\, \sum_{\ell=1}^{m(N-1)} (1-\alpha a)^{(n-\ell-1)/2} \\
&\leq 32 \qcond^{3/2} \bConst{A} a^{-1} \alpha \supconsteps \sqrt{m \taumix p}\eqsp,
\end{align}
where for the last inequality, we  additionally used that $\sqrt{1-x} \leq 1-x/2$ for $x \in [0,1]$. Similarly, with Minkowski's inequality and \Cref{lem:S_l_bound_lin}, we bound $T_3$:
\begin{align}
\PE^{1/p}_{\xi}\bigl[\norm{T_{3}}^{p}\bigr]
&\leq \alpha^{2} \sum_{\ell = m(N-1) + 1}^{n-1} \PE^{1/p}_{\xi}\bigl[\norm{S_{\ell+1:n} \funnoisew(\State_{\ell})}^{p}\bigr] \\
& \leq 16\sqrt{2} \alpha^{2} \qcond  \bConst{A} \sqrt{m \taumix p} \supconsteps  \sum_{\ell= m(N-1) + 1}^{n-1} (1-\alpha a)^{(n-\ell-1)/2} \\
&\leq 32\sqrt{2} \qcond \bConst{A} a^{-1} \alpha  \sqrt{m \taumix p} \supconsteps\eqsp.
\end{align}
In the bound above we used that $n-1-m(N-1) \leq 2m$. Combining the above,
\begin{equation}
\label{eq:T_1_3_bound}
\PE^{1/p}_{\xi}\bigl[\norm{T_{1}}^{p}\bigr] + \PE^{1/p}_{\xi}\bigl[\norm{T_{3}}^{p}\bigr] \leq \AuxconstM_{1} \alpha a  \sqrt{m \taumix p} \supconsteps\eqsp,
\end{equation}
where $\AuxconstM_{1} = 32 \qcond \bConst{A} a^{-2} (\sqrt{2}+\qcond^{1/2})$. It remains to bound $\PE_\xi^{1/p}[ \| T_2 \|^p]$. We switch to the extended space $(\tmszn,\tmczn,\PPext_{\xi})$,and, using \Cref{lem:construction_berbee}, we get that
$\PE_\xi^{1/p}[ \| T_2 \|^p]= \PEext_\xi^{1/p}[\| \tilde{T}_2 \|^p]$ with $\tilde{T}_2= \alpha^2 \sum_{\ell=1}^{m(N-1)} \tilde{S}_{\ell+m+1:n} (\Id - \alpha \bA)^{m} \funnoisew(\tZ_\ell)$. Here $\tilde{S}_{\ell+m+1:n}$ is a counterpart of $S_{\ell+m+1:n}$ defined on the extended space, that is,
\begin{equation}
\textstyle \tilde{S}_{\ell+m+1:n} = \sum_{k=\ell+m+1}^{n} (\Id - \alpha \bA)^{n-k} \zmfuncA{\tZ_{k}} (\Id - \alpha \bA)^{k-1 - \ell}\eqsp.
\end{equation}
We further decompose $\tilde{T}_2= \tilde{T}_{2,1} + \tilde{T}_{2,2}$, where
\begin{align}
  \tilde{T}_{2,1}&= \alpha^{2} \sum_{k=0}^{N-2} \sum_{i=1}^{m} \tilde{S}_{(k+1)m + i + 1:n} (\Id - \alpha \bA)^{m} \funnoisew(\tZs_{km+i})\eqsp, \\
  \label{eq:def_t_2_2}
\tilde{T}_{2,2}&= \alpha^{2} \sum_{k=0}^{N-2} \sum_{i=1}^{m} \tilde{S}_{(k+1)m + i + 1:n} (\Id - \alpha \bA)^{m} \{\funnoisew(\tZ_{km+i}) - \funnoisew(\tZs_{km+i})\}\eqsp.
\end{align}
We begin with bounding $\tilde{T}_{2,2}$. Set $V_{\ell} = \funnoisew(\tZ_{\ell}) - \funnoisew(\tZs_{\ell})$ and $\tilde{\mcf}^\star_\ell= \sigma( \tZ_k, \tZs_k \, : \, k \leq \ell)$.
Using \Cref{lem:construction_berbee} we get with the convention $0/0=0$,
\begin{align}
  & \PEext_{\xi}^{1/p}[\norm{\tilde{S}_{(k+1)m + i + 1:n} (\Id - \alpha \bA)^{m} V_{km+i}}^{p}] \\
    &\qquad = \PEext_{\xi}^{1/p}[\norm{\tilde{S}_{(k+1)m + i + 1:n} (\Id - \alpha \bA)^{m} V_{km+i} \indiacc{\tZ_{km+i} \neq \tZs_{km+i}}}^{p}] \\
&\qquad \leq \PEext_{\xi}^{1/p}\bigl[ \norm{V_{km+i}}^{p} \CPEtilde{\norm{\tilde{S}_{(k+1)m + i + 1:n} (\Id - \alpha \bA)^{m} V_{km+i}/\norm{V_{km+i}}}^{p}}{\tilde{\mcf}_{km+i}^\star}\bigr] \\
&\qquad \leq \PEext_{\xi}^{1/p}\bigl[ \norm{V_{km+i}}^{p} \sup_{u \in \sphere^{d-1}\, ,\,\xi' \in \mathcal{P}(\msz)}\PEext_{\xi'}[\norm{\tilde{S}_{(k+1)m + i + 1:n} (\Id - \alpha \bA)^{m} u}^{p}]\bigr]\eqsp,
\end{align}
where $\mathcal{P}(\msz)$ is the set of probability measure on $(\msz,\mcz)$.
Applying \Cref{lem:S_l_bound} and \Cref{fact:Hurwitzstability}, for any $u \in \sphere^{d-1}$ and probability measure $\xi'$,
\begin{multline}
\PEext^{1/p}_{\xi'}[\norm{\tilde{S}_{(k+1)m + i + 1:n} (\Id - \alpha \bA)^{m} u}^{p}]= \PE^{1/p}_{\xi'}[\norm{S_{(k+1)m + i + 1:n} (\Id - \alpha \bA)^{m} u}^{p}] \\ \leq
16 \qcond^{3/2} \bConst{A}  [(n - (k+1)m - i) \taumix (1-\alpha a)^{n-km -i-1} p]^{1/2}\eqsp.
\end{multline}
Moreover, under \Cref{assum:noise-level} and \Cref{assum:drift}, $\norm{V_{km+i}} \leq 2\supconsteps \indinD{\tZ_{km+i} \neq \tZs_{km+i}}$, and $\PPext_{\xi}(\tZs_{km+i} \neq \tZ_{km+i}) \leq \dobru{\MKQ^m}  \leq (1/4)^{\lfloor m/\taumix \rfloor}$ by \Cref{lem:construction_berbee} and \Cref{assum:drift}. Combining the bounds above,
\begin{align}
\label{eq:3}
&\PEext_{\xi}^{1/p}[\norm{\tilde{S}_{(k+1)m + i + 1:n} (\Id - \alpha \bA)^{m} V_{km+i}}^{p}] \leq 32 \qcond^{3/2} \bConst{A} \supconsteps (1/4)^{(1/p) \lfloor m/\taumix \rfloor} \times \\
& \qquad\qquad\qquad \bigl[(n-(k+1)m - i) (1-\alpha a)^{(n-k m -i-1)}\taumix p\bigr]^{1/2}\eqsp.
\end{align}
Substituting \eqref{eq:3}  into the definition \eqref{eq:def_t_2_2}  of $\tilde{T}_{2,2}$, and using \begin{equation}
\textstyle \sum_{\ell = 1}^{m(N-1)} \sqrt{n-\ell} (1-\alpha a)^{(n-\ell+1)/2} \leq \int_{0}^{\plusinfty}t^{1/2} \rme^{-\alpha a t/2} \rmd t = 2^{3/2}(a\alpha)^{-3/2} \Gamma(3/2)\eqsp,
\end{equation}
we get
\begin{equation}
\label{eq:T_2_2_final_bound}
\textstyle \PEext^{1/p}_{\xi}\bigl[\norm{\tilde{T}_{2,2}}^{p}\bigr] \leq \AuxconstM_{2} (\alpha a)^{1/2} (1/4)^{(1/p) \lfloor m/\taumix \rfloor} \sqrt{\taumix p} \supconsteps \eqsp,
\end{equation}
where $\AuxconstM_{2} = 64 \uppi^{1/2} \qcond^{3/2} \bConst{A} a^{-2}$. To obtain \eqref{eq:T_2_2_final_bound} we have additionally used that $m \geq 1$ and $\alpha a \leq 1/2$.
\par
Now we bound $\tilde{T}_{2,1}$. Define the function $g(z): \Zset \mapsto \rset^{d}$, $g(z) = (\Id - \alpha \bA)^{m} \funnoisew(z)$. \Cref{assum:noise-level} and \Cref{fact:Hurwitzstability} imply $\supnorm{g} \leq \qcond^{1/2}(1-\alpha a)^{m/2} \supconsteps$ and $\pi(g) = 0$. Then we apply  \Cref{lem:construction_berbee} and \Cref{lem:vector_valued_burkholder}, and obtain
\begin{align}
  \PEext^{1/p}_{\xi}\bigl[\norm{\tilde{T}_{2,1}}^{p}\bigr] \leq \alpha^{2} \sum_{i=1}^{m}\biggl[2p \supnorm{g} \bigl\{{\txts\sum_{k=0}^{N-2}\sup_{u \in \sphere^{d-1}}\PEext_{\xi}^{2/p}[\normop{\tilde{S}_{(k+1)m + i + 1:n} u}^{p}]}\bigr\}^{1/2}  \\
  +\biggr.
\biggl. \sum_{k=0}^{N-2} \norm{\xi \MKQ^{km+i}g} \sup_{u \in \sphere^{d-1}} \PE_{\xi}^{1/p}\big[\norm{S_{(k+1)m + i + 1:n} u}^p\big] \biggr]\eqsp.
\end{align}
Assumption \Cref{assum:drift} with $\pi(g) = 0$ implies $\norm{\xi \MKQ^{km+i}g} \leq \dobru{\MKQ^{km+i}} \supnorm{g}$. Combining it with \Cref{lem:S_l_bound},
\begin{align}
&\sum_{i=1}^{m}\sum_{k=0}^{N-2}\norm{\xi \MKQ^{km+i}g} \sup_{u \in \sphere^{d-1}} \PE_{\xi}^{1/p}\big[\norm{S_{(k+1)m + i + 1:n} u}^p\big] \\
&\qquad \leq 16 \qcond^{3/2} \bConst{A} (1-\alpha a)^{(m-1)/2} \sup_{x \geq 1 }\{x (1-\alpha a)^{x}\}^{1/2} \sqrt{\taumix p} \supconsteps \sum_{\ell = 0}^{+\infty} \dobru{\MKQ^{\ell}}  \\
  &\qquad
\leq \frac{64}{3\rme^{1/2}} (a \alpha)^{-1/2} \qcond^{3/2} \bConst{A} (1-\alpha a)^{(m-1)/2} \taumix^{3/2} \sqrt{p} \supconsteps \eqsp,
\end{align}
where we have used for the last inequality \eqref{eq:crr_koef_sum_tau_mix}, $a\alpha \leq 1/2$, and $\sup_{x \geq 1 }\{x (1-\alpha a)^{x}\}^{1/2} \leq \rme^{-1/2}(a \alpha)^{-1/2}$. Jensen's inequality together with \Cref{lem:S_l_bound} yields
\begin{align}
  &\sum_{i=1}^{m}\bigl\{{\txts \sum_{k=0}^{N-2}\sup_{u \in \sphere^{d-1}} \PE_{\xi}^{2/p}[\normop{S_{(k+1)m + i + 1:n} u}^{p}] }\bigr\}^{1/2}  \\
  & \qquad \leq \sqrt{m}\bigl\{{\txts \sum_{\ell=1}^{m(N-1)}\sup_{u \in \sphere^{d-1}} \PE_{\xi}^{2/p}[\normop{S_{\ell+m+1:n} u}^{p}]}\bigr\}^{1/2} \\
&\qquad \leq  16 \qcond \bConst{A} (m \taumix p)^{1/2} \bigl\{ {\txts \sum_{\ell=1}^{m(N-1)} (n-\ell-m) (1-\alpha a)^{n -\ell -m - 1} }\bigr\}^{1/2} \\
&\qquad \leq 16 \sqrt{ 2} \qcond \bConst{A} (m \taumix p)^{1/2} (\alpha a)^{-1}\eqsp.
\end{align}
Combining the bounds above with $\supnorm{g} \leq \qcond^{1/2}(1-\alpha a)^{m/2} \supconsteps$, we get
\begin{align}
\label{eq:T_2_1_final_bound_markov}
&\PEext^{1/p}_{\xi}\bigl[\norm{\tilde{T}_{2,1}}^{p}\bigr] \leq [32 \sqrt{2} \qcond^{3/2} \bConst{A} a^{-2} ]\sqrt{m \taumix} \alpha a p^{3/2} \supconsteps \\
&\qquad\qquad\qquad + [(64/3 \rme^{-1/2})\qcond^{3/2} \bConst{A} a^{-2}] (\alpha a \taumix)^{3/2} p^{1/2} \supconsteps\eqsp.
\end{align}
Now the proof is completed combining \eqref{eq:T_1_3_bound}, \eqref{eq:T_2_2_final_bound}, and \eqref{eq:T_2_1_final_bound_markov}, setting
\begin{equation}
m = \taumix \left\lceil \frac{p\log(1/\alpha a)}{2 \log(2)} \right\rceil\eqsp,
\end{equation}
and using $p^{1/2} \leq p$ and $\taumix^{1/2} \leq \taumix$.
Indeed, with this choice of $m$, $(1/4)^{(1/p) \lfloor m/\taumix \rfloor} \leq \sqrt{\alpha a}$, $m \geq \taumix$. In addition, note that $m\leq 2 \taumix p\log(1/\alpha a)/(2 \log(2))$  using $\alpha a \leq 1/2$ and $p \geq 2$.

We now prove  \eqref{eq:H_n_1_bound_markov}.
The decomposition \eqref{eq:jn_allexpansion_main} implies
$\Hnalpha{n}{1} = -\alpha \sum_{\ell=1}^{n}\ProdBa_{\ell+1:n} \zmfuncA{\State_\ell} \Jnalpha{\ell-1}{1}$.
Hence, with Minkowski's and Holder's inequalities,
\begin{align}
\textstyle
\PE^{1/p}_{\xi}[\normop{\Hnalpha{n}{1}}^{p}] \leq \alpha \sum_{\ell=1}^{n}  \PE^{1/2p}_{\xi}[\normop{\ProdBa_{\ell+1:n} \zmfuncA{\State_\ell}}^{2p}] \PE^{1/2p}_{\xi}[\norm{\Jnalpha{\ell-1}{1}}^{2p}]\eqsp.
\end{align}
Applying \Cref{prop:products_of_matrices_UGE} and \eqref{eq:j_n_1_bound_markov}, we get 
\begin{align}
\PE^{1/p}_{\xi}[\normop{\Hnalpha{n}{1}}^{p}]
  &\leq 4\rme^{2} \qcond^{1/2} \bConst{A} \ConstDM_{J,1} \taumix d^{1/q} p^{2} \alpha^2 a \sqrt{\log{(1/\alpha a)}}\,\supconsteps \sum_{\ell=1}^{n}\rme^{- a\alpha n/12} \\
& +2\rme^{2} \qcond^{1/2} d^{1/q} \ConstDM_{J,2}  p^{1/2} (\alpha a  \taumix)^{3/2} \,\supconsteps \sum_{\ell=1}^{n}\rme^{- a\alpha n/12}\eqsp.
\end{align}
Now the proof follows from elementary bound $\rme^{-x} \leq 1-x/2$, $x \in [0,1]$.
\endproof

\subsection{Proof of \Cref{prop:bias_estimate_PR_markov}}
\label{prop:bias_estimate_PR_markov_proof}
Define the constants
\begin{equation}
\label{eq:def:const_D_4_M}
\begin{split}
\ConstDM_{4} &= 48 \qcond^{1/2} \rme^{3} \eqsp, \\
\ConstDM_{5} &= 4\rme  (\ConstDM_{J,1} + \ConstDM_{H,1}) + \ConstDM_{7} / \sqrt{\log{2}} \eqsp, \\
\ConstDM_{6} &= \sqrt{2} \rme (\ConstDM_{J,2} + \ConstDM_{H,2})\eqsp.
\end{split}
\end{equation}
Note first that 
\[
\norm{\PE_\xi[\prtheta_{n}] - \thetas} \leq (2/n)\sum_{t=n/2}^{n-1}\norm{\PE_\xi[\theta_t] - \thetas}\eqsp.
\]
Proceeding as in \eqref{eq:p_norm_lsa_err_appendix}, we get for each $t \in \{n/2,\ldots,n\}$ that
\begin{equation}
\label{eq:theta_n_bound_bias_Markov}
\norm{\PE_\xi[\theta_t] - \thetas} = \sup_{u \in \sphere^{d-1}} \{\PE_{\xi}[u^{\top}\ProdBa_{1:t} (\thetainit - \thetas)] + \PE_{\xi}[u^{\top}\Jnalpha{t}{0}] + \PE_{\xi}[u^{\top}\Hnalpha{t}{0}]\}\eqsp.
\end{equation}
Now we bound each term above separately. For the first one note that 
\[
\bigl\vert \PE_{\xi}[u^{\top}\ProdBa_{1:t} (\thetainit - \thetas)] \bigr\vert \leq \PE_{\xi}^{1/2}\left[ \normop{\ProdBa_{1:t}}^{2} \right] \normop{\thetainit - \thetas} \leq \sqrt{\qcond} \rme^3 \rme^{- a\alpha t/12} \normop{\thetainit - \thetas}\eqsp.
\]
Using the representation \eqref{eq:jn0_main}, \Cref{fact:Hurwitzstability}, and \Cref{assum:drift}, we get that 
\[
\bigl\vert \PE_{\xi}[u^{\top}\Jnalpha{t}{0}] \bigr\vert = \alpha \bigl\vert \PE_{\xi}[u^{\top}\sum\nolimits_{j=1}^{t}(\Id - \alpha \bA)^{t-j}\funnoisew(\State_{j})] \bigr\vert \leq \alpha \qcond^{1/2} \bConst{A} \supconsteps \sum_{\ell=0}^{\infty} \dobru{\MKQ^{\ell}} \leq \ConstDM_{7} \alpha a \taumix \supconsteps \eqsp.
\]
Moreover, applying the Cauchy-Schwartz inequality and \Cref{prop:J_n_1_2_bound_markov}, we get 
\begin{align}
\bigl\vert \PE_{\xi}[u^{\top}\Hnalpha{t}{0}] \bigr\vert 
&\leq \PE_{\xi}^{1/2}[\norm{\Hnalpha{t}{0}}^{2}] \leq \PE^{1/2}_{\xi}[\normop{\Jnalpha{t}{1}}^{2}] + \PE^{1/2}_{\xi}[\normop{\Hnalpha{t}{1}}^{2}] \\
&\leq 4 \rme (\ConstDM_{J,1} + \ConstDM_{H,1}) \supconsteps (\alpha a \taumix) \sqrt{\log{(1/\alpha a)}} + \sqrt{2} \rme (\ConstDM_{J,2} + \ConstDM_{H,2}) \supconsteps (\alpha a \taumix)^{3/2}\eqsp.
\end{align}
Combining the bounds above yields \eqref{eq:bias_estimate_PR_markov}.

\subsection{Proof of \Cref{lem:bias_estimate_J_n_0_Markov}}
\label{prop:bias_Markov_PR}
\begin{proposition}
\label{lem:bias_estimate_J_n_0_Markov}
Assume \Cref{assum:A-b}, \Cref{assum:noise-level}, and \Cref{assum:drift}. Then, for any $\alpha \in \ocint{0,\alpha_{\infty}}$, $t \in \nset^{\star}$ and initial probability measure $\xi$ on $(\Zset,\Zsigma)$, it holds that
\begin{equation}
\label{eq:bias_estimate_J_n_0_appendix}
\norm{\PE_{\xi}[\zmfuncA{\State_{t+1}}\Jnalpha{t}{0}]} \leq \ConstDM_{7} \alpha a \taumix \supconsteps\eqsp,
\end{equation}
where the constant $\ConstDM_{7}$ is given by
\begin{equation}
\label{eq:def:const_D_7_M}
\ConstDM_{7} = (4/3) \qcond^{1/2} \bConst{A} a^{-1}\eqsp.
\end{equation}
\end{proposition}
\proof{Proof.} 
Using \eqref{eq:jn0_main}, we get
\[
\norm{\PE_{\xi}[\zmfuncA{\State_{t+1}}\Jnalpha{t}{0}]} = \sup_{u \in \sphere^{d-1}} \PE_{\xi}[\alpha u^{\top} \zmfuncA{\State_{t+1}}\sum\nolimits_{j=1}^{t}(\Id - \alpha \bA)^{t-j}\funnoisew(\State_{j})]\eqsp.
\]
Define for $z \in \Zset$ and $j \in \{1,\ldots,t\}$, the function $g_{j,t}(z): \Zset \mapsto \rset^{d}$ as
\begin{align}
\textstyle
g_{j,t}(z) = \int_{\Zset} \zmfuncA{z^{\prime}} (\Id - \alpha \bA)^{t-j}\funnoisew(z) \MKQ^{t-j+1}(z,\rmd z^{\prime})
\end{align}
Using that $\pi(\zmfuncAw) = 0$ together with \Cref{fact:Hurwitzstability} and \Cref{assum:drift}, for any $u \in \sphere^{d-1}$,
\begin{align}
\bigl\vert u^{\top} g_{j,t}(z) \bigr\vert \leq \qcond^{1/2} (1-\alpha a)^{(t-j)/2} \bConst{A} \supconsteps \dobru{\MKQ^{t-j+1}}\eqsp.
\end{align}
Using the Markov property of $(\State_{k})_{k \in \nset}$ and the definition of $\taumix$ (see \Cref{assum:drift}), we get from the previous bound that
\begin{align}
\bigl\vert \PE_{\xi}[\alpha u^{\top} \zmfuncA{\State_{t+1}}\sum\nolimits_{j=1}^{t}(\Id - \alpha \bA)^{t-j}\funnoisew(\State_{j})] \bigr\vert
&\leq \alpha \qcond^{1/2} \bConst{A} \supconsteps \sum\nolimits_{\ell = 0}^{\infty} \dobru{\MKQ^\ell} \\
&\leq \ConstDM_{7} \alpha a \taumix \supconsteps \eqsp,
\end{align}
and \eqref{eq:bias_estimate_J_n_0_appendix} follows.
\endproof.

\subsection{Proof of \Cref{th:theo_2_markov} and \Cref{th:LSA_PR_error_markov}}
\label{proof:LSA_PR_error_markov}
Define the quantities
\begin{equation}
\label{eq:def:const:LSA_PR_markov}
\begin{split}
&\MarDeltafl_{n,p,\alpha,\taumix} = \frac{8\, \ConstDM_{2} \rme^{1/p} \sqrt{a p \taumix}}{\sqrt{\alpha n}} + \frac{2^{1/2} \bConst{\sf{Ros}, 1}^{(\Markov)} \taumix^{3/4} p \log_{2}(2p)}{n^{1/4}} + \frac{2 \bConst{\sf{Ros}, 2}^{(\Markov)} \taumix p \log_2(2p)}{n^{1/2}} \\
& \qquad\qquad + 8 \rme^{1/p} \bigl((\ConstDM_{J,1} + \ConstDM_{H,1}) \alpha a \taumix\, \sqrt{\log{(1/\alpha a)}} p^{2}\bigr)(\alpha^{-1}n^{-1/2} + n^{1/2}\bConst{A}) \\
& \qquad\qquad + 8 \rme^{1/p} (\ConstDM_{J,2} + \ConstDM_{H,2}) (\alpha a \taumix)^{3/2} p^{1/2} (\alpha^{-1}n^{-1/2} + n^{1/2}\bConst{A}) \\
& \MarDeltatr_{n,p,\alpha,\taumix} = \rme^{2 + 1/p} \qcond^{1/2} \left(\frac{4}{\alpha n^{1/2}} + 2^{-1/2} n^{1/2} \bConst{A}\right)\eqsp.
\end{split}
\end{equation}
We begin with the proof of \Cref{th:LSA_PR_error_markov}. The result of \Cref{th:theo_2_markov} will directly follow from it using the step size $\alpha$ fixed in \eqref{eq:step_size_pth_moment_refined_markov}.
\proof{Proof of \Cref{th:LSA_PR_error_markov}.}
Let $p \geq 2$ and $q \geq p$ be a number to be fixed later and assume in addition that $\alpha \in \ocint{0, \alpha_{q,\infty}\taumix^{-1}}$. Below we use shorthand notations $\zmfuncAw[t], \Am_{t} ,\funnoisew_{t}$ for $\zmfuncA{\State_{t}}, \Am(\State_t)$, and $\funcnoise{\State_{t}}$, respectively.
  Proceeding as in \eqref{eq:bound_p_moment_iid_T_1_T_2_decomp} and \eqref{eq:T_1_bound_iid}, we decompose the $p$-th moment of LSA-PR error as
\begin{align}
\label{eq:err_representation_markov}
\textstyle (n/2) \PE_{\xi}^{1/p}\left[\norm{\bA\left(\prtheta_{n}-\thetas\right)}^{p}\right]
             \leq \PE_{\xi}^{1/p}\bigl[\norm{\sum\nolimits_{t=n/2}^{n-1}\funnoisew_{t+1}}^{p}\bigr]  +  T_1^{(\Markov)}+ T_2^{(\Markov)} + T_3^{(\Markov)} 
\end{align}
where we have set $T_1^{(\Markov)} =  \alpha^{-1}\PE_{\xi}^{1/p}[\norm{\theta_{n/2}-\theta_{n}}^p]$,
$T^{(\Markov)}_2 = \PE_{\xi}^{1/p}\bigl[\norm{\sum_{t=n/2}^{n-1}\zmfuncAw[t+1]
\ProdBa_{1:t} \{ \theta_0 - \thetas \}}^{p}\bigr]$ and 
\begin{equation}
\textstyle T^{(\Markov)}_3  = \PE_{\xi}^{1/p}\bigl[\norm{\sum_{t=n/2}^{n-1}\zmfuncAw[t+1]\Jnalpha{t}{0}}^{p}\bigr]  + \PE_{\xi}^{1/p}\bigl[\norm{\sum_{t=n/2}^{n-1}\zmfuncAw[t+1] \Hnalpha{t}{0}}^{p}\bigr]\eqsp.
\end{equation}
 Now we bound each term in the decomposition \eqref{eq:err_representation_markov}.  We begin with the first term, which is linear statistics of uniformly geometrically ergodic Markov chain. Applying \Cref{theo:rosenthal_uge_arbitrary_init}, we obtain  
\begin{align}
\label{eq:veps__markov_bounds}
&\txts \PE_{\xi}^{1/p}\bigl[\norm{\sum_{t=n/2}^{n-1}\funnoisew_{t+1}}^{p}\bigr] \leq \bConst{\sf{Rm}, 1} p^{1/2} n^{1/2} \{\trace{\noisecovmarkov}\}^{1/2} \\
& \qquad \qquad \qquad + \bConst{\sf{Ros}, 1}^{(\Markov)} \supconsteps (n/2)^{1/4}\taumix^{3/4}p\log_2(2p) + \bConst{\sf{Ros}, 2}^{(\Markov)} \supconsteps \taumix p \log_2(2p) \eqsp.
\end{align}
Applying \Cref{prop:LSA_error_Markov} and Minkowski's inequality, we get
\begin{equation}
    \label{eq:t_1_markov_bounds}
T^{(\Markov)}_{1} \leq 2 \alpha^{-1}\sqrt{\qcond} \rme^2 d^{1/q} \rme^{-\alpha a n/24} \norm{\thetainit- \thetas} + 2\ConstDM_{2} d^{1/q} \alpha^{-1/2} \sqrt{a p \taumix} \supconsteps\eqsp.
\end{equation}
Applying \Cref{prop:products_of_matrices_UGE}, Minkowski's inequality, and using \Cref{assum:A-b}, we get
\begin{equation}
  \label{eq:t_2_markov_bounds}
T_2^{(\Markov)}
\leq (n/2) \sqrt{\qcond} \rme^2 d^{1/q} \bConst{A} \rme^{- a\alpha n/24} \norm{\thetainit- \thetas} \eqsp.
\end{equation}
It remains to proceed with $T_3^{(\Markov)}$. Using the representation \eqref{eq:jn0_main},
\begin{equation}
\textstyle \sum_{t=n/2}^{n-1}\Hnalpha{t+1}{0} = \sum_{t=n/2}^{n-1}\{\Id - \alpha \funcA{\State_{t+1}}\}\Hnalpha{t}{0} - \alpha \sum_{t=n/2}^{n-1} \zmfuncAw[t+1]\Jnalpha{t}{0}\eqsp,
\end{equation}
which yields
\begin{equation}
\textstyle \sum_{t=n/2}^{n-1} \zmfuncAw[t+1]\Jnalpha{t}{0} = \alpha^{-1}(\Hnalpha{n/2}{0} - \Hnalpha{n}{0}) - \sum_{t=n/2}^{n-1}\funcA{\State_{t+1}}\Hnalpha{t}{0}\eqsp.
\end{equation}
Applying again Minkowski's inequality, we get
\begin{equation}
\textstyle \PE_{\xi}^{1/p}\bigl[\norm{\sum_{t=n/2}^{n-1}\zmfuncAw[t+1]\Jnalpha{t}{0}}^{p}\bigr] \leq \{2\alpha^{-1}  + (n/2)\bConst{A}\}\sup_{t \in\nsets}\PE_{\xi}^{1/p}\bigl[\norm{\Hnalpha{t}{0}}^{p}\bigr]\eqsp.
\end{equation}
Now it remains to combine the bounds above in \eqref{eq:err_representation_markov}, use \Cref{prop:J_n_1_2_bound_markov}, and set $q = p(1+\log{d})$.
\endproof


%% file: appendix_markov_technical.tex
Recall that $S_{\ell+1:\ell+m}$ is defined, for $\ell,m \in \nsets$, as
\begin{equation}
\label{eq:S_ell_n_def_tech_markov}
\textstyle
S_{\ell+1:\ell+m} =  \sum_{k = \ell+1}^{\ell+m} \funcBw_k(\State_{k}) \eqsp, \text{ with } \funcBw_k(z) =  (\Id - \alpha \bA)^{\ell+m-k} \zmfuncA{z} (\Id - \alpha \bA)^{k-1 - \ell} \eqsp.
\end{equation}
\begin{lemma}\vspace{-5mm}
\label{lem:S_l_bound_lin}
Assume \Cref{assum:A-b}, \Cref{assum:noise-level}, and \Cref{assum:drift}. Then, for any $p \geq 2$, any initial probability $\xi$ on $(\Zset,\Zsigma)$,  $\ell,m \in \nsets$, it holds that
\begin{equation}
\label{eq:s_l_n_moment_markov}
\PE^{1/p}_{\xi}[\norm{S_{\ell+1:\ell+m} \funnoisew(\State_{\ell})}^{p}] \leq \ConstDM_{S} m^{1/2} (1-\alpha a)^{(m-1)/2} \sqrt{\taumix p} \supconsteps\eqsp,
\end{equation}
where $\ConstDM_{S}= 16 \qcond \bConst{A}$.
\end{lemma}
\proof{Proof.}
Now, with $\mcf_\ell = \sigma\{\State_{j}, j \leq \ell\}$, it holds that
\begin{align}
\PE^{1/p}_{\xi}[\norm{S_{\ell+1:\ell+m} \funnoisew(\State_{\ell})}^{p}]
&= \PE_{\xi}^{1/p}\bigl[\norm{\funnoisew(\State_{\ell})}^{p} \CPE{S_{\ell+1:\ell+m} \funnoisew(\State_{\ell})/\norm{\funnoisew(\State_{\ell})}}{\mcf_\ell}\bigr] \\
&\leq \PE_{\xi}^{1/p}\bigl[\norm{\funnoisew(\State_{\ell})}^{p} \sup_{u \in \sphere^{d-1}, \,  \xi^{\prime} \in \mathcal{P}(\msz)}\PE_{\xi^{\prime}}[\norm{S_{\ell+1:\ell+m} u}^{p}]\bigr]\eqsp,
\end{align}
where $\mathcal{P}(\msz)$ denotes the set of probability measure on $(\msz,\mcz)$. Combining the above bounds with \Cref{lem:S_l_bound} and \Cref{assum:noise-level} yields the statement.
\endproof

\begin{lemma}
\label{lem:S_l_bound}
Assume \Cref{assum:A-b}, \Cref{assum:noise-level}, and \Cref{assum:drift}. For any $\ell,m \in \nsets$,  $t \geq 0$, $u \in \sphere^{d-1}$, and initial probability $\xi$ on $(\Zset,\Zsigma)$, it holds that
\begin{equation}
\label{eq:s_l_n_deviation_bound_markov}
\PP_{\xi}\biggl(\normop{S_{\ell+1:\ell+m} u} \geq t\biggr) \leq 2\exp\left\{-\frac{t^2}{2 \gamma_{m}^{2}}\right\}\eqsp, \text{ where } \gamma_{m} = 8 \qcond \bConst{A} [m \taumix (1-\alpha a)^{m-1}]^{1/2}\eqsp.
\end{equation}
Moreover,
\begin{align}
\textstyle \sup_{u \in \sphere^{d-1}}\PE^{1/p}_{\xi}[\norm{S_{\ell+1:\ell+m} u}^{p}] \leq 16 \qcond \bConst{A} [m \taumix (1-\alpha a)^{m-1} p]^{1/2}\eqsp.
\end{align}
\end{lemma}
\proof{Proof.}
Define $g_{k}(z) : \zset \mapsto \rset^{d}$ as $g_{k}(z) = \funcBw_{k}(z)u$ where $\funcBw_{k}$ is given in \eqref{eq:S_ell_n_def_tech_markov}.
Note that under \Cref{assum:A-b} and applying \Cref{fact:Hurwitzstability}, $\invariantQ(g_{k}) = 0$  and $\sup_{z\in\msz} \norm{g_k(z)} \leq \qcond \bConst{A} (1-\alpha a)^{(m-1)/2}$ for any $k \in \{\ell+1,\ldots,\ell+m\}$. The proof then follows from \Cref{lem:bounded_differences_norms_markovian} and \Cref{lem:bound_subgaussian}.
\endproof

\begin{lemma}
  \label{lem:vector_valued_burkholder}
  Let  $(\Omega,\mathfrak{G},\PP)$ be a probability space, $\{W_k,\Ws_k\}_{k\in \nset}$ be a sequence of $\msz^2$-valued random variables, and $\sequence{\checkA}[k][\{2,\ldots,N+1\}]$ be a sequence of $d\times d$ random matrices. Denote $\mathfrak{G}_k = \sigma(W_{\ell}, \ell \geq k)$  for $k \in\nsets$. Assume that for $k \in \nsets$,  that $\checkA_k$ is $\mathfrak{G}_k$-measurable and $\sigma(\Ws_k)$ and $\mathfrak{G}_{k+1}$ are independent. Then, for any family of measurable  functions $\{g_k\}_{k=1}^N$ from $\msz$ to $\rset^d$, with $\max_{k\in\{1,\ldots,N\}}\supnorm{g_k} \leq 1$, and $p \geq 2$,
\begin{multline}
\PE^{1/p}\big[\norm{\sum\nolimits_{k=1}^{N} \checkA_{k+1} g_{k}(\Ws_k)}^p\big] \\ \leq
2p \bigl\{\sum\nolimits_{k=1}^{N}\sup_{u \in \sphere^{d-1}}\PE^{2/p}\bigl[\norm{\checkA_{k+1} u}^{p}\bigr]\bigr\}^{1/2} + \PE^{1/p}\big[\norm{\sum\nolimits_{k=1}^{N} \checkA_{k+1}\, \PE^{\mathfrak{G}_{k+1}}[ g_k(\Ws_k)]}^p\big]\eqsp.
\end{multline}
\end{lemma}
\proof{Proof.}
 Applying Minkowski's inequality,
\begin{align}
  &  \PE^{1/p}\big[\norm{\sum\nolimits_{k=1}^{N} \checkA_{k+1} g_k(\Ws_k)}^p\big]
  \leq  \PE^{1/p}\big[\norm{\sum\nolimits_{k=1}^{N} \checkA_{k+1} \PE^{\mathfrak{G}_{k+1}}[ g_k(\Ws_k)]}^p\big] \\
  & \qquad \qquad \qquad \qquad \qquad \qquad  + \PE^{1/p}\big[\norm{\sum\nolimits_{k=1}^{N} \checkA_{k+1} \{g_k(\Ws_k) - \PE^{\mathfrak{G}_{k+1}}[ g_k(\Ws_k)]\}}^p\big] \eqsp.
\end{align}
The sequence $\{\checkA_{k}\bigl(g_k(\Ws_k) - \PE^{\mathfrak{G}_{k+1}}[ g_k(\Ws_k)]\bigr)\}_{k=1}^{N}$ is a reversed martingale difference sequence with respect to  $\{\mathfrak{G}_{k}\}_{k \geq 1}$. Hence,
applying the Burkholder inequality (see \citet[Theorem 8.6]{osekowski:2012}), we obtain
\begin{multline}
\PE^{1/p}\big[\norm{\sum\nolimits_{k=1}^{N} \checkA_{k} \{g_k(\Ws_k) - \PE^{\mathfrak{G}_{k+1}}[ g_k(\Ws_k)]\}}^p\big]   \\
\leq p \bigl(\sum\nolimits_{k=1}^{N}\PE^{2/p}[\norm{\checkA_{k+1} \{g_k(\Ws_k) -  \PE^{\mathfrak{G}_{k+1}}[ g_k(\Ws_k)] \}}^{p}]\bigr)^{1/2}\eqsp.
\end{multline}
\endproof

%% file: appendix_technical.tex
In this section we first provide a sharp Rosenthal inequality for the Markov chain $\sequence{\State}[n][\nset]$ under \Cref{assum:drift}. This result is due to \cite[Theorem~1]{moulines23_rosenthal}.
Under \Cref{assum:drift}, 
it is known (see e.g., \cite[Theorem 21.2.10]{douc:moulines:priouret:soulier:2018}) that, for bounded functions $f: \Zset \to \rset^{d}$, linear statistics $n^{-1/2} \sum_{i=0}^{n-1}\{f(\State_i) - \pi(f)\}$ converges in distribution to the
zero-mean Gaussian distribution with variance given by  
\begin{equation}
    \label{eq:def_sigma_pi_f}
    \sigma_{\pi}^2(f) = \txts \lim_{n\to\infty} n^{-1}\PE[\normLine{\sum_{i=0}^{n-1}\{f(\State_i) - \pi(f)\}}^2] \eqsp.
\end{equation}
\begin{theorem}
\label{theo:rosenthal_uge_arbitrary_init}
Assume \Cref{assum:drift}. Then, for any measurable function $f :\msz\to \rset^{d}$, $ \supnorm{f}  \leq 1$,  $p \geq 2$, and $n \geq 1$, it holds
\begin{equation}
\label{eq:rosenthal}
\txts \PE^{1/p}_{\xi}[\norm{\sum_{i=0}^{n-1} f(\State_i)- \pi(f)}^{p}]
\leq \bConst{\sf{Rm}, 1} \sqrt{2} p^{1/2} n^{1/2} \sigma_\pi(f)   + \bConst{\sf{Ros}, 1}^{(\Markov)} n^{1/4}\taumix^{3/4}p\log_2(2p) + \bConst{\sf{Ros}, 2}^{(\Markov)} \taumix p \log_2(2p) \eqsp,
\end{equation}
where
\begin{equation}
\label{eq:Const_ros_1_2}
\bConst{\sf{Ros}, 1}^{(\Markov)} = \frac{16\sqrt{19}}{3\sqrt{3}} \bConst{\sf{Rm}, 1}^{5/2}, \quad \bConst{\sf{Ros}, 2}^{(\Markov)} = 64 (\bConst{\sf{Rm}, 1}^2\bConst{\sf{Rm}, 2}^{1/2}+\bConst{\sf{Rm}, 2})\eqsp,
\end{equation}
where the constants
$\bConst{\sf{Rm}, 1}, \bConst{\sf{Rm}, 2}$ are given in \Cref{appendix:constants} and $\sigma^2_{\pi}(f)$ is defined in \eqref{eq:def_sigma_pi_f}.
\end{theorem}

Now we provide a standard moment bounds for sub-Gaussian random variable, which is proven for completeness.
\begin{lemma}
\label{lem:bound_subgaussian}
Let $X$ be an $\rset^{d}$-valued random variable satisfying $\PP(\norm{X} \geq t) \leq 2 \exp(-t^2/(2\sigma^2))$ for any $t \geq 0$ and some $\sigma^2 >0$. Then, for any $p \geq 2$, it holds that $ \PE[\normLine{X}^p] \leq 2 p^{p/2}\sigma^p$.
\end{lemma}
\proof{Proof.}
Using Fubini's theorem and the change of variable formula, \begin{equation}
\textstyle \PE[\normLine{X}^p] = \int_{0}^{\infty} p t^{p-1} \PP(\normLine{X} \geq t) \, \rmd t = p2^{p/2}  \sigma^{p}\Gammabf(p/2)\eqsp,
\end{equation}
where $\Gammabf$ is the Gamma function. It remains to apply the bound $\Gammabf(p/2) \leq (p/2)^{p/2-1}$, which holds for $p \geq 2$ due to \citet[Theorem 1.5]{anderson:qiu:1997}.
\endproof
Now we present the general version of Hoeffding inequality for martingale-difference sequences, taking values in Banach spaces. This result is due to \citet[Theorem 3.5]{pinelis_1994}. Below we specify this inequality to the special case of sum of zero-mean independent random vectors.
\begin{lemma}
\label{lem:bounded_differences_norms}
Let $X_1,\ldots,X_n \in \rset^{d}$ be independent random vectors satisfying $\norm{X_i} \leq \beta_{i}$ $\PP$-a.s. and $\PE[X_i] = 0$, $i \in \{1,\ldots,n\}$. Then, for any $t \geq 0$, it holds
\begin{equation}
\label{eq:prob_for_norms}
\PP\parenthese{ \norm{\sum_{i=1}^{n}X_{i}} \geq t} \leq 2\exp\left\{-\frac{t^2}{2\sum_{j=1}^{n}\beta_{j}^{2}}\right\}\eqsp.
\end{equation}
\end{lemma}
The result above can be generalized for bounded $\rset^{d}$-valued functions of the Markov chains with kernel satisfying \Cref{assum:drift}.

\begin{lemma}
\label{lem:bounded_differences_norms_markovian}
Assume \Cref{assum:drift}. Let $\{g_i\}_{i=1}^n$ be a family of measurable functions from $\Zset$ to  $\rset^{d}$ such that $\supnorm{g}= \max_{i \in\{1,\ldots,n\}}\supnorm{g_i} < \infty$
and $\pi(g_i)= 0$ for any $i \in\{1,\ldots,n\}$.
Then, for any initial probability $\xi$ on $(\Zset,\Zsigma)$, $n \in \nset$, $t \geq 0$, it holds
\begin{equation}
\label{eq:prob_for_norms_markov}
\PP_{\xi}\biggl(\normop{\sum\nolimits_{i=1}^{n}g_i(\State_{i})}\geq t\biggr) \leq 2 \exp\biggl\{-\frac{t^2}{2 u_n^{2}}\biggr\}\eqsp, \text{ where } u_n = 8 \supnorm{g} \sqrt{n} \sqrt{\taumix}\eqsp.
\end{equation}
\end{lemma}
\proof{Proof.} The function $\varphi(\state_1,\dots,\state_n) := \norm{\sum_{i=1}^{n}g_i(\state_{i})}$ on $\Zset^n$ satisfies the bounded differences property. Moreover, $(1/2)\sup_{z,z' \in \Zset} \norm{\MKQ^{\taumix}(z, \cdot) - \MKQ^{\taumix}(z',\cdot)}[\sf{TV}] \leq 1/4$ by definition of $\taumix$. Thus, applying \citet[Corollary 2.10]{paulin_concentration_spectral}, we get for $t \geq \PE_{\xi}[\norm{\sum_{i=1}^{n}g_i(\State_{i})}]$,
\begin{align}
\PP_{\xi}\biggl(\norm{\sum\nolimits_{i=1}^{n}g_i(\State_{i})} \geq t\biggr) \leq  \exp\left\{-\frac{2(t-\PE_{\xi}[\norm{\sum_{i=1}^{n}g_i(\State_{i})}])^{2}}{9n\supnorm{g}^2 \taumix }\right\}\eqsp.
\end{align}
It remains to upper bound $\PE_{\xi}[\norm{\sum_{i=1}^{n}g_i(\State_{i})}]$. Note that
\begin{align}
\PE_{\xi}[\norm{\sum\nolimits_{i=1}^{n}g_i(\State_{i})}^{2}] = \sum\nolimits_{i=1}^n  \PE_{\xi}[\norm{g_i(\State_{i})}^{2}] + 2\sum\nolimits_{k=1}^{n-1}\sum\nolimits_{\ell = 1}^{n-k} \PE_{\xi}[g_k(\State_{k})^{\top} g_{k+\ell}(\State_{k+\ell})]\eqsp.
\end{align}
and, using \Cref{assum:drift} and $\invariantQ(g_{k+\ell}) = 0$, we obtain
\begin{align}
\textstyle
 \bigl\vert \PE_{\xi}[g_k(\State_{k})^{\top} g_{k+\ell}(\State_{k+\ell})] \bigr\vert &= \absD{ \int_{\Zset}g_k(z)^{\top}\left(\MKQ^{\ell}g_{k+\ell}(z) - \invariantQ(g_{k+\ell})\right) \xi\MKQ^{k}(\rmd z) } \leq \supnorm{g}^{2} \dobru{\MKQ^{\ell}}\eqsp.
\end{align}
Together with \eqref{eq:crr_koef_sum_tau_mix}, this implies
\begin{align}
\textstyle
\sum_{k=1}^{n-1}\sum_{\ell = 1}^{n-k}\abs{ \PE_{\xi}[g_k(\State_{k})^{\top} g_{k+\ell}(\State_{k+\ell})] } \leq \sum_{k=1}^{n-1}\supnorm{g}^{2} \dobru{\MKQ^{\ell}}  \leq (4/3) \abs{g}_{\infty}^2 \taumix n \eqsp.
\end{align}
Combining the bounds above, we upper bound $\PE_{\xi}[\norm{\sum_{i=1}^{n}g_{i}(\State_{i})}]$ as
\begin{align}
\PE_{\xi}[\norm{\sum\nolimits_{i=1}^{n}g_i(\State_{i})}] &\leq \bigl\{ \PE_{\xi}[\norm{\sum\nolimits_{i=1}^{n}g_i(\State_{i})}^2] \bigr\}^{1/2} \leq
2 \sqrt{n} \supnorm{g} \sqrt{\taumix} =: v_n \eqsp.
\end{align}
Plugging this result in \eqref{eq:prob_for_norms_markov},  we obtain that
\begin{equation}
\label{eq:MacDiarmid_markov_new}
\PP_{\xi}\biggl(\norm{\sum\nolimits_{i=1}^{n}g_i(\State_{i})} \geq t\biggr) \leq
\begin{cases}
1, \quad t < v_n, \\
 \exp\left\{-\frac{2(t-v_{n})^{2}}{3 v_{n}^2}\right\}\eqsp, \quad t \geq v_{n}\eqsp.
\end{cases}
\end{equation}
Now it is easy to see that \rhs\ of \eqref{eq:MacDiarmid_markov_new} is upper bounded by $2\exp\{-t^2/(8 v_n^2)\}$ for any $t \geq 0$, and the statement follows.
\endproof

\newpage 
